\documentclass[reqno,10pt,a4paper]{amsart}
\usepackage{amscd}
\usepackage{amsmath,amsthm,amssymb}
\usepackage[percent]{overpic}
\numberwithin{equation}{section}
\usepackage{times}
\usepackage{bbm}
\usepackage{booktabs}
\usepackage{enumitem}
\usepackage[left=1.1in,right=1.1in,top=1.13in,bottom=1.13in]{geometry} 
\usepackage{calc}
\usepackage{url}
\usepackage[normalem]{ulem}
\usepackage{ifthen,tabularx,graphicx,multirow}
\graphicspath{ %
	{Figures/} %
	{Figures/Zoom} %
}

\usepackage{rotating}
\usepackage{csquotes}
\usepackage[normalem]{ulem}
\usepackage{listings}
\usepackage{etoolbox}
\usepackage{tcolorbox}
\usepackage{enumitem}
\usepackage{scrextend}
\usepackage{amsmath}
\usepackage{amsthm}
\usepackage{algorithm2e}
\usepackage{cite}
\newcolumntype{Y}{>{\centering\arraybackslash}X}
\newtheorem{lemma}{Lemma}

\newtheorem{theorem}{Theorem}
\newtheorem{definition}{Definition}
\numberwithin{lemma}{section}

\theoremstyle{definition}
\newtheorem{remark}[lemma]{Remark}

	\definecolor{ddg}{rgb}{0.2, 0.2, 0.2}
	\definecolor{llg}{rgb}{0.95, 0.95, 0.95}

\usepackage[margin=0.3cm]{caption}
\usepackage{pdfpages}
\newcount\colveccount
\newcommand*\colvec[1]{
	\global\colveccount#1
	\begin{pmatrix}
		\colvecnext
	}
	\def\colvecnext#1{
		#1
		\global\advance\colveccount-1
		\ifnum\colveccount>0
		\\
		\expandafter\colvecnext
		\else
	\end{pmatrix}
	\fi
}

\usepackage{mathtools}

\DefineNamedColor{named}{Purple}{cmyk}{0.45,0.86,0,0}

\DefineNamedColor{named}{orange} {rgb}{1,0.55,0}

\DeclareMathOperator{\argmin}{arg\!\min}

\usepackage[hidelinks]{hyperref}
\usepackage[foot]{amsaddr}
\makeatletter
\newcommand{\ssymbol}[1]{^{\@fnsymbol{#1}}}
\makeatother

\usepackage{xcolor}
\usepackage{pict2e}
\usepackage{booktabs}
\usepackage{tikz}
\tikzstyle{stuff_fill}=[rectangle,draw,fill=pink,minimum size=0.5em]
\usetikzlibrary{calc}
\usetikzlibrary{decorations.pathreplacing,calc}

\newcommand{\tikzmark}[2][-3pt]{\tikz[remember picture, overlay, baseline=-0.5ex]\node[#1](#2){};}

\newcounter{arrow}
\newcommand{\drawcurvedarrow}[3][]{%
 \refstepcounter{arrow}
 \tikz[remember picture, overlay]\draw (#2.center)edge[#1]node[coordinate,pos=0.5, name=arrow-\thearrow]{}(#3.center);
}

\newcommand{\annote}[3][]{%
 \tikz[remember picture, overlay]\node[#1] at (#2) {#3};
}

\title[The barriers of deep learning and Smale's 18th problem]{Can stable and accurate neural networks be computed? -- On the barriers of deep learning and
Smale's 18th problem}

\author{Matthew J. Colbrook$\ssymbol{1}$$\ssymbol{2}$}
\author{Vegard Antun$\ssymbol{1}$$\ssymbol{3}$}
\author{Anders C. Hansen$\ssymbol{2}$$\ssymbol{3}$}

\email{m.colbrook@damtp.cam.ac.uk}
\email{vegarant@math.uio.no}
\email{a.hansen@damtp.cam.ac.uk}

\address{$\ssymbol{1}$M.J.C. and V.A. contributed equally to this work.}
\address{$\ssymbol{2}$Department of Applied Mathematics and Theoretical Physics, University of Cambridge.}
\address{$\ssymbol{3}$Department of Mathematics, University of Oslo.}

\begin{document}
\begin{abstract}
Deep learning (DL) has had unprecedented success and is now entering scientific computing with full force. However, current DL methods typically suffer from instability, even when universal approximation properties guarantee the existence of stable neural networks (NNs). We address this paradox by demonstrating basic well-conditioned problems in scientific computing where one can prove the existence of NNs with great approximation qualities, however, there does not exist any algorithm, even randomised, that can train (or compute) such a NN. For any positive integers $K > 2$ and $L$, there are cases where simultaneously: (a) no randomised training algorithm can compute a NN correct to $K$ digits with probability greater than $1/2$, (b) there exists a deterministic training algorithm that computes a NN with $K-1$ correct digits, but any such (even randomised) algorithm needs arbitrarily many training data, (c) there exists a deterministic training algorithm that computes a NN with $K-2$ correct digits using no more than $L$ training samples. These results imply a classification theory describing conditions under which (stable) NNs with a given accuracy can be computed by an algorithm. We begin this theory by establishing sufficient conditions for the existence of algorithms that compute stable NNs in inverse problems. We introduce Fast Iterative REstarted NETworks (FIRENETs), which we both prove and numerically verify are stable. Moreover, we prove that only $\mathcal{O}(|\log(\epsilon)|)$ layers are needed for an $\epsilon$-accurate solution to the inverse problem.
\end{abstract}


\maketitle

\begin{figure*}[t]
    \centering
    \setlength{\tabcolsep}{2pt} 
    \begin{tabular}{@{}>{\centering}m{0.24\textwidth}>{\centering}m{0.24\textwidth}>{\centering}m{0.24\textwidth}>{\centering\arraybackslash}m{0.24\textwidth}@{}}   
        U-net rec.\ from $y=Ax$ 
      & U-net rec.\ from $y = Ax+e_1$ 
      & U-net rec.\ from $y = Ax+e_2$ 
      & Original image $x$ \\
        (cropped) 
      & (cropped) 
      & (cropped) 
      & (full size)  \\
      \begin{overpic}[width=\linewidth]{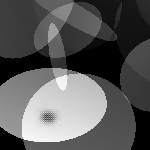}
			\linethickness{3pt}
\put(20,52){\color{green}\vector(1,-2.5){10}}
\end{overpic}
    & \begin{overpic}[width=\linewidth]{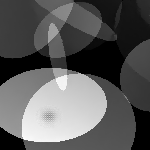}
		\linethickness{3pt}
\put(20,52){\color{green}\vector(1,-2.5){10}}
\end{overpic}
    & \includegraphics[width=\linewidth]{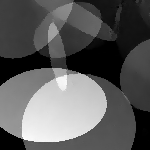}
    & \includegraphics[width=\linewidth]{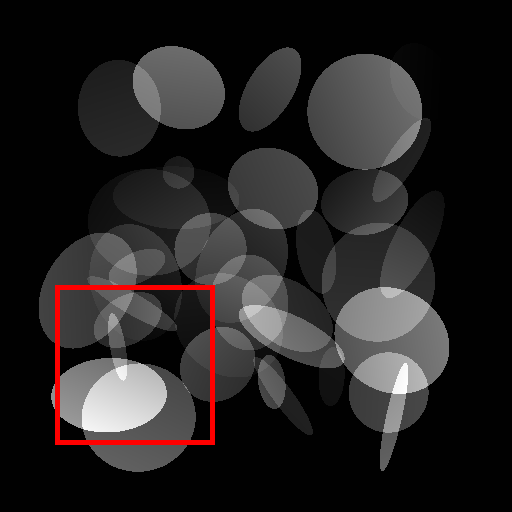} \\
        FIRENET rec.\ from $y=Ax$ 
      & FIRENET rec.\ from $y = Ax+e_1$ 
      & FIRENET rec.\ from $y = Ax+e_2$ 
      & Original image $x$ \\
        (cropped) 
      & (cropped) 
      & (cropped) 
      & (cropped)  \\
	  \includegraphics[width=\linewidth]{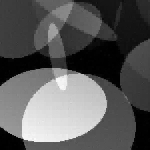}
    & \includegraphics[width=\linewidth]{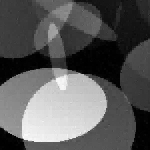}
    & \includegraphics[width=\linewidth]{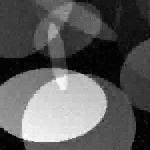}
    & \includegraphics[width=\linewidth]{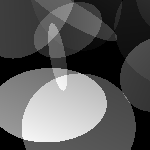} \\
    \end{tabular}
    \vspace{-2mm}\caption{(\textbf{AI-generated hallucinations}) A trained NN, based on a U-net architecture and trained on a set of ellipses images, generates a black area in a white ellipse (upper left image, shown as green arrow) when reconstructing the original image $x$ from noiseless measurements. By adding random Gaussian noise $e_1$ and $e_2$ (where $\|e_1\|_{l^2}/\|e_2\|_{l^2} \approx 2/5$) to the measurements, we see that the trained NN removes the aspiring black ellipse (row 1, columns 2-3). FIRENET on the other hand is completely stable with and without random Gaussian noise (row 2, cloumn 1-3 ). In the forth column,  we show the original image $x$, with a red square indicating the cropped area. In this example $A \in \mathbb{C}^{m\times N}$ is a subsampled discrete Fourier transform with $m/N \approx 0.12$. }
\label{fig:Aihallucination}
\end{figure*}

Deep learning (DL) has demonstrated unparalleled accomplishments in fields ranging from image classification and computer vision \cite{krizhevsky2012imagenet}, voice recognition and automated diagnosis in medicine \cite{hinton2012deep}, to inverse problems and image reconstruction \cite{kamilov2015learning,jin17,hammernik2018learning}. However, there is now overwhelming empirical evidence that current DL techniques typically lead to unstable methods, a phenomenon that seems universal and present in all of the applications listed above \cite{SzZ-14,moosavi2016deepfool,antun2020instabilities, finlayson2019adversarial,tyukin2020adversarial}, and in most of the new artificial intelligence (AI) technologies. These instabilities are often detected by what has become commonly known in the literature as ``adversarial attacks''. Moreover, the instabilities can be present even in random cases and not just worst-case scenarios \cite{gottschling2020troublesome} - see Figure \ref{fig:Aihallucination} for an example of \textit{AI-generated hallucinations}. There is a growing awareness of this problem in high-stake applications and society as a whole \cite{finlayson2019adversarial,baker2019workshop,hamonrobustness}, and instability seems to be the Achilles' heel of modern AI and DL -- see Figure \ref{fig:automap_pert} (top row). For example, this is a problem in real-world clinical practice. Facebook and NYU's 2019 FastMRI challenge reported that networks that performed well in terms of standard image quality metrics were prone to false negatives, failing to reconstruct small, but physically-relevant image abnormalities \cite{knoll2020advancing}. {AI-generated hallucinations} poses a serious danger in applications such as medical diagnosis. Subsequently, the 2020 FastMRI challenge \cite{muckley2020state} focused on pathologies, noting, ``\textit{Such hallucinatory features are not acceptable and especially problematic if they mimic normal structures that are either not present or actually abnormal. Neural network models can be unstable as demonstrated via adversarial perturbation studies \cite{antun2020instabilities}.}''

Nevertheless, classical approximation theorems show that a continuous function can be approximated arbitrarily well by a neural network (NN). Thus, stable problems described by stable functions can always be solved stably with a NN. This leads to the following fundamental question:

{\textbf{Question:}\it Why does DL lead to unstable methods and AI-generated hallucinations, even in scenarios where one can prove that stable and accurate neural networks exist?}

\subsection*{Foundations of AI for inverse problems} To answer the above question we initiate a program on the foundations of AI, determining the limits of what DL can achieve in inverse problems. It is crucial to realise that an existence proof of suitable NNs does not always imply that they can be constructed by a training algorithm. Furthermore, it is not difficult to compute stable NNs. For example, the zero network is stable, but not particularly useful. The big problem is to compute NNs that are both accurate and stable \cite{devore2020neural,adcock2020gap}. Scientific computing itself is based on the pillars stability and accuracy. However, there is often a trade-off between the two. There may be barriers preventing the existence of accurate and stable algorithms, and sometimes accuracy must be sacrificed to secure stability.

\subsection*{Main results}
\begin{figure*}[t!]
    \setlength{\tabcolsep}{2pt} 
    \begin{tabular}{@{}>{\centering}m{0.24\textwidth}>{\centering}m{0.24\textwidth}>{\centering}m{0.24\textwidth}>{\centering\arraybackslash}m{0.24\textwidth}@{}}   
     $ \Psi(Ax)$ &  $\Psi(Ax+e_1)$ &  $\Psi(Ax + e_2)$  & $\Psi(Ax + e_3)$\\
     \includegraphics[width=\linewidth]{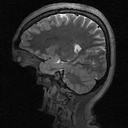}
    &\includegraphics[width=\linewidth]{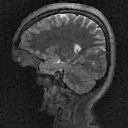}
    &\includegraphics[width=\linewidth]{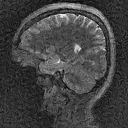}
    &\includegraphics[width=\linewidth]{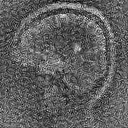}\\
     $ \Phi(Ax)$ &  $\Phi(Ax+\tilde{e}_1)$ &  $\Phi(Ax + \tilde{e}_2)$  & $\Phi(Ax + \tilde{e}_3)$\\
     \includegraphics[width=\linewidth]{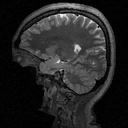}
    &\includegraphics[width=\linewidth]{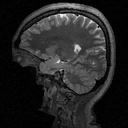}
    &\includegraphics[width=\linewidth]{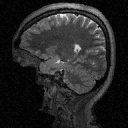}
    &\includegraphics[width=\linewidth]{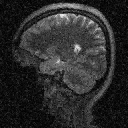}\\
\end{tabular}\vspace{-2mm}
\caption{\textbf{Top row} (\textbf{Unstable neural network in image reconstruction}): The neural network AUTOMAP (\textit{Nature} (2018) \cite{Bo-18}) represents the tip of the iceberg of DL in inverse problems. The paper promises that one can ``... observe superior immunity to noise...''. Moreover, the follow-up announcement (\textit{Nature Methods} ``\textit{AI transforms image reconstruction}," \cite{Str-18}) proclaims: ``A deep-learning-based approach improves speed, accuracy and robustness of biomedical image reconstruction". However, {as we see in the first row},  the AUTOMAP reconstruction $\Psi(Ax+e_j)$ from the subsampled {noisy} Fourier MRI data $Ax+e_j$, is completely unstable. Here, {$A \in \mathbb{C}^{m \times N}$ is a subsampled Fourier transform,} $x$ is the original image and the $e_j$'s are perturbations meant to simulate worst-case effect. Note that the condition number $\mathrm{cond}(AA^*) = 1$, so the instabilities are not caused by poor condition. The network weights were provided by the authors of \cite{Bo-18} (trained on brain images from the MGH--USC HCP dataset \cite{fan2016mgh}). 
\\\textbf{Bottom row} (\textbf{The FIRENET is stable to worst-case perturbations}): Using the same method, we compute perturbations $\tilde{e}_j$ to simulate worst-case effect for the FIRENET $\Phi \colon \mathbb{C}^m \to \mathbb{C}^{N}$. {As can be seen from the figure, FIRENET is stable to these worst-case perturbations.} Here $x$ and $A \in \mathbb{C}^{m\times N}$ are the same image and sampling matrix as for AUTOMAP. Moreover, for each $j=1,2,3$ we have ensured that $\|\tilde{e}_j\|_{l^2} \geq \|e_j\|_{l^2}$, where the $e_j$'s are the perturbations for AUTOMAP (we have denoted the perturbations for FIRENET by $\tilde{e}_j$ to emphasise that these adversarial perturbations are sought for FIRENET and have nothing to do with the perturbations for AUTOMAP).}
\label{fig:automap_pert} \vspace{-2mm}
\end{figure*}

We consider the canonical inverse problem of an underdetermined system of linear equations:
\begin{equation}
\label{eq:inv_prob0}
\mbox{Given measurements $y = A x + e \in \mathbb{C}^m$, recover $x \in \mathbb{C}^N$.}
\end{equation}
Here, $A \in \mathbb{C}^{m \times N}$ represents a sampling model ($m < N$), such as a subsampled DFT in Magnetic Resonance Imaging (MRI). The problem in \eqref{eq:inv_prob0} forms the basis for much of inverse problems and image analysis. The possibility of $y\neq Ax$ models noise or perturbations. Our results demonstrate fundamental barriers preventing NNs (despite their existence) from being computed by algorithms. This helps shed light on the intricate question of why current algorithms in DL produce unstable networks, despite the fact that stable NNs often exist in the particular application. We show:
\begin{enumerate}[leftmargin=12pt]
\setlength\itemsep{0pt}%
    \setlength\parskip{5pt}%
	\item Theorems \ref{thrm:exists} and \ref{impossibility_theorem}: There are well-conditioned problems (suitable condition numbers bounded by $1$) where, paradoxically, mappings from training data to suitable NNs exist, but no training algorithm (even randomised) can compute approximations of the NNs from the training data.
	\item Theorem \ref{impossibility_theorem}: The existence of algorithms computing NNs depends on the desired accuracy. For any $K\in\mathbb{Z}_{\geq 3}$, there are well-conditioned classes of problems where simultaneously: (i) algorithms may compute NNs to $K-1$ digits of accuracy, but not $K$, (ii) achieving $K-1$ digits of accuracy requires arbitrarily many training data, (iii) achieving $K-2$ correct digits requires only one training datum.
	\item Theorems \ref{firenet_informal} and \ref{NN_FW_recov_thm}: Under specific conditions that are typically present in, for example, MRI, there are algorithms that compute stable NNs for the problem in \eqref{eq:inv_prob0}. These NNs, which we call {F}ast {I}terative {RE}started {NET}works (FIRENETs), converge exponentially in the number of hidden layers. Crucially, we prove that FIRENETs withstand adversarial attacks, see Figure \ref{fig:automap_pert} (bottom row), and they can even be used to stabilise unstable NNs, see Figure \ref{fig:stabilisation}.
	\item There is a trade-off between stability and accuracy in DL, with limits on how well a stable NN can perform in inverse problems. Figure \ref{fig:false_neg} demonstrates this with a U-net trained on images consisting of ellipses and which is quite stable. However, when a detail not in the training set is added, it washes it out almost entirely. FIRENETs offer a blend of both stability and accuracy. However, they are by no means the end of the story. Tracing out the optimal stability vs. accuracy trade-off is crucial for applications and will no doubt require a myriad of different techniques to tackle different problems and stability tolerances.
\end{enumerate}

\section*{Fundamental barriers}

We first consider basic mappings used in modern mathematics of information, inverse problems and optimisation. Given a matrix $A \in \mathbb{C}^{m \times N}$ and a vector $y \in \mathbb{C}^m$, we consider the following three popular minimisation problems:\\
\noindent\vspace*{-0.7cm}\begin{align}
(P_1)\quad&\argmin_{x \in \mathbb{C}^{N}} F_{1}^A(x)\coloneqq \|x\|_{l^1_w},\text{ s.t. }\|Ax-y\|_{l^2}\!\leq\! \epsilon,\label{def_P1}\\
(P_2)\quad&\argmin_{x \in \mathbb{C}^{N}} F_2^A(x,y,\lambda)\coloneqq \lambda\|x\|_{l^1_w}+\|Ax-y\|_{l^2}^2,\label{def_P2}\\
(P_3)\quad&\argmin_{x \in \mathbb{C}^{N}} F_3^A(x,y,\lambda)\coloneqq \lambda\|x\|_{l^1_w}+\|Ax-y\|_{l^2},\label{def_P3}
\end{align}

\noindent{}known respectively as quadratically constrained basis pursuit (we always assume existence of a feasible $x$ for $(P_1)$), unconstrained LASSO and unconstrained square-root LASSO. Such sparse regularisation problems are often used as benchmarks for \eqref{eq:inv_prob0} and we prove impossibility results for approximations of the solution maps of $(P_j)$. This is a striking computational barrier given that one can prove the existence of accurate neural networks (Theorem \ref{thrm:exists}).

The parameters $\lambda$ and $\epsilon$ are positive rational numbers, and the weighted $l^1_w$ norm is given by
$
\|x\|_{l^1_w}\coloneqq \sum_{l=1}^Nw_l|x_{l}|,
$
where each weight $w_j$ is a positive rational. Throughout, we let 
\begin{equation}\label{eq:Ay}\setlength\abovedisplayskip{6pt}\setlength\belowdisplayskip{6pt}
\Xi_j(A,y) \text{ be the set of minimisers for } (P_j).
\end{equation}
Let $A\in\mathbb{C}^{m\times N}$ and let $ \mathcal{S} = \{y_k\}_{k=1}^{R} \subset \mathbb{C}^m$ be a collection of samples ($R\in\mathbb{N}$). We consider the following key question:

\vspace{1mm}
\textbf{Question: }\textit{Given a collection $\Omega$ of such pairs $(A, \mathcal{S})$, does there exists a neural network approximating $\Xi_j$, and if so, can such an approximation be trained by an algorithm?}
\vspace{1mm}

To make this question precise, note that $A$ and samples in $\mathcal{S}$ will typically never be exact, but can be approximated/stored to arbitrary precision. For example, this would occur if $A$ was a subsampled discrete cosine transform. Thus, we assume access to rational approximations $\{y_{k,n}\}_{k=1}^R$ and $A_n$ with
\begin{equation}
\label{delt1_inf1}
\|y_{k,n} - y_k\|_{l^2} \leq 2^{-n},\quad \|A_n - A\| \leq 2^{-n}, \quad \forall n \in \mathbb{N},
\end{equation}
where $\|\cdot\|$ refers to the usual Euclidean operator norm. The bounds $2^{-n}$ are simply for convenience and can be replaced by any other sequence converging to zero. We also assume access to rational $\left\{x_{k,n}\right\}_{k=1}^R$ with
\begin{equation}
\label{delt1_inf2}
\inf_{x^*\in\Xi_j(A_n,y_{k,n})}\left\|x_{k,n}-x^*\right\|_{l^2} \leq 2^{-n}, \quad \forall n \in \mathbb{N}.
\end{equation}
Hence, the training set associated with $(A,\mathcal{S})\in\Omega$ for training a suitable NN must be 
\begin{equation}
\label{training_TT}
\iota_{A,\mathcal{S}} \coloneqq  \left\{\left(y_{k,n}, A_n, x_{k,n}\right) \, \vert \, k = 1,\hdots, R, \text{ and }n \in \mathbb{N}\right\}.
\end{equation}
The class of all such admissible training data is denoted by 
\[
\Omega_{\mathcal{T}}\coloneqq \left\{   \iota_{A,\mathcal{S}} \text{ as in \eqref{training_TT}} \, \vert \, (A,\mathcal{S})\in\Omega,\text{ (\ref{delt1_inf1}) and (\ref{delt1_inf2}) hold}\right\}.
\]
Statements addressing the above question are summarised in the following two theorems. We use $\mathcal{N}_{m,N}$ to denote the class of NNs from $\mathbb{C}^m$ to $\mathbb{C}^N$. We use standard definitions of feedforward NNs \cite{higham2019deep}, precisely given in the SI Appendix.

\begin{theorem}\label{thrm:exists}
For any collection $\Omega$ of such $(A,\mathcal{S})$ described above, there exists a mapping
\begin{align*}
&\mathcal{K}\colon \Omega_{\mathcal{T}}\rightarrow \mathcal{N}_{m,N}, \quad \mathcal{K}(\iota_{A,\mathcal{S}}) = \varphi_{A,\mathcal{S}},\\
&\text{s.t.}\quad\varphi_{A,\mathcal{S}}(y)\in \Xi_j(A,y),\quad \forall y\in\mathcal{S}.
\end{align*}
In words, $\mathcal{K}$ maps the training data $\Omega_{\mathcal{T}}$ to NNs that solve the optimisation problem $(P_j)$ for each $(A,\mathcal{S})\in\Omega$.
\end{theorem}


Despite the existence of NNs guaranteed by Theorem \ref{thrm:exists}, computing or training such a NN from training data is most delicate. The following is stated precisely and proven in the SI Appendix. We also include results for randomised algorithms, which are common in DL (e.g. stochastic gradient descent).

\begin{theorem}
\label{impossibility_theorem}
Consider the optimisation problem $(P_j)$ for fixed parameters $\lambda\in(0,1]$ or $\epsilon\in(0,1/2]$ and $w_l=1$, where $N \geq 2$ and $m<N$. Let $K > 2$ be a positive integer and let $L \in \mathbb{N}$. Then there exists a class $\Omega$ of elements $(A,\mathcal{S})$ as above, with the following properties. The class $\Omega$ is well-conditioned with relevant condition numbers bounded by 1 independent of all parameters. However, the following hold simultaneously (where accuracy is measured in the $l^2$-norm):
\begin{itemize}[leftmargin=13pt]
\setlength\itemsep{0pt}%
    \setlength\parskip{4pt}%
\item[(i)] {\upshape{\textbf{($K$ digits of accuracy impossible)}}}  There does not exist any algorithm that, given a training set $\iota_{A,\mathcal{S}}\! \in\! \Omega_{\mathcal{T}}$, produces a NN with $K$ digits of accuracy for any element of $\mathcal{S}$. Furthermore, for any $p>1/2$, no probabilistic algorithm (BSS, Turing or any model of computation) can produce a NN with $K$ digits of accuracy with probability at least $p$. 
\item[(ii)] {\upshape{\textbf{($K-1$ digits of accuracy possible but requires arbitrarily many training data)}}} There does exist a deterministic Turing machine that, given a training set $\iota_{A,\mathcal{S}}\! \in\! \Omega_{\mathcal{T}}$, produces a NN accurate to $K-1$ digits over $\mathcal{S}$. However, for any probabilistic Turing machine, $M\in\mathbb{N}$ and $p\in\left[0,\frac{N-m}{N+1-m}\right)$ that produces a NN, there exists a training set $\iota_{A,\mathcal{S}} \in \Omega_{\mathcal{T}}$ such that for all $y\in\mathcal{S}$, the probability of failing to achieve $K-1$ digits or requiring more than $M$ training data is greater than $p$.
\item[(iii)] {\upshape{\textbf{($K-2$ digits of accuracy possible with $L$ training data)}}}  There does exist a deterministic Turing machine that, given a training set $\iota_{A,\mathcal{S}}\in\Omega_{\mathcal{T}}$ and using only $L$ training data from each $\iota_{A,\mathcal{S}}$, produces a NN accurate to $K-2$ digits over $\mathcal{S}$.
\end{itemize}
\end{theorem}

\vspace{-1mm}

\begin{remark}[{\bf Condition}]
The statement in Theorem \ref{impossibility_theorem} refers to standard condition numbers used in optimisation and scientific computing. For precise definitions, see the SI Appendix.
\end{remark}

\vspace{-1mm}

\begin{remark}[{\bf G\"{o}del, Turing, Smale and Theorem \ref{impossibility_theorem}}] Theorem \ref{thrm:exists} and Theorem \ref{impossibility_theorem} demonstrate basic limitations on the existence of algorithms that can compute NNs despite their existence. This relates to Smale's 18th problem, ``What are the limits of intelligence, both artificial and human?'', from the list of mathematical problems for the 21st century \cite{21century_Smale}, which echoes the famous Turing test from 1950 \cite{Turing_1950}. Smale's discussion is motivated by the results of G\"{o}del \cite{godel1931formal} and Turing \cite{turing1937computable} establishing impossibility results on what mathematics and digital computers can achieve \cite{Weinberger}. Our results are actually stronger, however, than what can be obtained with Turing's techniques. Theorem \ref{impossibility_theorem} holds even for any randomised Turing or Blum--Shub--Smale (BSS) machine that can solve the halting problem. It immediately opens up for a classification theory on which NNs can be computed by randomised algorithms. Theorem \ref{firenet_informal} is a first step in this direction. Below, we argue that a program on the foundations of AI, similar to Hilbert's program on the foundations of mathematics, is needed, where impossibility results are provided in order to establish the boundaries of DL and AI.
\end{remark}


\subsection*{Numerical example}
To highlight the impossibility of computing NNs (Theorem \ref{impossibility_theorem}) -- despite their existence by Theorem \ref{thrm:exists} -- we consider the following numerical example.  Consider the problem $(P_3)$, with $w_l=1$ and $\lambda = 1$. Theorem \ref{impossibility_theorem} is stated for a specific input class $\Omega$, and in this example we consider three different such classes $\Omega_{K}$, depending on the accuracy parameter $K$. In Theorem \ref{impossibility_theorem}, we required  that $K>2$ so that $K-2>0$, but this is not necessary to show the impossibility statement (i), so we consider $K=1,3,6$. Full details of the following experiment are given in the SI Appendix.

To show that it is impossible to compute NNs which can solve $(P_3)$ to arbitrary accuracy we consider FIRENETs $\Phi_{A_n}$ (the NNs in Theorem \ref{firenet_informal}) and learned ISTA (LISTA) networks $\Psi_{A_n}$ based on the architecture choice from \cite{chen2018theoretical}. The networks are trained to high accuracy on training data on the form \eqref{training_TT} with $R=8000$ training samples and $n$ given as in Table \ref{tab:inexact}. In all cases $N=20$, $m=N-1$ and the $x_{k,n}$'s minimising $(P_3)$ with input data $(y_{k,n}, A_n)$, are all $6$-sparse. The choice of $N$, $m$ and sparsity is to allow for fast training, other choices are certainly possible. 

Table \ref{tab:inexact} shows the errors for both LISTA and FIRENETs. Both network types are given input data $(y_n,A_n)$, approximating the true data $(y,A)$. As is clear from the table, none of the networks are able to compute an approximation to the true minimiser in $\Xi_{3}(A,y)$ to $K$ digits accuracy. However, both networks compute an approximation with $K-1$ digits accuracy. These observations agree precisely with Theorem \ref{impossibility_theorem}. 

\begin{table}
\begin{tabular}{ccccc}
\toprule 
$ \Psi_{A_n}$ & $\Phi_{A_n}$ & \begin{tabular}{c} $\|A_n-A\| \leq 2^{-n} $ \\ $\|y_n - y\|_{l^2} \leq 2^{-n}$ \end{tabular}  &  $10^{-K}$  & $\Omega_K$ \\
\midrule
 0.2999690 & 0.2597827 & $n = 10$ &   $10^{ -1  }$ & $K = 1$ \\
 0.3000000 & 0.2598050 & $n = 20$ &   $10^{ -1  }$ & $K = 1$ \\
 0.3000000 & 0.2598052 & $n = 30$ &   $10^{ -1  }$ & $K = 1$ \\
 0.0030000 & 0.0025980 & $n = 10$ &   $10^{ -3  }$ & $K = 3$ \\
 0.0030000 & 0.0025980 & $n = 20$ &   $10^{ -3  }$ & $K = 3$ \\
 0.0030000 & 0.0025980 & $n = 30$ &   $10^{ -3  }$ & $K = 3$ \\
 0.0000030 & 0.0000015 & $n = 10$ &   $10^{ -6  }$ & $K = 6$ \\
 0.0000030 & 0.0000015 & $n = 20$ &   $10^{ -6  }$ & $K = 6$ \\
 0.0000030 & 0.0000015 & $n = 30$ &   $10^{ -6  }$ & $K = 6$ \\
\bottomrule \\
\end{tabular}
\vspace{-2mm}
\caption{\label{tab:inexact} (\textbf{Impossibility of computing approximations of the existing neural network to arbitrary accuracy}). We demonstrate statement (i) from Theorem \ref{impossibility_theorem} on FIRENETs $\Phi_{A_n}$, and LISTA networks $\Psi_{A_n}$. The table shows the shortest $l^2$ distance between the output from the networks, and the true solution of the problem $(P_3)$, with $w_l=1$ and $\lambda =1$, for different values of $n$ and $K$. Note that none of the networks can compute the existing correct NN (that exists by Theorem \ref{thrm:exists} and coincides with $\Xi_3$) to $10^{-K}$ digits accuracy, while all of them are able to compute approximations that are accurate to $10^{-K+1}$ digits (for the input class $\Omega_K$). This agrees exactly with Theorem \ref{impossibility_theorem}.}
\end{table}

\subsection*{The subtlety and difficulty of removing instabilities, and the need for additional assumptions} Theorem \ref{impossibility_theorem} shows that the problems $(P_j)$ cannot, in general, be solved by any training algorithm. Hence any attempt at using the problems $(P_{j})$ as approximate solution maps of the \textit{general} inverse problem in \eqref{eq:inv_prob0}, without additional assumptions, is doomed to fail. This is not just the case for reconstruction using sparse regularisation, but also applies to other methods. In fact, any stable and accurate reconstruction procedure must be ``kernel aware'' \cite{gottschling2020troublesome}, a property that most DL methods do not enforce. A reconstruction method {$\Psi\colon \mathbb{C}^{m}\to \mathbb{C}^N$} lacks kernel awareness if it approximately recovers two vectors 
\begin{equation}\label{eq:recxx}\setlength\abovedisplayskip{6pt}\setlength\belowdisplayskip{6pt}
\|\Psi(Ax)-x\| \leq \epsilon 
\quad\text{and}\quad
\|\Psi(Ax')-x'\| \leq \epsilon 
\end{equation}
whose difference {$\|x-x'\| \gg 2\epsilon$ is large, but where the difference} lies close to the null space of $A$ (which is non-trivial due to $m<N$) so that $\| A(x-x') \| < \epsilon$. In particular, by applying \eqref{eq:recxx} and the triangle inequality twice, we have that
\begin{equation}\label{eq:instab1}\setlength\abovedisplayskip{6pt}\setlength\belowdisplayskip{6pt}
\|\Psi(Ax) - \Psi(Ax')\| \geq \|x-x'\| - 2\epsilon
\end{equation}
implying instability, as it only requires a perturbation $e=A(x'-x)$ of size $\|e\|<\epsilon$ for $\Psi(Ax+e) = \Psi(Ax')$ to reconstruct the wrong image. The issue here is that if we want to accurately recover $x$ and $x'$, i.e., we want \eqref{eq:recxx} to hold, then we cannot simultaneously have that $x-x'$ lies close to the kernel. Later we shall see conditions that circumvent this issue for our model class, thereby allowing us to compute stable and accurate NNs.

While training can encourage the conditions in \eqref{eq:recxx} to hold, it is not clear how many of the defense techniques in DL, simultaneously, will protect against the condition $\|A(x-x')\|<\epsilon$. One standard attempt to remedy instabilities is adversarial training \cite{raj2020improving}. However, while this strategy can potentially avoid \eqref{eq:recxx}, it may yield poor performance. For example, consider the following optimisation problem, which generates a reconstruction in the form of a NN given samples $\Theta=\{(y_s,x_s):s=1,...,R, Ax_s=y_s\}$ and $\epsilon,\lambda>0$:
\begin{equation}\setlength\abovedisplayskip{6pt}\setlength\belowdisplayskip{6pt}
\label{opt_GANS}
\min_{\phi\in\mathcal{N}_{m,N}}\! \sum_{s=1}^R\! \max_{\|e\|_{l^2}\leq\epsilon}\! \! \left\{\! \|x_s\! -\! \phi(y_s)\|_{l^2}^2\! +\! \lambda \|x_s\! -\! \phi(y_s\! +\! e)\|_{l^2}^2\! \right\}\! .\! 
\end{equation}
In other words, for each training point $(y,x)\in\Theta$ we find the worst-case perturbation $e$ in the $\epsilon$-ball around $y$. This is a simplified model of what one might do using Generative Adversarial Networks (GANs) to approximate adversarial perturbations \cite{goodfellow2014generative,arjovsky2017wasserstein}. For simplicity, assume that $A$ has full row rank $m$ and that we have access to exact measurements $y_s=Ax_s$. Suppose that our sample is such that $\min_{i\neq j}\|y_i-y_j\|_{l^2}>2\epsilon$. Any $\phi$ that minimises \eqref{opt_GANS} must be such that $\phi(y_s+e)=x_s$ for all $e$ with $\|e\|_{l^2}\leq \epsilon.$ Such networks can easily be constructed using, say, ReLU activation functions. Now suppose that $x_2$ is altered so that $x_1-x_2$ lies in the kernel of $A$. Then for any minimiser $\phi$, we must have $\phi(y_1+e)=\phi(y_2+e)=({x_1+x_2})/{2}$ for any $e$ with $\|e\|_{l^2}\leq \epsilon$, and hence we can never be more than $\|x_1+x_2\|_{l^2}/2$ accurate over the whole test sample. Similar arguments apply to other methods aimed at improving robustness such as adding noise to training samples (known as `jittering' - see Figure \ref{fig:false_neg}). Given such examples and Theorem \ref{impossibility_theorem}, we arrive at the following question:

\vspace{1mm}
 {\textbf{Question:} \it Are there sufficient conditions on $A$ that imply the existence of an algorithm that can compute a neural network that is both accurate and stable for the problem in \eqref{eq:inv_prob0}?}

\section*{Sufficient conditions for algorithms to compute stable and accurate NNs}

Sparse regularisation, such as the problems $(P_j)$, forms the core of many start-of-the-art reconstruction algorithms for inverse problems. We now demonstrate  a sufficient condition (from compressed sensing) guaranteeing the existence of algorithms for stable and accurate NNs. Sparsity in levels is a standard sparsity model for natural images \cite{adcock2017breaking,bastounis2017absence} as images are sparse in levels in X-lets (wavelets, curvelets, shearlets etc.)

\begin{definition}[Sparsity in levels]
\label{levels}
Let $\text{\upshape{\textbf{M}}}=(M_1,...,M_r) \in {\mathbb{N}^r}$, $1\leq M_1<...<M_r=N$, and $\text{\upshape{\textbf{s}}}=(s_1,...,s_r)\in {\mathbb{N}^{r}_{0}},$ where $s_l\leq M_l-M_{l-1}$ for $l=1,...,r$ ($M_0=0$). 
$x\in\mathbb{C}^N$ is $(\text{\upshape{\textbf{s}}},\text{\upshape{\textbf{M}}})$-sparse in levels if
$\left|\mathrm{supp}(x)\cap\{M_{l-1}+1,...,M_l\}\right|\leq s_l$ for $l=1,...,r.$ The total sparsity is $s=s_1+...+s_r$. We denote the set of $(\text{\upshape{\textbf{s}}},\text{\upshape{\textbf{M}}})$-sparse vectors by $\Sigma_{\text{\upshape{\textbf{s}}},\text{\upshape{\textbf{M}}}}$. We also define the following measure of distance of a vector $x$ to $\Sigma_{\text{\upshape{\textbf{s}}},\text{\upshape{\textbf{M}}}}$ by
$$\setlength\abovedisplayskip{6pt}\setlength\belowdisplayskip{6pt}
\sigma_{\text{\upshape{\textbf{s}}},\text{\upshape{\textbf{M}}}}(x)_{l^1_w}=\inf\{\|x-z\|_{l^1_w}:z\in\Sigma_{\text{\upshape{\textbf{s}}},\text{\upshape{\textbf{M}}}}\}.
$$
\end{definition}
This model has been used to explain the effectiveness of compressed sensing \cite{candes2006stable,donoho2006compressed,cohen2009compressed} in real life applications \cite{wang2014novel, jones2016continuous}. For simplicity, we assume that each $s_l > 0$ and that $w_i=w_{(l)}$ if $M_{l-1}+1\leq i\leq M_{l}$ (the weights in the $l^1_w$ norm are constant in each level). For a vector $c$ which is compressible in the wavelet basis, $\sigma_{\textbf{s},\textbf{M}}(x)_{l^1_w}$ is expected to be small if $x$ is the vector of wavelet coefficients of $c$ and the levels correspond to wavelet levels \cite{devore1998nonlinear}. In general, the weights are a prior on anticipated support of the vector \cite{friedlander2012recovering}, and we discuss some specific optimal choices in the SI Appendix.

For $\mathcal{I} \subset \{1,\ldots, N\}$, let $P_{\mathcal{I}} \in \mathbb{C}^{N\times N}$ denote the projection $(P_{\mathcal{I}}x)_i=x_i$ if $i \in \mathcal{I}$ and $(P_{\mathcal{I}}x)_i=0$ otherwise. The key ``kernel aware'' property that allows for stable and accurate recovery of $(\mathbf{s},\mathbf{M})$-spare vectors for the inverse problem \eqref{eq:inv_prob0}, is the \emph{weighted robust null space property in levels} (weighted rNSPL):
\begin{definition}[weighted rNSP in levels]\label{def_rNSPL}
Let $(\textup{\textbf{s}},\textup{\textbf{M}})$ be local sparsities and sparsity levels respectively. For weights $\{w_i\}_{i=1}^N$, $A\in\mathbb{C}^{m\times N}$ satisfies the weighted robust null space property in levels (weighted rNSPL) of order $(\textup{\textbf{s}},\textup{\textbf{M}})$ with constants $0<\rho<1$ and $\gamma>0$ if for any $(\textup{\textbf{s}},\textup{\textbf{M}})$ support set $\mathcal{I}\subset \{1,\ldots,N\}$,
$$\setlength\abovedisplayskip{6pt}\setlength\belowdisplayskip{6pt}
\|P_{\mathcal{I}} x\|_{l^2}\leq\frac{\rho}{\sqrt{\xi}}\|P_{\mathcal{I}^{c}}x\|_{l^1_w}+\gamma\|Ax\|_{l^2},
\quad\quad \text{for all $x\in\mathbb{C}^N$.}
$$
\end{definition}
We highlight that if $A$ satisfies the weighted rNSPL, then
$$\setlength\abovedisplayskip{6pt}\setlength\belowdisplayskip{6pt}
\| x-x' \|_{l^2} \leq C \|A(x-x')\|_{l^2}, \quad\forall x,x' \in \Sigma_{\mathbf{s},\mathbf{M}},
$$
where $C = C(\rho, \gamma) > 0$ is a constant depending only on $\rho$ and $\gamma$ (see SI Appendix). This ensures that if $\|x-x'\|_{\ell^2} \gg 2\epsilon$, then we cannot, simultaneously, have that $\|A(x-x')\| < \epsilon$, causing the instability in \eqref{eq:instab1}. Below, we give natural examples of sampling in compressed imaging where such a property holds, for known $\rho$ and $\gamma$, with large probability. We can now state a simplified version of our result (the full version with explicit constants is given and proven in the SI Appendix):

\begin{theorem}\label{firenet_informal}
There exists an algorithm such that for any input sparsity parameters $(\textup{\textbf{s}},\textup{\textbf{M}})$, weights $\{w_i\}_{i=1}^N$, $A \in \mathbb{C}^{m \times N}$ (with the input $A$ given by $\{A_l\}$) satisfying the rNSPL with constants $0<\rho<1$ and $\gamma>0$ (also input), and input parameters $n \in \mathbb{N}$ and $\{\delta,b_1,b_2\}\subset\mathbb{Q}_{>0}$, the algorithm outputs a neural network $\phi_n$ with $\mathcal{O}(n)$ hidden layers and $\mathcal{O}(N)$ width with following property. For any $x\in\mathbb{C}^N$, $y\in\mathbb{C}^m$ with
$$
\sigma_{\textup{\textbf{s}},\textup{\textbf{M}}}(x)_{l^1_w}+\|Ax-y\|_{l^2}\lesssim \delta,\quad \|x\|_{l^2}\lesssim b_1,\quad \|y\|_{l^2}\lesssim b_2,
$$
we have $\|\phi_{n}(y)-x\|_{l^2}\lesssim{\delta}+e^{-n}.$
\end{theorem}

Hence, up to the small error term $\sigma_{\text{\upshape{\textbf{s}}},\text{\upshape{\textbf{M}}}}(x)_{l^1_w}$, as $n\rightarrow\infty$ (with exponential convergence), we recover $x$ stably with an error proportional to the measurement error $\|Ax-y\|_{l^2}$. The explicit constant in front of the $\|Ax-y\|_{l^2}$ term can be thought of as an asymptotic \textit{local Lipschitz constant} for the NNs as $n\rightarrow\infty$, and thus measures \textit{stability of inexact input $y$}. The error of order $\sigma_{\textup{\textbf{s}},\textup{\textbf{M}}}(x)_{l^1_w}$ measures how close the vector $x$ is from the model class of sparse in levels vectors. In the full version of Theorem \ref{firenet_informal}, we also bound the error when we only approximately apply the nonlinear maps of the NNs, and show that these errors can only accumulate slowly as $n$ increases. In other words, we also gain a form of \textit{numerical stability} of the forward pass of the NN. The architectures of the NNs in Theorem \ref{firenet_informal} are based on the optimisation problem $(P_3)$ and unrolled primal-dual iterations. In addition to providing stability, the rNSPL allows us to prove exponential convergence through a careful restarting and reweighting scheme. We call our NNs {F}ast {I}terative {RE}started {NET}works (FIRENETs).  

\begin{remark}[{\bf Unrolling does not in general yield an algorithm producing an accurate network}] 
Unrolling iterative methods has a rich history in DL \cite{mccann2017convolutional,monga2021algorithm}. Note, however, that Theorem \ref{impossibility_theorem} demonstrates that despite the existence of an accurate neural network, there are scenarios where no algorithm exists that can compute it. Thus, unrolling optimisation methods can only work under certain assumptions. Our results are related to \cite{benlectures}, which shows how key assumptions such as the {robust nullspace property} help bound the error of the approximation to a minimiser in terms of error bounds on the approximation to the objective function.
\end{remark}

In the case that we do not know $\rho$ or $\gamma$ (the constants in the definition of rNSPL), we can perform a log-scale grid search for suitable parameters. By increasing the width of the NNs to $\mathcal{O}(N\log(n))$, we can still gain exponential convergence in $n$ by choosing the parameters in the grid search that lead to the vector with minimal $F_3^A$ (the objective function of $(P_3)$). In {other} cases, such as Theorem \ref{NN_FW_recov_thm} below, it is possible to prove probabilistic results where $\rho$ and $\gamma$ are known.

\subsection*{Examples in image recovery}

As an application, we consider Fourier and Walsh sampling, using Haar wavelets as a sparsifying transform. Our results can also be generalised to infinite-dimensional settings via higher-order Daubechies wavelets.

Let $K = 2^{r}$ for $r \in\mathbb{N}$, and set $N=K^d$ so that the objective is to recover a vectorised $d$-dimensional tensor $c\in\mathbb{C}^N$. Let $V \in \mathbb{C}^{N \times N}$ correspond to the $d-$dimensional discrete Fourier or Walsh transform (see SI Appendix). Let $\mathcal{I} \subset \{1,\ldots, N\}$ be a sampling pattern with cardinality $m=|\mathcal{I}|$ and let $D =\mathrm{diag}(d_1,\ldots,d_m) \in \mathbb{C}^{m\times m}$ be a suitable diagonal scaling matrix, whose entries along the diagonal depends only on $\mathcal{I}$. We assume we can observe the subsampled, scaled and noisy measurements $y= D P_{\mathcal{I}}V c+e \in \mathbb{C}^{m}$, where projection $P_{\mathcal{I}}$ is treated as a $m\times N$ matrix by ignoring the zero entries.

To recover a sparse representation of $c$, we consider Haar wavelet coefficients. Denote the discrete $d$-dimensional Haar Wavelet transform by $\Psi \in \mathbb{C}^{N \times N}$, and note that $\Psi^* = \Psi^{-1}$ since $\Psi$ is unitary. To recover the wavelet coefficients $x=\Psi c$ of $c$, we consider the matrix $A = DP_{\mathcal{I}} V\Psi^{*}$, and observe that $y=Ax +e= DP_{\mathcal{I}}Vc+e$. A key result in this work is that we can design a probabilistic sampling strategy (see SI Appendix), for both Fourier and Walsh sampling in $d$-dimensions, requiring no more than $m \gtrsim (s_1+\ldots + s_r) \cdot \mathcal{L}$ samples, which can ensure with high probability that $A$ satisfies the weighted rNSPL with certain constants. The sparsity in levels structure (Definition \ref{levels}) is chosen to correspond to the $r$ wavelet levels. Here $\mathcal{L}$ is a logarithmic term in $N,m,s$ and $\epsilon_{\mathbb{P}}^{-1}$ (where $\epsilon_{\mathbb{P}} \in (0,1)$ is a probability). This result is crucial, as it makes $A$ kernel aware for vectors that are approximately $(\mathbf{s},\mathbf{M})$-sparse, and allows us (using Theorem \ref{firenet_informal}) to design NNs which can stably and accurately recover approximately $(\mathbf{s},\mathbf{M})$-sparse vectors. Moreover, due to the exponential convergence in Theorem \ref{firenet_informal}, the depth of these NNs depends only logarithmically on the error $\delta$. Below follows a simplified version of our result (the full precise version is given and proven in the SI Appendix). 

%

\begin{theorem}
\label{NN_FW_recov_thm}
Consider the above setup of recovering wavelet coefficients $x=\Psi c$ of a tensor $c \in \mathbb{C}^{K^d}$ from subsampled, scaled and noisy Fourier or Walsh measurements $y = DP_{\mathcal{I}} V c+e$. Let $A=DP_{\mathcal{I}}V\Psi^{*}$, $m=|\mathcal{I}|$ and $\epsilon_{\mathbb{P}} \in (0,1)$. We then have
\begin{enumerate}[label=(\roman*)]
    \item If $\mathcal{I} \subset \{1,\ldots,N\}$ is a random sampling pattern drawn according to the strategy specified in SI Appendix, and
$$\setlength\abovedisplayskip{6pt}\setlength\belowdisplayskip{6pt}
m\gtrsim (s_1 + \cdots+s_r) \cdot \mathcal{L}.
$$
Then with probability $1-\epsilon_{\mathbb{P}}$, $A$ satisfies the weighted rNSPL of order $(\mathbf{s}, \mathbf{M})$ with constants $(\rho,\gamma)=(1/2,\sqrt{2})$. Here $\mathcal{L}$ denotes a term logarithmic in $\epsilon_{\mathbb{P}}^{-1},N,m$ and $s=\sum_{l=1}^rs_l$.
    \item Suppose $\mathcal{I}$ is chosen as above. For any $\delta\in(0,1)$, let $\mathcal{J}({\delta,\mathbf{s},\mathbf{M},w})$ be the set of all $y\!=\! Ax\!+\! e\!\in\mathbb{C}^m$ where
\begin{equation}\setlength\abovedisplayskip{6pt}\setlength\belowdisplayskip{6pt}
\label{bdd_for_delta}
\|x\|_{l^2}\leq 1,\quad \max\left\{{\sigma_{\text{\upshape{\textbf{s}}},\text{\upshape{\textbf{M}}}}(x)_{l^1_w}},\|e\|_{l^2}\right\}\leq \delta.
\end{equation}
We provide an algorithm that constructs a neural network $\phi$ with $\mathcal{O}(\log(\delta^{-1}))$ hidden layers (and width bounded by $2(N+m)$) such that with probability at least $1-\epsilon_{\mathbb{P}}$,
\begin{equation*}\setlength\abovedisplayskip{6pt}\setlength\belowdisplayskip{6pt}
\label{FW_recov_bound_J}
\left\|\phi(y)-c\right\|_{l^2}\lesssim \delta,\quad \forall y=Ax+e\in\mathcal{J}({\delta,\mathbf{s},\mathbf{M},w}).
\end{equation*}
\end{enumerate}
\end{theorem}

%
%
\section*{Balancing the stability and accuracy trade-off}
Current DL methods for image reconstruction can be unstable in the sense that (1) a tiny perturbation, in either the image or sampling domain, can cause severe artefacts in the reconstructed image (see Figure \ref{fig:automap_pert}, top row), and/or (2) a tiny detail in the image domain might be washed out in the reconstructed image (lack of accuracy), resulting in potential false negatives. Inevitably, there is a stability-accuracy trade-off, for this type of linear inverse problem, making it impossible for any reconstruction method to become arbitrarily stable without sacrificing accuracy or visa versa. Here, we show that the NNs computed by our algorithm (FIRENETs) are stable with respect to adversarial perturbations and accurate for images which are sparse in wavelets. As most images are sparse in wavelets, these networks also show great generalisation properties to unseen images.

\begin{figure*}
    \centering
    \setlength{\tabcolsep}{2pt} 
    \resizebox{1\textwidth}{!}{
    \begin{tabular}{@{}>{\centering}m{0.24\textwidth}>{\centering}m{0.24\textwidth}>{\centering}m{0.24\textwidth}>{\centering\arraybackslash}m{0.24\textwidth}@{}}   
    $\Psi(\tilde{y}), ~\tilde{y} = Ax+e_3$ & $\Phi\left(\tilde{y}, \Psi(\tilde{y}) \right)$ & FIRENET rec. from $y=Ax+\tilde e_3$ & AUTOMAP$+$FIRENET rec. from $y=Ax+\hat e_3$  \\
     \includegraphics[width=0.24\textwidth]{runner_19_r_idx_4_noisy_rec.png}
    &\includegraphics[width=0.24\textwidth]{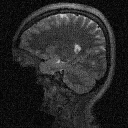} 
    &\includegraphics[width=0.24\textwidth]{im_rID_19_im_nbr_3_experi_001_pert_nbr_4_rec.png} 
    &\includegraphics[width=0.24\textwidth]{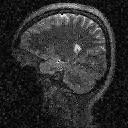}
    \end{tabular}
    }\vspace{-1mm}
    \caption{(\textbf{Adding a few FIRENET layers at the end of AUTOMAP makes it stable}). The FIRENET $\Phi\colon \mathbb{C}^m\times\mathbb{C}^N \to \mathbb{C}^N$ takes as input measurements $y \in \mathbb{C}^m$ and an initial guess for $x$, which we call $x_0 \in \mathbb{C}^N$. We now concatenate a 25-layer ($p=5$, $n=5$) FIRENET $\Phi$ and the AUTOMAP network $\Psi \colon \mathbb{C}^{m} \to \mathbb{C}^{N}$, by using the output from AUTOMAP as initial guess $x_0$, i.e., we consider the neural network mapping $y\mapsto \Phi(y,\Psi(y))$. In this experiment we consider the image $x$ from Figure \ref{fig:automap_pert} and the perturbed measurements $\tilde{y} = Ax+e_3$ (here $A$ is as in Figure \ref{fig:automap_pert}). The first column of the figure shows the reconstruction of AUTOMAP from Figure \ref{fig:automap_pert}. The second column shows the reconstruction of FIRENET with $x_0=\Psi(\tilde{y})$. The third column shows the reconstruction of FIRENET from Figure \ref{fig:automap_pert}. The fourth column shows the reconstruction of the concatenated network with a worst case perturbation $\hat e_3$ such that $\|\hat e_3\|_{l^2}\geq \|e_3\|_{l^2}$. In all other experiments we set $x_0=0$, and consider $\Phi$ as a mapping $\Phi \colon \mathbb{C}^m \to \mathbb{C}^N$.}
\label{fig:stabilisation}
\end{figure*}

\subsection*{Adversarial perturbations for AUTOMAP and FIRENETs} Figure \ref{fig:automap_pert} (top row) shows the stability test of \cite{antun2020instabilities} applied to the AUTOMAP \cite{Bo-18} network used for MRI reconstruction with 60\% subsampling. The image $x$ seen in Figure \ref{fig:automap_pert} is taken from the mentioned dataset, the stability test is run on the AUTOMAP network to find a sequence of perturbations $\left|e_1\right| < \left|e_2\right| < \left| e_3\right|$. As can be seen from the first row in the figure, the network reconstruction completely deforms the image and the reconstruction is severely unstable (similar results for other networks are demonstrated in \cite{antun2020instabilities}).

In contrast, we have applied the stability test, but now for the new NNs (FIRENETs) reported in this paper. Figure \ref{fig:automap_pert} (bottom row) shows the results for the constructed FIRENETs, where we rename the perturbations $\tilde e_j$ to emphasise the fact that these perturbations are sought for the new NNs and have nothing to do with the adversarial perturbations for AUTOMAP. We now see that despite the search for adversarial perturbations, the reconstruction remains stable. The error in the reconstruction was also found to be at most of the same order of the perturbation (as expected from the stability in Theorem \ref{firenet_informal}). In applying the test to FIRENETs, we tested/tuned the parameters in the gradient ascent algorithm considerably (much more so than was needed for applying the test to AUTOMAP, where finding instabilities was straightforward) in order to find the worst reconstruction results, yet the reconstruction remained stable. Note also that this is just one form of stability test and it is likely that there are many other tests for creating instabilities for NNs for inverse problems. This highlights the importance of results such as Theorem \ref{firenet_informal}, which guarantees stability regardless of the perturbation.

To demonstrate the generalisation properties of our NNs, we show the stability test applied to FIRENETs for a range of images in the SI Appendix. This shows stability across different types of images and highlights that conditions such as Definition \ref{def_rNSPL} allow great generalisation properties.

\subsection*{Stabilising unstable NNs with FIRENETs} Our NNs also act as a stabiliser. For example, Figure \ref{fig:stabilisation} shows the adversarial example for AUTOMAP (taken from Figure \ref{fig:automap_pert}), but now shows what happens when we take the reconstruction from AUTOMAP as an input to our FIRENETs. Here we use the fact that we can view our networks as approximations of unrolled and restarted iterative methods, allowing us to use the output of AUTOMAP as an additional input for the reconstruction. We see that FIRENETs fix the output of AUTOMAP and stabilise the reconstruction. Moreover, the concatenation itself of the networks remains stable to adversarial attacks.

\begin{figure*}
    \centering
    \setlength{\tabcolsep}{2pt} 
    \resizebox{1\textwidth}{!}{
	
    \begin{tabular}{@{}>{\centering}m{0.15\textwidth}>{\centering}m{0.27\textwidth}>{\centering}m{0.27\textwidth}>{\centering\arraybackslash}m{0.27\textwidth}@{}}   
     &Original $x$ & Original & Original + detail ($x+h_1$)\\
     &(full size)  & (cropped, red frame) & (cropped, blue frame) \\
    &\includegraphics[width=0.27\textwidth]{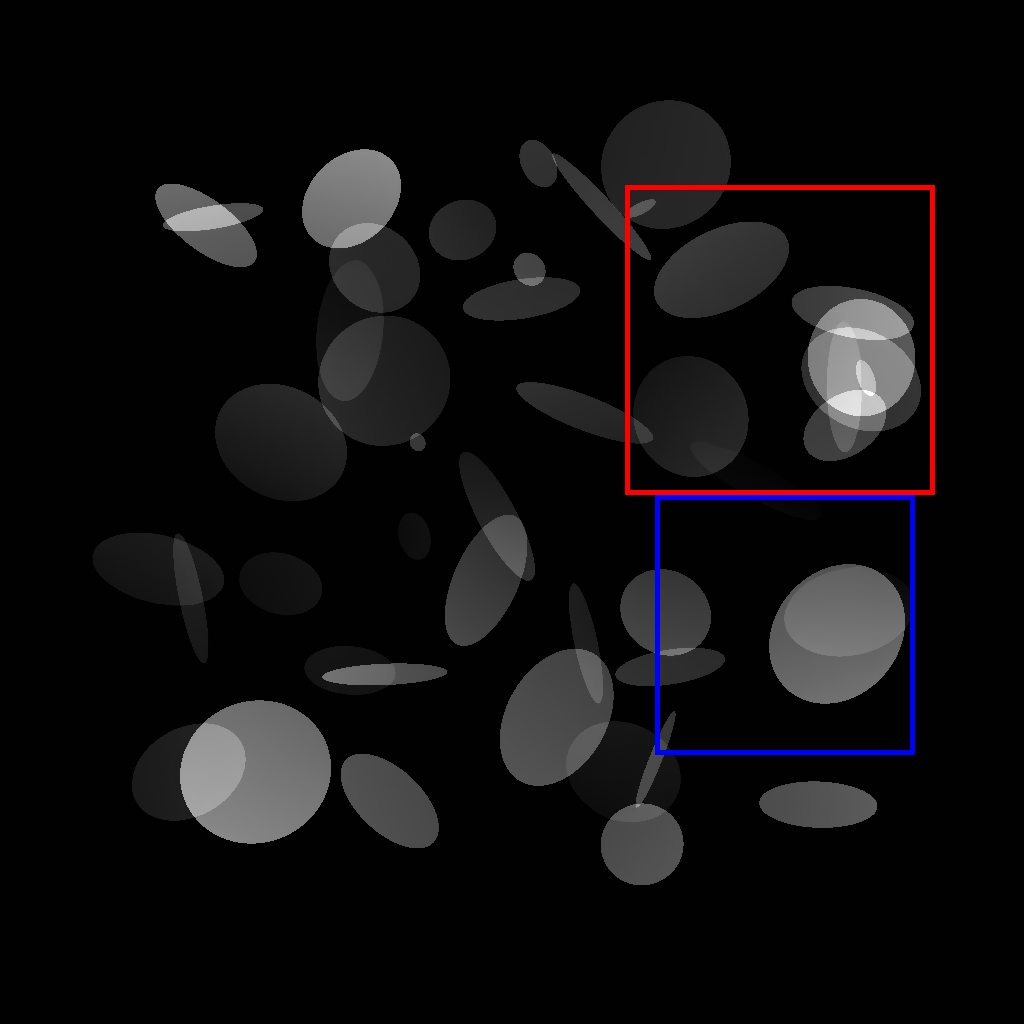}
    &\includegraphics[width=0.27\textwidth]{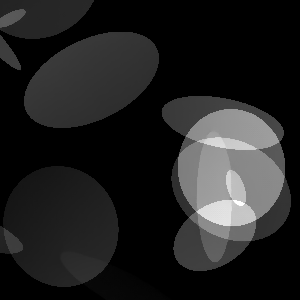}
    &\includegraphics[width=0.27\textwidth]{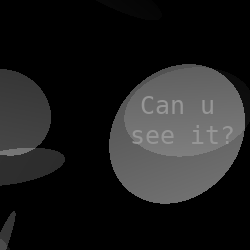}      
     \\
    \toprule 
    & Orig.\ + worst-case noise & Rec.\ from worst-case noise & Rec. of detail \\
    \begin{turn}{90}FIRENET ($\Phi_{1}$) $(n=5, p=5)$\end{turn}
    &\includegraphics[width=0.27\textwidth]{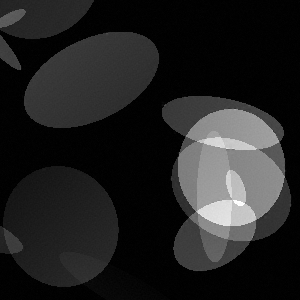}
    &\includegraphics[width=0.27\textwidth]{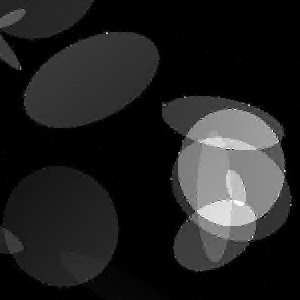}
    &\includegraphics[width=0.27\textwidth]{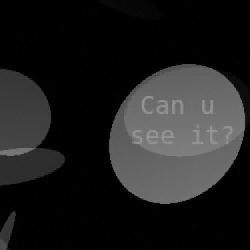}
    \\   
    \toprule
    & Orig.\ + worst-case noise & Rec.\ from worst-case noise & Rec. of detail \\
    \tikzmark[xshift=-19pt,yshift=25ex]{x}\begin{turn}{90}{\parbox{1.4in}{\centering Network $\Phi_{2}$ trained without noise}}\end{turn}
    & \includegraphics[width=0.27\textwidth]{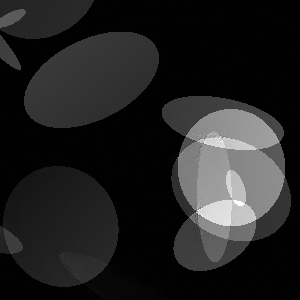}
    & \includegraphics[width=0.27\textwidth]{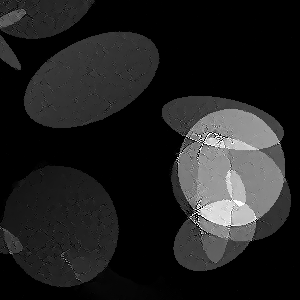}
    & \includegraphics[width=0.27\textwidth]{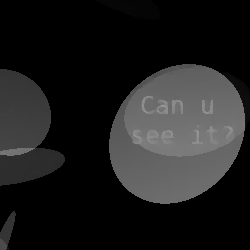}
     \\
    \tikzmark[xshift=-13pt,yshift=-1ex]{y}\begin{turn}{90}{\parbox{1.6in}{\centering Network $\Phi_{3}$ trained with measurements contaminated with random noise.}}\end{turn}
    & \includegraphics[width=0.27\textwidth]{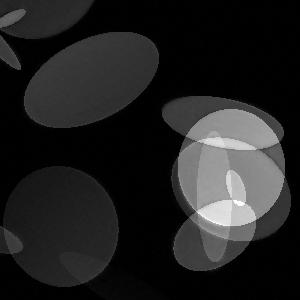}
    & \includegraphics[width=0.27\textwidth]{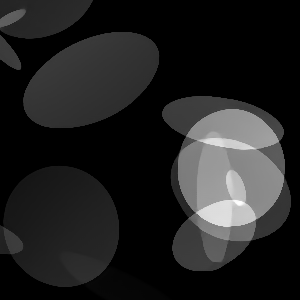}
    & \includegraphics[width=0.27\textwidth]{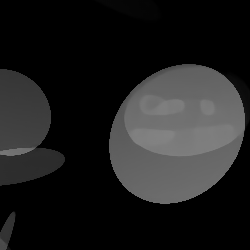}
     \\
 \end{tabular}
\drawcurvedarrow[-latex,ultra thick]{x}{y}
\annote[left]{arrow-1}{\begin{turn}{90}{\parbox{1.6in}{\centering \textbf{Increased stability, but Decreased accuracy}}}\end{turn}}
}
    \caption{\label{fig:automap_pert4} (\textbf{Trained neural networks with limited performance can be stable}). We examine the accuracy/stability trade-off for linear inverse problems by considering three reconstruction networks $\Phi_{j}\colon \mathbb{C}^{m} \to \mathbb{C}^{N}$, $j=1,2,3$. Here $\Phi_1$ is a FIRENET, whereas $\Phi_{2}$ and $\Phi_{3}$ are the U-nets mentioned in the main text, trained without and with noisy measurements, respectively. For each network, we compute a perturbation $w_j \in \mathbb{C}^{N}$ meant to simulate the worst-case effect, and we show a cropped version of the perturbed images $x+w_j$ in the first column (row 2-4). In the middle column (row 2-4), we show the reconstructed images $\Phi_{j}(A(x+w_j))$ from each of the networks. In the last column (row 2-4) we test the networks' ability to reconstruct a tiny detail $h_1$, in the form of the text \enquote{can u see it?}. As we see from the figure, the network trained on noisy measurements is stable to worst-case perturbations, but it is not accurate. Conversely, the network trained without noise is accurate but not stable. The FIRENET is balancing this trade-off and is accurate for images which are sparse in wavelets, and stable to worst-case perturbations.}\label{fig:false_neg}

\end{figure*}

\subsection*{The stability vs. accuracy trade-off and false negatives} It is easy to produce a perfectly stable network: the zero network is the obvious candidate! However, this network would obviously have poor performance and produce many false negatives. The challenge is to simultaneously ensure performance and stability. Figure \ref{fig:false_neg} highlights this issue. Here we have trained two NNs to recover a set of ellipses images from noise-free and noisy Fourier measurements. The noise-free measurements are generated as $y=Ax$, where $A \in \mathbb{C}^{m\times N}$ is a subsampled discrete Fourier transform, with $m/N = 0.15$ and $N=1024^2$. The noisy measurements are generated as $y=Ax+ce$, where $A$ is as before, and the real and imaginary components of $e \in \mathbb{C}^{m}$ are drawn from a zero mean and unit variance normal distribution $\mathcal{N}(0,1)$, and $c \in \mathbb{R}$ is drawn from the uniform distribution $\mathrm{Unif([0, 100])}$. The noise $ce \in \mathbb{C}^m$, is generated on the fly during the training process.  

The trained networks use a standard benchmarking architecture for image reconstruction, and maps $y \mapsto \phi(A^* y)$, where $\phi\colon \mathbb{C}^{N} \to \mathbb{R}^N$ is a trainable U-net NN \cite{long2015fully, jin17}. Training networks with noisy measurements, using for example this architecture, have previously been used as an example of how to create NNs which are robust towards adversarial attacks \cite{genzel2020solving}. As we can see from Figure \ref{fig:automap_pert4} (bottom row) this is the case, as it does indeed create a NN which is stable with respect to worst-case perturbations.  However, a key issue is that it is also producing false negatives due to its inability to reconstruct details. Similarly, as reported in the 2019 FastMRI challenge, trained NNs that performed well in terms of standard image quality metrics were prone to false negatives: they failed to reconstruct small, but physically-relevant image abnormalities \cite{knoll2020advancing}. Pathologies, generalisation and AI-generated hallucinations were subsequently a focus of the 2020 challenge \cite{muckley2020state}. FIRENET, on the other hand, has a guaranteed performance (on images being sparse in wavelet bases) and stability, given specific conditions on the sampling procedure. The challenge is to determine the optimal balance between accuracy/stability, a problem that is well known in numerical analysis.

\section*{Concluding remarks}

\begin{itemize}[leftmargin=13pt]
\setlength\itemsep{0pt}%
    \setlength\parskip{4pt}%
\item[(i)] ({\bf Algorithms may not exist -- Smale's 18th problem}). There are well-conditioned problems where accurate NNs exist, but no algorithm can compute them. Understanding this phenomenon is essential in order to address Smale's 18th problem on the limits of AI. Moreover, the limitations established in this paper suggest a classification theory describing the conditions needed for the existence of algorithms that can compute stable and accurate NNs. 

\item[(ii)] ({\bf Classifications and Hilbert's program})
The strong optimism regarding the abilities of AI is comparable to the optimism surrounding mathematics in the early 20th century, led by D. Hilbert. Hilbert believed that mathematics could prove or disprove any statement and, moreover, that there were no restrictions on which problems could be solved by algorithms. G\"{o}del \cite{godel1931formal} and Turing \cite{turing1937computable} turned Hilbert's optimism upside down by their foundational contributions establishing impossibility results on what mathematics and digital computers can achieve.

Hilbert's program on the foundations of mathematics led to a rich mathematical theory and modern logic and computer science, where substantial efforts were made to classify which problems can be computed. We have sketched a similar program for modern AI, where we provide certain sufficient conditions for the existence of algorithms to produce stable and accurate NNs. We believe that such a program on the foundations of AI is necessary and will act as an invaluable catalyst for the advancement of AI.

\item[(iii)] ({\bf Trade-off between stability and accuracy})
For ML in inverse problems there is an intrinsic trade-off between stability and accuracy. We demonstrated NNs that offer a blend of both stability and accuracy. Balancing these two interests is crucial for applications and will no doubt require a myriad of future techniques to be developed. Tracing out the optimal stability vs. accuracy trade-off remains largely an open problem, one that we expect to be of particular relevance in the increasing number of real-world implementations of ML in inverse problems.

\item[(iv)] ({\bf Future work})
Our results are just the beginning of a potential vast mathematical theory on which NNs can be computed by algorithms. Note that our sufficient conditions are just for a specific class of images. Thus, even for inverse problems, this opens up for a substantial theory covering other sufficient (and potentially necessary) conditions guaranteeing stability and accuracy, and extensions to other inverse problems such as phase retrieval \cite{candes2013phaselift,fannjiang2020numerics}.
\end{itemize}

\vspace{-4mm}

\section*{Methods and Data Availability}
Full theoretical derivations are given in the SI Appendix. Our proof techniques for fundamental barriers in Theorem \ref{impossibility_theorem} stem from the SCI hierarchy that has recently been used to settle longstanding questions in scientific computing \cite{Hansen_JAMS,colbrook_spec_meas,opt_big}. We introduce the concept of Sequential General Algorithms which also captures the notion of adaptive and/or probabilistic choice of training data (thus strengthening the proven lower bounds). All the code and data used to produce the figures in this manuscript are available from \url{www.github.com/Comp-Foundations-and-Barriers-of-AI/firenet}.



\section*{Acknowledgments} 
We are grateful to Kristian Haug for allowing us to use parts of unrolled primal-dual algorithms developed in TensorFlow. 
This work was supported by a Research Fellowship at Trinity College, Cambridge (M.J.C.), and a Leverhulme Prize and a Royal Society University Research Fellowship (A.C.H.).

\bibliographystyle{abbrv}
\bibliography{NNBib}



\section*{Supplementary material}

# CAN STABLE AND ACCURATE NEURAL NETWORKS BE COMPUTED? – ON THE BARRIERS OF DEEP LEARNING AND SMALE'S 18TH PROBLEM
# SI APPENDIX

MATTHEW J. COLBROOK*†, VEGARD ANTUN*‡, AND ANDERS C. HANSEN†‡

Here we provide statements of theorems from the main text, proofs of theorems, detailed explanations of the experimental setup and further numerical examples. We briefly collect some basic notation, and further notation will be introduced throughout where appropriate. We use $\mathcal{N}_{m,N}$ to denote the class of neural networks (NNs) from $\mathbb{C}^m$ to $\mathbb{C}^N$ (see §1.1.2.1 for the precise definition). Given a metric space $(\mathcal{M}, d)$, $x \in \mathcal{M}$ and $X \subset \mathcal{M}$, $d(x, X) = \text{dist}(x, X) = \inf_{y \in X} d(x, y)$. For a matrix $A \in \mathbb{C}^{m \times N}$, the norm $\|A\|$ refers to the operator norm of $A$ when $\mathbb{C}^m$ and $\mathbb{C}^N$ are equipped with the standard $l^2$-norm. For $x \in \mathbb{C}^N$ and $p \in [1, \infty]$, $\|x\|_{l^p}$ refers to the $l^p$-norm of $x$. For a set of indices $S$ and vector $x$, $x_S$ is the vector defined by $(x_S)_j = x_j$ if $j \in S$ and $(x_S)_j = 0$ if $j \notin S$. Complex rationals $\mathbb{Q} + i\mathbb{Q}$ are denoted by $\mathbb{Q}[i]$. We use □ to denote the end of a proof and ⊠ to denote the end of a remark.

## Contents

*M.J.C. AND V.A. CONTRIBUTED EQUALLY TO THIS WORK.
†DEPARTMENT OF APPLIED MATHEMATICS AND THEORETICAL PHYSICS, UNIVERSITY OF CAMBRIDGE.
‡DEPARTMENT OF MATHEMATICS, UNIVERSITY OF OSLO.

1. STATEMENT OF THEOREMS AND RESULTS

We now state our main theorems. Proofs are given in §3 – §5. Recall that we study the canonical inverse problem of solving an underdetermined system of linear equations:

$$\text{Given noisy measurements } y = Ax + e \in \mathbb{C}^m \text{ of } x \in \mathbb{C}^N, \text{ recover } x. \tag{1.1}$$

Here $A \in \mathbb{C}^{m \times N}$ represents a model of typically undersampled sampling ($m < N$), such as a subsampled discrete Fourier transform as in Magnetic Resonance Imaging (MRI). Specific choices of $A$ are discussed in §1.1.3. Problem (1.1) forms the basis for much of inverse problems and image analysis. The possibility of $y \neq Ax$ models noise or perturbations.

1.1. **Existence of NNs is not enough, algorithms may not compute them sufficiently accurately.** Here we present Theorems 1 and 2 of the main paper. Given a matrix $A \in \mathbb{C}^{m \times N}$ and a vector $y \in \mathbb{C}^m$, recall that we consider the following three minimisation problems:

$$(P_1) \quad \operatorname{argmin}_{x \in \mathbb{C}^N} F_1^A(x) := \|x\|_{l_w^1}, \text{ such that } \|Ax - y\|_{l^2} \leq \epsilon, \tag{1.2}$$

$$(P_2) \quad \operatorname{argmin}_{x \in \mathbb{C}^N} F_2^A(x, y, \lambda) := \lambda \|x\|_{l_w^1} + \|Ax - y\|_{l^2}^2, \tag{1.3}$$

$$(P_3) \quad \operatorname{argmin}_{x \in \mathbb{C}^N} F_3^A(x, y, \lambda) := \lambda \|x\|_{l_w^1} + \|Ax - y\|_{l^2}. \tag{1.4}$$

The parameters $\lambda$ and $\epsilon$ are positive rational numbers, and the weighted $l_w^1$ norm is given by $\|x\|_{l_w^1} := \sum_{l=1}^N w_l |x_l|$, where each weight $w_j$ is a positive rational. Throughout, we use the following notation:

$$\Xi(A, y) \text{ is the set of minimisers for } (P_j) \text{ given input } A \in \mathbb{C}^{m \times N}, y \in \mathbb{C}^m, \tag{1.5}$$

where, for notational convenience, we have suppressed the dependence on $\epsilon$ or $\lambda$ (which are usually fixed parameters) and the index $j$. In certain cases, we will write $\Xi_j$ to specify minimisers of problem $(P_j)$. Let

$$A \in \mathbb{C}^{m \times N}, \quad \mathcal{S} = \{y_k\}_{k=1}^R \subset \mathbb{C}^m, \quad R < \infty.$$

In the main text we considered the following key question:

*Given a collection $\Omega$ of such pairs $(A, \mathcal{S})$, does there exists a neural network approximating the mapping $\Xi$, and if so, can such an approximation be trained by an algorithm?*

To make this question precise, we first note that $A$ and the elements in $\mathcal{S}$ will typically never be exact, but can be approximated to arbitrary precision. For example, this would be the case if $A$ was a subsampled discrete cosine transform. Thus, we can access approximations $\{y_{k,n}\}_{k=1}^R \subset \mathbb{Q}[i]^m$ and $A_n \in \mathbb{Q}[i]^{m \times N}$ such that

$$\|y_{k,n} - y_k\| \leq 2^{-n}, \quad \|A_n - A\| \leq 2^{-n}, \quad \forall n \in \mathbb{N}. \tag{1.6}$$

We also assume access to $\{x_{k,n}\}_{k=1}^R \subset \mathbb{Q}[i]^N$ such that

$$\inf_{x^* \in \Xi(A_n, y_{k,n})} \|x_{k,n} - x^*\| \leq 2^{-n}, \quad \forall n \in \mathbb{N}. \tag{1.7}$$

Hence, the training set associated with $(A, \mathcal{S}) \in \Omega$ for training a suitable NN must be

$$\iota_{A,\mathcal{S}} := \{(y_{k,n}, A_n, x_{k,n}) \mid k = 1, \ldots, R, \text{ and } n \in \mathbb{N}\}. \tag{1.8}$$

Thus, given a collection of $(A, \mathcal{S})$, we denote the class of all such admissible training data by

$$\Omega_{\mathcal{T}} := \{\iota_{A,\mathcal{S}} \text{ as in (1.8)} \mid (A, \mathcal{S}) \in \Omega, \text{ (1.6) and (1.7) hold}\}.$$

Precise statements addressing the above question are summarised in the following theorems, the first of which follows directly from standard universal approximation theorems.

**Theorem 1** (**Neural networks exist for $\Xi$**). *Consider the problem $(P_j)$ ($j = 1, 2, 3$) for fixed dimensions $m < N$ and parameters $\lambda$ or $\epsilon$. Then, for any family $\Omega$ of such $(A, \mathcal{S})$ described above, there exists a mapping*

$$\mathcal{K} : \Omega_{\mathcal{T}} \to \mathcal{N}_{m,N}, \quad \mathcal{K}(\iota_{A,\mathcal{S}}) = \varphi_{A,\mathcal{S}}, \quad \text{such that} \quad \varphi_{A,\mathcal{S}}(y) \in \Xi(A, y), \quad \forall y \in \mathcal{S}.$$

*In words, $\mathcal{K}$ maps the training data $\Omega_{\mathcal{T}}$ to NNs that solve the optimisation problem $(P_j)$ for each $(A, \mathcal{S}) \in \Omega$.*

Despite the existence of NNs guaranteed by Theorem 1, the problem of computing such a NN from training data is a most delicate issue, as described in the following theorem (proven in §3).

**Theorem 2** (**Despite existence, neural networks may only be computed to a certain accuracy**)**.** *For $j = 1, 2$ or $3$, consider the optimisation problem $(P_j)$ for fixed parameters $\lambda \in (0, 1]$ or $\epsilon \in (0, 1/2]$ and $w_l = 1$, where $N \geq 2$ and $m < N$. Let $K > 2$ be a positive integer and let $L \in \mathbb{N}$. Then there exists a class $\Omega$ of elements $(A, \mathcal{S})$ as in (1.5), with the following properties. The class $\Omega$ is well-conditioned with condition numbers of the matrices $AA^*$ and the solution maps $\Xi$, as well as the feasibility primal local condition number (see §3.3.1), all bounded by 1 independent of all parameters. However, the following hold:*

(i) *There does not exist any algorithm that, given a training set $\iota_{A,\mathcal{S}} \in \Omega_\mathcal{T}$, produces a NN $\phi_{A,\mathcal{S}}$ with*

$$\min_{y \in \mathcal{S}} \inf_{x^* \in \Xi(A,y)} \|\phi_{A,\mathcal{S}}(y) - x^*\|_{l^2} \leq 10^{-K}, \quad \forall (A, \mathcal{S}) \in \Omega. \tag{1.9}$$

*Furthermore, for any $p > 1/2$, no probabilistic algorithm (BSS, Turing or any model of computation) can produce a NN $\phi_{A,\mathcal{S}}$ such that (1.9) holds with probability at least $p$.*

(ii) *There does exist a deterministic Turing machine that, given a training set $\iota_{A,\mathcal{S}} \in \Omega_\mathcal{T}$, produces a NN $\phi_{A,\mathcal{S}}$ with*

$$\max_{y \in \mathcal{S}} \inf_{x^* \in \Xi(A,y)} \|\phi_{A,\mathcal{S}}(y) - x^*\|_{l^2} \leq 10^{-(K-1)}, \quad \forall (A, \mathcal{S}) \in \Omega. \tag{1.10}$$

*However, for any probabilistic Turing machine $(\Gamma, \mathbb{P})$, $M \in \mathbb{N}$ and $p \in \left[0, \frac{N-m}{N+1-m}\right)$ that produces a NN $\phi_{A,\mathcal{S}}$, there exists a training set $\iota_{A,\mathcal{S}} \in \Omega_\mathcal{T}$ such that for all $y \in \mathcal{S}$,*

$$\mathbb{P}\Big(\inf_{x^* \in \Xi(A,y)} \|\phi_{A,\mathcal{S}}(y) - x^*\|_{l^2} > 10^{1-K} \text{ or the training data size needed to construct } \phi_{A,\mathcal{S}} > M\Big) > p. \tag{1.11}$$

(iii) *There does exist a deterministic Turing machine that, given a training set $\iota_{A,\mathcal{S}} \in \Omega_\mathcal{T}$ and using only $L$ training data from each $\iota_{A,\mathcal{S}}$, produces a NN $\phi_{A,\mathcal{S}}(y)$ such that*

$$\max_{y \in \mathcal{S}} \inf_{x^* \in \Xi(A,y)} \|\phi_{A,\mathcal{S}}(y) - x^*\|_{l^2} \leq 10^{-(K-2)}, \quad \forall (A, \mathcal{S}) \in \Omega. \tag{1.12}$$

**Remark 1.1** (Meaning of the notation $(\Gamma, \mathbb{P})$)**.** The notation $(\Gamma, \mathbb{P})$ in (ii) is used to denote a (possibly) randomised algorithm $\Gamma$ and its law $\mathbb{P}$. This includes scenarios such as stochastic gradient descent, random selection of training data, random computation with training data etc. The precise setup is detailed in §3.3.1.⊠

**Remark 1.2** (Generalisations of Theorem 2)**.** For simplicity, we have stated Theorem 2 for errors measured in the $l^2$-norm and the case of unweighted $l^1$ regularisation (all the $w_l = 1$) in the problems $(P_j)$. However, the proof can be adapted, and similar results hold for any norm replacing the $l^2$-norm, and any non-singular weighted $l^1$ regularisation. Moreover, result (i) in Theorem 2 holds regardless of the model of computation, even if we allowed real number arithmetic (see Definition 3.3). For further details on the precise setup, including the definition of condition numbers, which are standard in the literature, see §3.3.1. Finally, the theorem remains true if we restrict ourselves to real-valued matrices and vectors. ⊠

Further details on the experiment following this theorem (that was given in the main text) can be found in §3.3.4.

1.2. **Computing stable and accurate neural networks.**

1.2.1. *Neural networks and notational conventions.* To state our theorems, we need to be precise about the definition of a NN. For introductions to the field of DL and NNs, we refer the reader to [20, 18] and [21], respectively, and the references therein. To capture standard architectures used in practice such as skip connections, we consider the following definition of a NN. Without loss of generality and for ease of exposition, we also work with complex-valued NNs. Such networks can be realised by real-valued NNs by splitting into real and imaginary parts. A NN is a mapping $\phi \colon \mathbb{C}^m \to \mathbb{C}^N$ that can be written as a composition

$$\phi(y) = V_T(\rho_{T-1}(...\rho_1(V_1(y)))), \quad \text{where:}$$

- Each $V_j$ is an affine map $\mathbb{C}^{N_{j-1}} \to \mathbb{C}^{N_j}$ given by $V_j(x) = W_j x + b_j(y)$ where $W_j \in \mathbb{C}^{N_j \times N_{j-1}}$ and the $b_j(y) = R_j y + c_j \in \mathbb{C}^{N_j}$ are affine functions of the input $y$.
- Each $\rho_j : \mathbb{C}^{N_j} \to \mathbb{C}^{N_j}$ is one of two forms:
  (i) There exists an index set $I_j \subset \{1, ..., N_j\}$ (possibly a strict subset) such that $\rho_j$ applies a possibly non-linear function $f_j : \mathbb{C} \to \mathbb{C}$ element-wise on the input vector's components with indices in $I_j$:
  $$\rho_j(x)_k = \begin{cases} f_j(x_k), & \text{if } k \in I_j \\ x_k, & \text{otherwise.} \end{cases}$$
  (ii) There exists a possibly non-linear function $f_j : \mathbb{C} \to \mathbb{C}$ such that, after decomposing the input vector $x$ as $(x_0, X^\top, Y^\top)^\top$ ($\top$ denotes transpose) for scalar $x_0$ and $X \in \mathbb{C}^{m_j}$ ($Y \in \mathbb{C}^{N_j - 1 - m_j}$), we have
  $$\rho_j : \begin{pmatrix} x_0 \\ X \\ Y \end{pmatrix} \to \begin{pmatrix} 0 \\ f_j(x_0) X \\ Y \end{pmatrix}. \tag{1.13}$$

The affine dependence of $b_j(y)$ on $y$ allows skip connections from the input to the current level as in standard definitions of feed-forward NNs [23, p. 269], and the above architecture has become standard [19, 17, 15].

**Remark 1.3** (On the use of multiplication). The use of non-linear functions of the form (ii) may be re-expressed using the following element-wise squaring trick:

$$\begin{pmatrix} x_0 \\ X \\ Y \end{pmatrix} \to \begin{pmatrix} f_j(x_0) \\ X \\ Y \end{pmatrix} \to \begin{pmatrix} f_j(x_0)\mathbf{1} \\ X \\ f_j(x_0)\mathbf{1} + X \\ Y \end{pmatrix} \to \begin{pmatrix} f_j(x_0)^2 \mathbf{1} \\ X^2 \\ [f_j(x_0)\mathbf{1} + X]^2 \\ Y \end{pmatrix} \to \begin{pmatrix} 0 \\ \frac{1}{2}\left[[f_j(x_0)\mathbf{1} + X]^2 - f_j(x_0)^2 \mathbf{1} - X^2\right] \\ Y \end{pmatrix},$$

where $\mathbf{1}$ denotes a vector of ones of the same size as $X$ (so that $f_j(x_0) \to f_j(x_0)\mathbf{1}$ is a linear map). However, this is not done in practice since the map in (1.13) is directly trainable via backpropagation. $\boxtimes$

Note that we do not allow the matrices $W_j$ to depend on $y$. We denote the collection of all NNs of the above form by $\mathcal{N}_{\mathbf{D}, T, q}$, where the vector $\mathbf{D} = (N_0 = m, N_1, ..., N_T = N)$ denotes the dimensions in each layer, $T$ denotes the number of layers and $q$ denotes the number of different non-linear functions applied (including the count of different $I_j$ and $m_j$). In general, we will require that the layer sizes $N_j$ do not grow with $j$ so that the size of each layer is of the same order as the sampling matrix $A$.

We consider stable reconstruction from noisy undersampled measurements, as in (1.1), and NNs that can be constructed via algorithms. To make this precise, we assume that we have access to a sequence of matrices $A_l \in \mathbb{Q}[i]$ such that $\|A - A_l\| \leq q_l$ for some known null sequence $\{q_l\}$. This is consistent with the training set given by (1.8). To construct NNs via an algorithm, care must be taken with the non-linear activation functions. We assume that for $\theta \in \mathbb{Q}_{>0}$ we have access to a routine "$\mathrm{sqrt}_\theta$" such that $|\mathrm{sqrt}_\theta(x) - \sqrt{x}| \leq \theta$ for all $x \in \mathbb{R}_{\geq 0}$. In what follows, the non-linear maps $f_j$ used in the NNs are either arithmetic or constructed using arithmetic operations and $\mathrm{sqrt}_\theta$. We always ensure that $\mathrm{sqrt}_\theta$ acts on non-negative real numbers and on rational inputs if the input to the NN is rational. We refer to the pair $(\phi, \theta)$ as a NN.

**Remark 1.4** (Approximating $\sqrt{\cdot}$ with neural networks). On any bounded set (for bounded input our constructed NNs only require the routine $\mathrm{sqrt}_\theta$ on a bounded set), we can construct an approximation to $\sqrt{\cdot}$ using standard non-linear activation functions such as ReLU (more efficient approximations may be achieved by using other activation functions such as rational maps [11]). The choice of the square root function is somewhat arbitrary, but simplifies our proof of Theorem 3. Similar results hold for other activation functions. $\boxtimes$

**Remark 1.5** (An interpretation of $\theta$). As well as being necessary from a foundations point of view, an important interpretation of $\theta$ is numerical stability, or accumulation of errors, of the forward pass of the NN. Larger values of $\theta$ show greater stability when applying the NN in finite precision. We prove results of the form

$$\|\phi_n(y) - x\|_{l^2} \leq \varepsilon + c_1(A, x)\|Ax - y\|_{l^2} + c_2(A, x) v^n, \quad \forall x \in S \subset \mathbb{C}^N, y \in \mathbb{C}^m, \tag{1.14}$$

where $(\phi_n, \theta_n)$ is a (sequence of) NN(s) with $\mathcal{O}(n)$ layers that is computed by an algorithm, $v \in (0,1)$ describes the exponential rate of convergence in the number of layers, $\varepsilon > 0$ (in our results $\varepsilon$ will be related to the distance to vectors that are sparse in levels: $\sigma_{\mathbf{s},\mathbf{M}}(x)_{l_w^1}$ in Definition 1.6), and $\theta_n^{-1} = \theta^{-1}$ is bounded independent of $n$. Up to the error tolerance $\varepsilon$, the constant $c_1(A,x)$ can be thought of as an asymptotic *local Lipschitz constant* for the NNs as $n \to \infty$, and thus measures *stability of inexact input $y$*. In practice one would use floating point arithmetic to approximate square roots. Hence, the boundedness of $\theta_n^{-1}$ is a *numerical notion of stability* - the accuracy needed for approximating square roots (and the non-linear maps) does not become too great and errors do not accumulate as $n$ increases. Moreover, in practice the value of $\theta^{-1}$ needed is well below what is achieved using standard floating-point formats. ⊠

1.2.2. *The construction of stable and accurate neural networks.* The main result of this subsection, Theorem 3, uses the concept of sparsity in levels and weighted robust null space property in levels defined in the main text. We repeat these definitions here for convenience of the reader.

**Definition 1.6** (Sparsity in levels). *Let* $\mathbf{M} = (M_1, ..., M_r) \in \mathbb{N}^r$, $1 \leq M_1 < ... < M_r = N$, *and* $\mathbf{s} = (s_1, ..., s_r) \in \mathbb{N}_0^r$, *where* $s_k \leq M_k - M_{k-1}$ *for* $k = 1, ..., r$ $(M_0 = 0)$. $x \in \mathbb{C}^N$ *is* $(\mathbf{s}, \mathbf{M})$-*sparse in levels if*

$$|\mathrm{supp}(x) \cap \{M_{k-1}+1, ..., M_k\}| \leq s_k, \quad k = 1, ..., r.$$

*The total sparsity is $s = s_1 + ... + s_r$. We denote the set of $(\mathbf{s}, \mathbf{M})$-sparse vectors by $\Sigma_{\mathbf{s},\mathbf{M}}$. We also define the following measure of distance of a vector $x$ to $\Sigma_{\mathbf{s},\mathbf{M}}$ by*

$$\sigma_{\mathbf{s},\mathbf{M}}(x)_{l_w^1} = \inf\{\|x - z\|_{l_w^1} : z \in \Sigma_{\mathbf{s},\mathbf{M}}\}.$$

For simplicity, we assume throughout that each $s_k > 0$ and that

$$w_i = w_{(j)}, \quad \text{if } M_{j-1} + 1 \leq i \leq M_j. \tag{1.15}$$

**Definition 1.7** (weighted rNSP in levels). *Let $(\mathbf{s}, \mathbf{M})$ be local sparsities and sparsity levels respectively. For weights $\{w_i\}_{i=1}^N$ ($w_i > 0$), we say that $A \in \mathbb{C}^{m \times N}$ satisfies the weighted robust null space property in levels (weighted rNSPL) of order $(\mathbf{s}, \mathbf{M})$ with constants $0 < \rho < 1$ and $\gamma > 0$ if for any $(\mathbf{s}, \mathbf{M})$ support set $\Delta$,*

$$\|x_\Delta\|_{l^2} \leq \rho \|x_{\Delta^c}\|_{l_w^1}/\sqrt{\xi} + \gamma \|Ax\|_{l^2}, \quad \text{for all } x \in \mathbb{C}^N.$$

We also define the following quantities:

$$\xi = \xi(\mathbf{s},\mathbf{M},w) := \sum_{k=1}^r w_{(k)}^2 s_k, \quad \zeta = \zeta(\mathbf{s},\mathbf{M},w) := \min_{k=1,...,r} w_{(k)}^2 s_k, \quad \kappa = \kappa(\mathbf{s},\mathbf{M},w) := \xi/\zeta.$$

Unless there is ambiguity, we will drop the $(\mathbf{s}, \mathbf{M}, w)$ from the notation of these parameters. Recall the setup throughout this paper of a matrix $A \in \mathbb{C}^{m \times N}$ ($m < N$), where we have access to an approximation sequence $A_l$ such that $\|A - A_l\| \leq q_l$ with known $q_l \to 0$ as $l \to \infty$. In this regard, the following simple perturbation lemma is useful (whose proof is given in §4).

**Lemma 1.8** (**The weighted rNSP in levels is preserved under perturbations or approximations**). *Assume that (1.15) holds and that $A$ satisfies the weighted rNSPL of order $(\mathbf{s}, \mathbf{M})$ with constants $0 < \rho < 1$ and $\gamma > 0$. Let $\hat{A}$ be an approximation of $A$ such that $\|\hat{A} - A\| < (1-\rho)\gamma^{-1}\left(1 + \frac{\sqrt{\xi}}{\min_{k=1,...,r} w_{(k)}}\right)^{-1}$. Then $\hat{A}$ satisfies the weighted rNSPL of order $(\mathbf{s}, \mathbf{M})$ with new constants*

$$\hat{\rho} = \frac{\rho + \gamma\sqrt{\xi}\|\hat{A} - A\|/\min_{k=1,...,r} w_{(k)}}{1 - \gamma\|\hat{A} - A\|}, \quad \hat{\gamma} = \frac{\gamma}{1 - \gamma\|\hat{A} - A\|}.$$

Lemma 1.8 says that if $A$ satisfied the weighted rNSPL of order $(\mathbf{s}, \mathbf{M})$, then so does $A_l$ for large enough $l$. Moreover, given the sequence $\{q_l\}$, we can compute how large $l$ must be and the new constants. For ease of exposition, we drop the notational hats from these constants. We can now state our main result, proven in §4.

**Theorem 3** (**Stable and accurate neural networks with uniform recovery guarantees can be constructed**).
*There exists an algorithm such that for any input sparsity parameters $(\mathbf{s}, \mathbf{M})$, weights $\{w_i\}_{i=1}^N$, $A \in \mathbb{C}^{m \times N}$ (with the input $A$ given by $\{A_l\}$) satisfying the rNSPL with constants $0 < \rho < 1$ and $\gamma > 0$ (also input), and input parameters $n \in \mathbb{N}$, $\{\delta, b_1, b_2\} \subset \mathbb{Q}_{>0}$ and $\upsilon \in (0,1) \cap \mathbb{Q}_{>0}$, the algorithm outputs a neural network $\phi_n$ such that the following holds. For*

$$C_1 = \left(\frac{1+\rho}{2} + (3+\rho)\frac{\kappa^{1/4}}{4}\right)\left(\frac{3+\rho}{1-\rho}\right) \sim \frac{\kappa^{1/4}}{1-\rho}, \quad C_2 = 2\left(\frac{3+\rho}{1-\rho} + \frac{7+\rho}{1-\rho}\frac{\kappa^{1/4}}{2}\right)\gamma \sim \frac{\kappa^{1/4}\gamma}{1-\rho},$$

(1) *(Size)* $\phi_n \in \mathcal{N}_{\mathbf{D}_{(n,p)}, 3np+1, 3}$ with $\mathbf{D}_{(n,p)} = (m, \underbrace{2N+m, 2(N+m), 2N+m+1}_{np \text{ times}}, N)$, where $p \in \mathbb{N}$ with the bound $p \leq \left\lceil \frac{3C_2 \|A\|}{\upsilon} \right\rceil$. Moreover, $\theta^{-1} \sim p^2 (1 + \|w\|_{l^2}) \max\left\{1, \frac{\|w\|_{l^2}}{\|A\|\gamma\sqrt{\xi}}\right\}$.

(2) *(Exponentially Convergent, Uniform and Stable Recovery)* For any pair $(x, y) \in \mathbb{C}^N \times \mathbb{C}^m$ with

$$\frac{2C_1}{C_2\sqrt{\xi}} \cdot \sigma_{\mathbf{s},\mathbf{M}}(x)_{l^1_w} + 2\|Ax - y\|_{l^2} \leq \delta, \quad \|x\|_{l^2} \leq b_1, \quad \|y\|_{l^2} \leq b_2,$$

*we have the following exponentially convergent, uniform and stable recovery guarantees:*

$$\|\phi_n(y) - x\|_{l^2} \leq \frac{2C_1}{\sqrt{\xi}} \cdot \sigma_{\mathbf{s},\mathbf{M}}(x)_{l^1_w} + 2C_2 \cdot \|Ax - y\|_{l^2} + \left(\frac{1+\upsilon}{1-\upsilon}\right)C_2 \cdot \delta + b_2 C_2 \cdot \upsilon^n, \quad (1.16)$$

$$\|\phi_n(y) - x\|_{l^1_w} \leq \left(\frac{3+\rho}{1-\rho}\right)\frac{\sqrt{\xi}}{C_1}\left(\frac{2C_1}{\sqrt{\xi}} \cdot \sigma_{\mathbf{s},\mathbf{M}}(x)_{l^1_w} + 2C_2 \cdot \|Ax-y\|_{l^2} + \left(\frac{1+\upsilon}{1-\upsilon}\right)C_2 \cdot \delta + b_2 C_2 \cdot \upsilon^n\right). \quad (1.17)$$

**Remark 1.9** (The optimal choice of $\upsilon$). For a total budget of $T = 3pn + 1$ layers,

$$\upsilon^n = \exp\left(\frac{(T-1)}{3}\left\lceil\frac{3C_2\|A\|}{\upsilon}\right\rceil^{-1}\log(\upsilon)\right)$$

If we ignore the ceiling function, the optimal choice is $\upsilon = e^{-1}$ (strictly speaking Theorem 3 is only stated for rational $\upsilon$, but we can easily approximate $e^{-1}$). This yields the error term $\upsilon^n = \exp\left(-\frac{(T-1)}{3}\lceil 3C_2 e\|A\|\rceil^{-1}\right)$ and exponential convergence in the number of layers $T$. This is not optimal. For example, a study of the proof of Theorem 3 shows that we can replace $3C_2$ in the exponential by

$$2\left(\frac{1+\rho}{1-\rho} + \frac{3+\rho}{1-\rho}\frac{\kappa^{1/4}}{2}\right)\gamma + \epsilon$$

for arbitrary $\epsilon > 0$. Suppose that we want $b_2 C_2 \cdot r^n \sim \delta$, then the number of layers required is proportional to $C_2\|A\|\log(b_2\delta^{-1})$, and only grows logarithmically with the precision $\delta^{-1}$. This is made precise in Theorem 4, where we apply Theorem 3 to examples in compressive imaging. ⊠

The proof of Theorem 3 uses the optimisation problem $(P_3)$ (defined in (1.2)), in the construction of $\phi_n$. It is also possible to prove similar results using $(P_1)$ and $(P_2)$, but we do not provide the details. The NNs constructed are approximations of unrolled primal-dual iterations for $(P_3)$, with a careful restart scheme to ensure exponential convergence in the number of layers. Further computational experiments beyond the main test are given in §2. The bounds in (1.16) and (1.17) are not quite optimal. If we were able to work in exact arithmetic (taking $\theta \to 0$ and $A_l \to A$), we obtain slightly smaller constants, though these do not affect the asymptotic rates.

**Remark 1.10** (What happens without restart or with unknown $\delta$?). Without the restart scheme, convergence in the number of layers scales as $\mathcal{O}(n^{-1})$. However, one can get rid of the assumption $2C_1/(C_2\sqrt{\xi})\sigma_{\mathbf{s},\mathbf{M}}(x)_{l^1_w} + 2\|Ax - y\|_{l^2} \leq \delta$. (The assumption is to ensure that the reweighting of the restarts do not become too small - in practice, we found that this was not an issue and the assumption was not needed, with (up to small constants) $\delta$ replaced by $2C_1/(C_2\sqrt{\xi})\sigma_{\mathbf{s},\mathbf{M}}(x)_{l^1_w} + 2\|Ax - y\|_{l^2}$ in (1.16). See also the discussion in §2.) More precisely,

the proof of Theorem 3 can be adapted to show the following. For an additional input $\beta \in \mathbb{Q}_{>0}$ (and without inputs $b_2$, $\delta$ and $\upsilon$), there exists an algorithm that computes $\widehat{\phi}_n \in \mathcal{N}_{\mathbf{D}_n, 3n+1, 3}$ with

$$\mathbf{D}_n = (m + N, \underbrace{2N + m, 2(N+m), 2N + m + 1}_{n \text{ times}}, N),$$

such that for any $x, x_0 \in \mathbb{C}^N$ with $\|x\|_{l^2} \leq b_1$ and all $y \in \mathbb{C}^m$, the following reconstruction guarantees hold:

$$\|\widehat{\phi}_n(y, x_0) - x\|_{l^2} \leq \frac{2C_1}{\sqrt{\xi}} \sigma_{\mathbf{s}, \mathbf{M}}(x)_{l^1_w} + 2C_2 \left[ \|Ax - y\|_{l^2} + \frac{\|A\|}{n} \left( \frac{\|x - x_0\|_{l^2}^2}{\beta} + \beta \right) \right], \quad (1.18)$$

$$\|\widehat{\phi}_n(y, x_0) - x\|_{l^1_w} \leq \left( \frac{3+\rho}{1-\rho} \right) \frac{\sqrt{\xi}}{C_1} \left( \frac{2C_1}{\sqrt{\xi}} \sigma_{\mathbf{s}, \mathbf{M}}(x)_{l^1_w} + 2C_2 \left[ \|Ax - y\|_{l^2} + \frac{\|A\|}{n} \left( \frac{\|x - x_0\|_{l^2}^2}{\beta} + \beta \right) \right] \right). \quad (1.19)$$

Here, $x_0$ should be interpreted as an initial guess (an arbitrary input to the NNs) and $\beta$ should be interpreted as a scaling parameter, with optimal scaling $\beta \sim \|x - x_0\|_{l^2}$. A good choice for $\beta$ is $\|x\|_{l^2}$, or, when this is unknown, $\|y\|_{l^2}/\|A\|$. For completeness, we provide a proof sketch of (1.18) and (1.19) at the end of §4.4.3. ⊠

Algorithm unrolling is particularly well-suited to scenarios where it is difficult to collect large training samples. However, training a finite fixed number of layers typically incurs the same stability and generalisation issues mentioned above. Moreover, learning the weights and biases usually prevents the convergence analysis of standard (unlearned) iterative methods carrying over. In particular, there is no guarantee of objective function minimisation (let alone convergence of the iterated arguments) or any form of convergence as the number of layers increases. A subtle, yet fundamental, point regarding iterative methods, whether they are unrolled as a NN and supplemented with learned parameters or not, is the following. Theorem 2 states that, in general, the optimisation problems $(P_1)$, $(P_2)$, and $(P_3)$ are non-computable. This is despite the fact that there are many results in the literature describing rates of convergence for iterative methods. The resolution of this apparent puzzle is that convergence results regarding iterative methods are typically given in terms of the *objective function* that is being minimised (see also Theorem 5, which we use to prove Theorem 3). As the proof of Theorem 3 shows, it is crucial to have conditions such as the rNSPL to convert these objective function bounds to the desired error bounds on the distance to the *minimisers* or vector $x$. Moreover, this property has the key effect of allowing exponential convergence through restarting and reweighting.

1.3. **Examples in image recovery.** As an example application of Theorem 3, we consider the case of Fourier and Walsh sampling, using the Haar wavelets as the sparsifying transform. Our results can be generalised to the infinite-dimensional setting with the use of higher-order Daubechies wavelets (though the results are more complicated to write down), and we refer the reader to [3] for compressed sensing in infinite dimensions. We first define the concept of multilevel random subsampling [4].

**Definition 1.11** (Multilevel random subsampling). *Let $\mathbf{N} = (N_1, \ldots, N_l) \in \mathbb{N}^l$, where $1 \leq N_1 < \cdots < N_l = N$ and $\mathbf{m} = (m_1, \ldots, m_l) \in \mathbb{N}^l$ with $m_k \leq N_k - N_{k-1}$ for $k = 1, \ldots, l$, and $N_0 = 0$. For each $k = 1, \ldots, l$, let $\mathcal{I}_k = \{N_{k-1} + 1, \ldots, N_k\}$ if $m_k = N_k - N_{k-1}$ and if not, let $t_{k,1}, \ldots, t_{k,m_k}$ be chosen uniformly and independently from the set $\{N_{k-1} + 1, \ldots, N_k\}$ (with possible repeats), and set $\mathcal{I}_k = \{t_{k,1}, \ldots, t_{k,m_k}\}$. If $\mathcal{I} = \mathcal{I}_{\mathbf{N}, \mathbf{m}} = \mathcal{I}_1 \cup \cdots \cup \mathcal{I}_l$ we refer to $\mathcal{I}$ as an $(\mathbf{N}, \mathbf{m})$-multilevel subsampling scheme.*

**Definition 1.12** (Multilevel subsampled unitary matrix). *A matrix $A \in \mathbb{C}^{m \times N}$ is an $(\mathbf{N}, \mathbf{m})$-multilevel subsampled unitary matrix if $A = P_{\mathcal{I}} DU$ for a unitary matrix $U \in \mathbb{C}^{N \times N}$ and $(\mathbf{N}, \mathbf{m})$-multilevel subsampling scheme $\mathcal{I}$. Here $D$ is a diagonal scaling matrix with diagonal entries*

$$D_{ii} = \sqrt{\frac{N_k - N_{k-1}}{m_k}}, \quad i = N_{k-1} + 1, \ldots, N_k, \quad k = 1, \ldots, l$$

*and $P_{\mathcal{I}}$ denotes the projection onto the linear span of the subset of the canonical basis indexed by $\mathcal{I}$.*

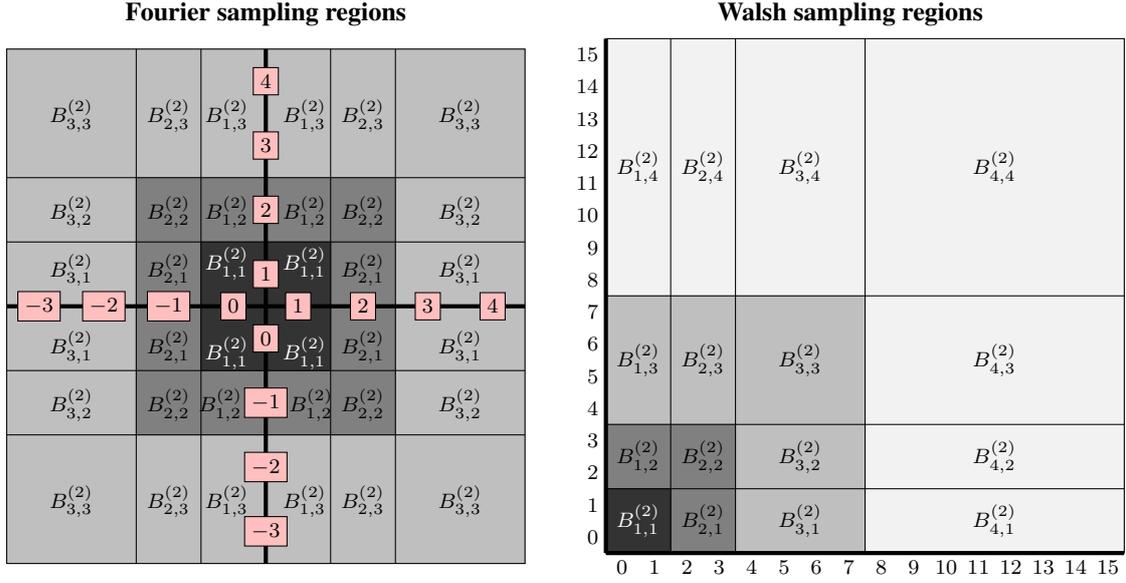

FIGURE 1. The different sampling regions used for the sampling patterns for Fourier (left, $r = 3$) and Walsh (right, $r = 4$). The axis labels correspond to the frequencies in each band and the annular regions are shown as the shaded greyscale regions.

Throughout this subsection, we let $K = 2^r$ for $r \in \mathbb{N}$, and consider vectors on $\mathbb{C}^K$ or $d$-dimensional tensors on $\mathbb{C}^{K \times \cdots \times K}$. To keep consistent notation with previous sections, we set $N = K^d$ so that the objective is to recover a vectorised $x \in \mathbb{C}^N$. The following can also be generalised to rectangles (i.e. $\mathbb{C}^{2^{r_1} \times \cdots \times 2^{r_d}}$ with possibly different $r_1, ..., r_d$) or dimensions that are not powers of two.

Let $V \in \mathbb{C}^{N \times N}$ be either the matrix $F^{(d)}$ or $W^{(d)}$, corresponding to the $d$-dimensional discrete Fourier or Walsh transform (see §5.1). In the Fourier case, we divide the different frequencies $\{-K/2 + 1, \ldots, K/2\}^d$ into dyadic bands. For $d = 1$, we let $B_1 = \{0, 1\}$ and $B_k = \{-2^{k-1}+1, \ldots, -2^{k-2}\} \cup \{2^{k-2}+1, \ldots, 2^{k-1}\}$ for $k = 2, \ldots, r$. In the Walsh case, we define the frequency bands $B_1 = \{0, 1\}$ and $B_k = \{2^{k-1}, \ldots, 2^k - 1\}$ for $k = 2, \ldots, r$ in the one-dimensional case. In the general $d$-dimensional case for Fourier or Walsh sampling, we set

$$B_{\mathbf{k}}^{(d)} = B_{k_1} \times \ldots \times B_{k_d}, \quad \mathbf{k} = (k_1, \ldots, k_d) \in \mathbb{N}^d.$$

For a $d$-dimensional tensor $c \in \mathbb{C}^{K \times \cdots \times K}$, we assume we can observe subsampled measurements of $V \text{vec}(c)$, where $\text{vec}(c) \in \mathbb{C}^N$ is a vectorised version of $c$. To recover a sparse representation, we consider the Haar wavelet coefficients. We denote the discrete Haar Wavelet transform by $\Phi \in \mathbb{C}^{N \times N}$, and note that $\Psi^* = \Psi^{-1}$ since $\Psi$ is unitary. In other words, we consider a multilevel subsampled unitary matrix (Definition 1.12), with $U = V\Psi^*$. Given $\{m_{\mathbf{k}=(k_1,\ldots,k_d)}\}_{k_1,\ldots,k_d=1}^r$, we use a multilevel random sampling such that $m_{\mathbf{k}}$ measurements are chosen from $B_{\mathbf{k}}^{(d)}$ according to Definition 1.11. This corresponds to $l = r^d$ and the $N_i$'s can be chosen given a suitable ordering of the Fourier/Walsh basis. The sparsity in levels structure (Definition 1.6) is chosen to correspond to the $r$ wavelet levels. A pictorial representation is given in Figure 1. Finally, we define

$$\mathcal{M}_{\mathcal{F}}(\mathbf{s}, \mathbf{k}) := \sum_{j=1}^{\|\mathbf{k}\|_{l^\infty}} s_j \prod_{i=1}^d 2^{-|k_i - j|} + \sum_{j=\|\mathbf{k}\|_{l^\infty}+1}^r s_j 2^{-2(j - \|\mathbf{k}\|_{l^\infty})} \prod_{i=1}^d 2^{-|k_i - j|} \quad (1.20)$$

$$\mathcal{M}_{\mathcal{W}}(\mathbf{s}, \mathbf{k}) := s_{\|\mathbf{k}\|_{l^\infty}} \prod_{i=1}^d 2^{-|k_i - \|\mathbf{k}\|_{l^\infty}|}. \quad (1.21)$$

For notational convenience, we also define $\mathcal{Z} = \max\left\{1, \frac{\max_{j=1,\ldots,r} w_{(j)} \sqrt{(M_j - M_{j-1})}}{\sqrt{\xi(\mathbf{s}, \mathbf{M}, w)}}\right\}$.

We now state the main theorem of this subsection (proven in §5), which states how many samples are needed and the number of layers of the NN needed, which only depends logarithmically on the error $\delta$, a consequence of the exponential convergence in Theorem 3. We discuss the sampling conditions below.

**Theorem 4.** *Consider the above setup of recovering a d-dimensional tensor $c \in \mathbb{C}^{K^d}$ ($N = K^d$) from subsampled Fourier or Walsh measurements $Vc$, such that $A$ is a multilevel subsampled unitary matrix with respect to $U = V\Psi^*$. Let $\epsilon_{\mathbb{P}} \in (0,1)$ and $\mathcal{L} = d \cdot r^2 \cdot \log(2m) \cdot \log^2(s \cdot \kappa(\mathbf{s}, \mathbf{M}, w)) + \log(\epsilon_{\mathbb{P}}^{-1})$. Suppose that:*

- *(a) In the Fourier case*
$$m_{\mathbf{k}} \gtrsim \kappa(\mathbf{s}, \mathbf{M}, w) \cdot \mathcal{M}_{\mathcal{F}}(\mathbf{s}, \mathbf{k}) \cdot \mathcal{L}. \tag{1.22}$$

- *(b) In the Walsh case*
$$m_{\mathbf{k}} \gtrsim \kappa(\mathbf{s}, \mathbf{M}, w) \cdot \mathcal{M}_{\mathcal{W}}(\mathbf{s}, \mathbf{k}) \cdot \mathcal{L}. \tag{1.23}$$

*Then with probability at least $1 - \epsilon_{\mathbb{P}}$, $A$ satisfies the weighted rNSPL of order $(\mathbf{s}, \mathbf{M})$ with constants $\rho = 1/2$ and $\gamma = \sqrt{2}$. The conclusion of Theorem 3 then holds for the uniform recovery of the Haar wavelet coefficients*

$$x = \Psi c \in \mathbb{C}^N. \tag{1.24}$$

*Moreover, for any $\delta \in (0,1)$, let $\mathcal{J}(\delta, \mathbf{s}, \mathbf{M}, w)$ be the collection of all $y \in \mathbb{C}^m$ with $y = P_{\mathcal{I}}DVc + e$ where*

$$\|c\|_{l^2} \leq 1, \quad \max\left\{\frac{\sigma_{\mathbf{s}, \mathbf{M}}(\Psi c)_{l^1_w}}{\sqrt{\xi}}, \|e\|_{l^2}\right\} \leq \delta. \tag{1.25}$$

*Then we construct via an algorithm, a neural network $\phi \in \mathcal{N}_{\mathbf{D}, 3n+1, 3}$ such that with probability at least $1 - \epsilon_{\mathbb{P}}$,*

$$\|\phi(y) - c\|_{l^2} \lesssim \kappa^{1/4}\delta, \quad \forall y = P_{\mathcal{I}}DVc + e \in \mathcal{J}(\delta, \mathbf{s}, \mathbf{M}, w). \tag{1.26}$$

*The network parameters are*

$$\mathbf{D} = (m, \underbrace{2N+m, 2(N+m), 2N+m+1}_{n \text{ times}}, N), \quad n \leq \left\lceil \log\left(\delta^{-1}\mathcal{Z}\right)\kappa^{1/4}\mathcal{Z} \right\rceil. \tag{1.27}$$

The sampling conditions in (1.22) and (1.23) are optimised by minimising $\kappa(\mathbf{s}, \mathbf{M}, w)$. Up to a constant scale, this corresponds to the choice $w_{(j)} = \sqrt{s/s_j}$ and

$$n = \left\lceil \log\left(\delta^{-1} \max_{j=1,\ldots,r}\sqrt{\max\left\{1, \frac{M_j - M_{j-1}}{rs_j}\right\}}\right) r^{1/4} \max_{j=1,\ldots,r}\sqrt{\max\left\{1, \frac{M_j - M_{j-1}}{rs_j}\right\}} \right\rceil.$$

Up to log-factors, the measurement condition then becomes equivalent to the currently best-known oracle estimator (where one assumes apriori knowledge of the support of the vector) [2, Prop. 3.1]. For Fourier measurements, we can interpret the condition as follows. For $d = 1$, this estimate yields the sampling estimates

$$m_k \gtrsim \left(\sum_{j=1}^{k} s_j 2^{-|k-j|} + \sum_{j=k+1}^{r} s_j 2^{-3|k-j|}\right) \cdot r \cdot \mathcal{L}.$$

In other words, up to logarithmic factors and exponentially small terms, $s_j$ measurements are needed in each level. Furthermore, if $s_1 = \ldots = s_r = s_*$ and $d = 2$ then (1.22) holds if

$$m_{(k_1, k_2)} \gtrsim s_* 2^{-|k_1 - k_2|} \cdot r \cdot \mathcal{L}. \tag{1.28}$$

Another interpretation is gained by considering

$$m_k = \sum_{\|\mathbf{k}\|_{l^\infty} = k} m_{\mathbf{k}}, \quad k = 1, \ldots, r,$$

the number of samples per annular region. We then have

$$m_k \gtrsim 3^d d \left(s_k + \sum_{l=1}^{k-1} s_l 2^{-(k-l)} + \sum_{l=k+1}^{r} s_l 2^{-3(l-k)}\right) \cdot r \cdot \mathcal{L}, \tag{1.29}$$

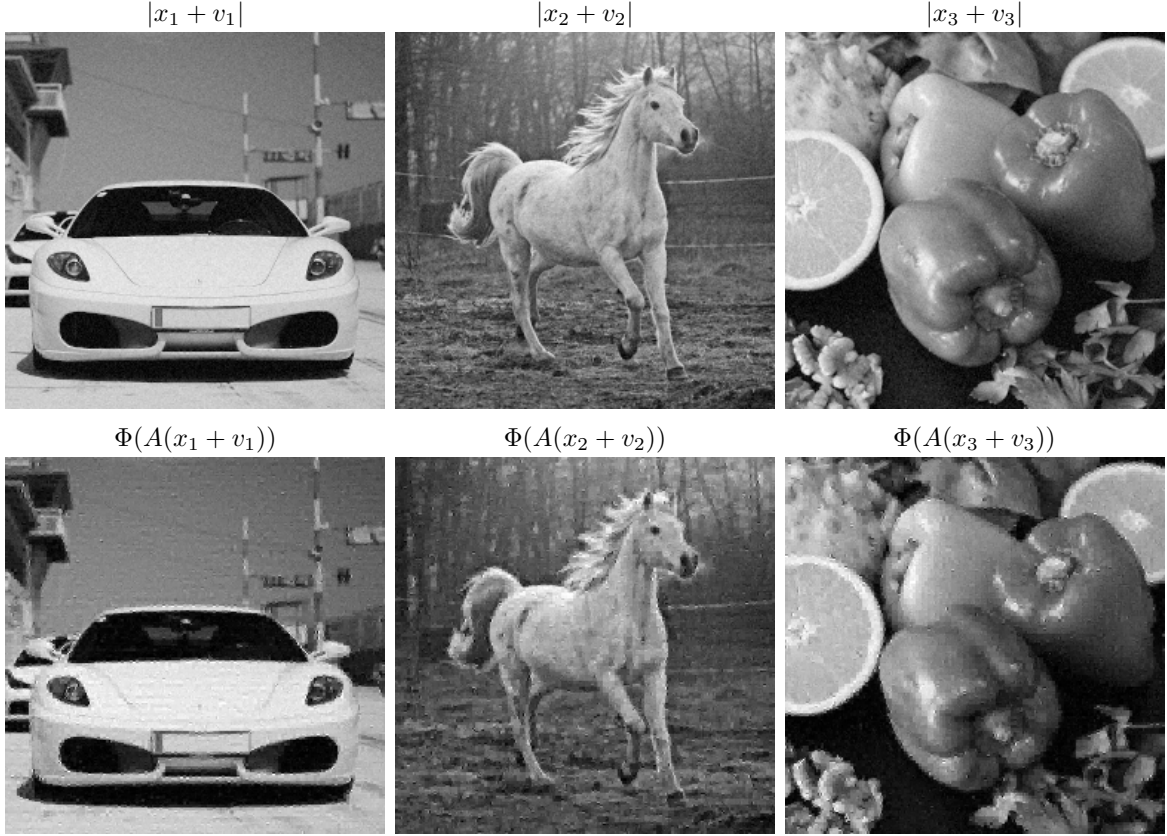

FIGURE 2. (**FIRENET withstands worst-case perturbations and generalises well**). To show that FIRENET generalises well and is stable, we consider three different images $x_j$, $j = 1, 2, 3$. For each image $x_j$ we compute a perturbation $v_j$ meant to simulate worst-case effect for a FIRENET $\Phi$ with $n = 5$ and $p = 5$. The first row shows the perturbed images $x_j + v_j$, whereas the second row shows the FIRENET reconstructions from data $A(x_j + v_j)$. Here $A \in \mathbb{C}^{m \times N}$ is a subsampled discrete Fourier transform with $m/N = 0.25$ and $N = 256^2$. The perturbations $v_j$ have magnitude $\|Av_j\|_{l^2}/\|Ax_j\|_{l^2} \geq 0.05$ in the measurement domain.

which is the same estimate as the one-dimensional case for bounded $d$. Note that the number of samples required in each annular region is (up logarithmic factors) proportional to the corresponding sparsity $s_k$ with additional exponentially decaying terms dependent on $s_l, l \neq k$. In the case of Walsh sampling, (1.28) remains the same whereas (1.29) becomes $m_k \gtrsim 2^d \cdot d \cdot r \cdot \mathcal{L} \cdot s_k$, with no terms from the sparsity levels $s_l, l \neq k$.

## 2. Further examples of FIRENET

2.1. **Generalisation properties.** To demonstrate the generalisation properties of our NNs, Figure 2 shows the stability test (see main text) applied to FIRENETs for a range of images. This shows stability across different types of images and highlights an important fact. Namely, methods based on conditions such as Definition 1.7 allow great generalisation properties and avoid time-consuming and expensive retraining of NNs for different classes of images. As well as being rigorously proven to be stable, FIRENETs are accurate for images which are sparse in wavelets. As most images are sparse in wavelets, these networks also show great generalisation properties to unseen images.

2.2. **Exponential convergence.** We now provide a computational experiment to demonstrate the convergence in the number of layers stated in Theorem 4 (and Theorem 3). Note that the matrix $A$ and its adjoint can be implemented rapidly using the fast Fourier transform (or fast Walsh–Hadamard transform). We take the image

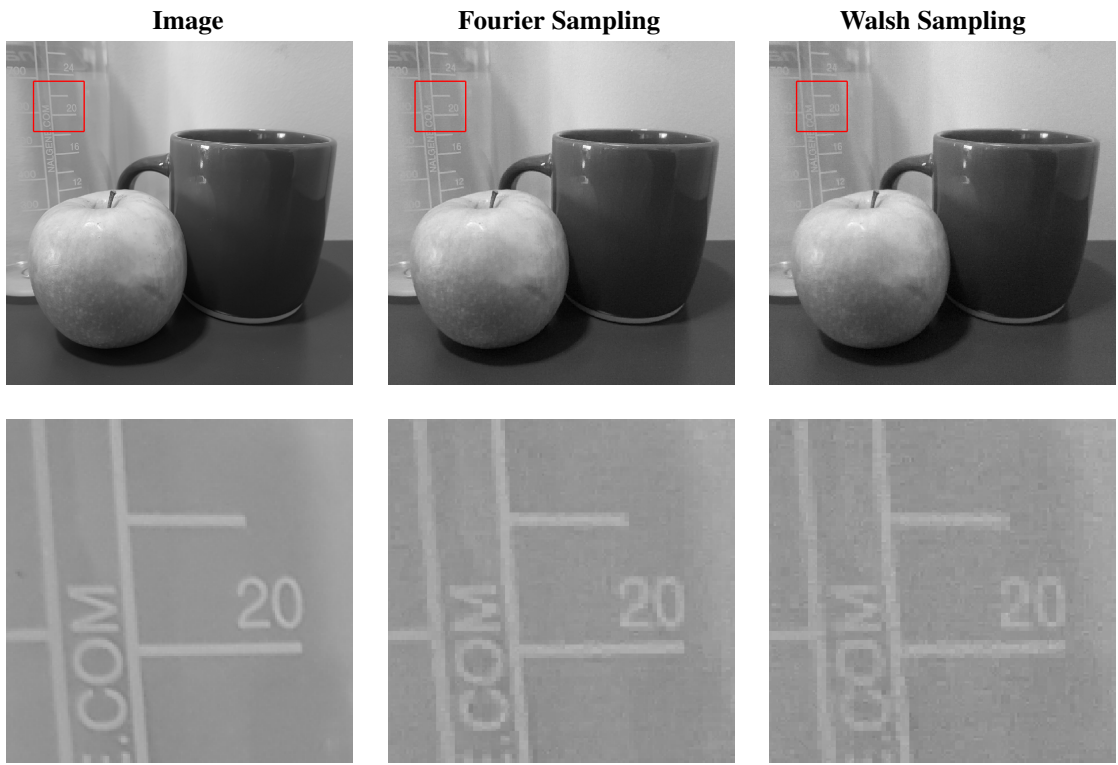

FIGURE 3. Left: The true image. Middle: Reconstruction from noisy Fourier measurements. Right: Reconstruction from noisy Walsh measurements. Both images were reconstructed using only a 15% sampling rate according to the sampling patterns in Figure 1 and $n = p = 5$. The top row shows the full image and the bottom row shows a zoomed in section (corresponding to the red boxes in the top row).

shown in Figure 3, a subsampling rate of only 15%, and corrupt the measurements by adding 2% Gaussian noise. Figure 3 shows the reconstructions using Fourier and Walsh sampling and Haar wavelets. Similar results hold for other wavelets, such as Daubechies wavelets with a larger number of vanishing moments. In fact, the reconstruction results are better than those shown for the Haar wavelet system. We have chosen to show the Haar wavelet results because this is the system for which Theorem 4 is stated. For the reconstruction, we take $\lambda = 0.00025, \tau = \sigma = 1, p = 5$ and the weights as discussed in §1.1.3. In the spirit of no parameter tuning, the weights were selected based on a standard phantom image, and not the image we use to test the algorithm. These parameters are certainly not optimal, and instead were chosen simply to emphasise that we have deliberately avoided parameter tuning. Moreover, we found that the choice of $\delta$ in the algorithm was of little consequence, so have taken $\delta = 10^{-9}$.

Figure 4 shows the convergence in the number of inner iterations (or, equivalently, $n$ - the total number of inner iterations is $np$, and hence we have not specified $n$, which is typically chosen to be 5). We show the error between the constructed image after $j$ iterations (denoted by $c_j$) and the true image (denoted by $c$), as well as the convergence of the objective function which we denote by $F$ in the figure caption. To compute the minimum of $F$, denoted $F^*$, we ran several thousand iterates of the non-restarted version of the algorithm so that the error in the value of $F^*$ is at least an order of magnitude smaller than the shown values of $F(c_j) - F^*$. Whilst the objective function is guaranteed to converge to the minimum value when computing $F^*$ this way, there is no guarantee that the vectors computed by the non-restarted version converge to a minimiser, as demonstrated by the non-computability results in Theorem 2. However, in this case, the non-restarted version converged to a vector $c^*$ up to an error much smaller than $\|c - c^*\|_{l^2}$. Hence $\|c - c^*\|_{l^2}$ indicates the minimum error we can expect from using $(P_3)$ to recover the image.

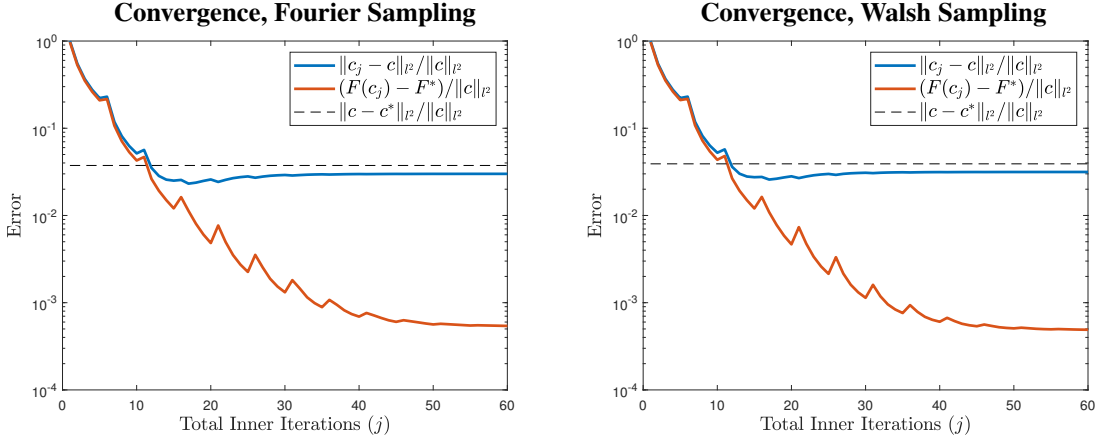

FIGURE 4. The convergence of the algorithm in the number of inner iterations. The dashed line shows $\|c - c^*\|_{l^2}/\|c\|_{l^2}$. In both cases, the error between the reconstruction and the image decreases exponentially until this bound is reached. The objective function gap decreases exponentially slightly beyond this point, demonstrating that the robust null space property (in levels) controls the $l^2$-norm difference between vectors (locally around $c^*$) down to the error $\|c - c^*\|_{l^2}$ (see the bound in (4.18) in our proof).

The figure shows the expected exponential convergence, as the number of inner iterations increases, of the objective function values as well as $c_j$ to $c$ until the error is of the order $\|c - c^*\|_{l^2}$. This corresponds to an initial phase of exponential convergence, where the $\upsilon^{-n}$ term (with $\upsilon = e^{-1}$) is dominant in Theorem 3, followed by a plateau to the minimal error $\|c - c^*\|_{l^2}$ (shown as the dotted line). This plateau occurs due to inexact measurements (the noise) and the fact that the image does not have exactly sparse wavelet coefficients. This corresponds to the robust null space property (in levels) only being able to bound the distance $\|c - c_j\|_{l^2}$ up to the same order as $\|c - c^*\|_{l^2}$. In other words, we can only accelerate convergence up to this error bound. The error plateau disappears in the limit of exactly sparse vectors and zero noise (in the limit $\delta \downarrow 0$ in Theorem 3), and one gains exponential convergence down to essentially machine precision. Finally, the acceleration is of great practical interest. Rather than the several hundreds (or even thousands) of iterations that are typically needed for solving compressed sensing optimisation problems with first-order iterative methods, we obtain optimal accuracy in under 20 iterations. This was found for a range of different images, subsampling rates etc. The fact that so few layers are needed, coupled with the fast transforms for implementing the affine maps in the NNs, makes the NNs very computationally efficient and competitive speed-wise with state-of-the-art DL.

## 3. PROOF OF THEOREM 2 AND TOOLS FROM THE SCI HIERARCHY

In this section, we prove Theorem 2. To do this, we need some analytical results regarding phase transitions of solutions of $(P_1), (P_2)$ and $(P_3)$ which are given in §3.3.2. However, before analysing these phase transitions, we need some preliminary definitions regarding algorithms, inexact inputs, and condition numbers. There are two main reasons for this framework. First, because our definitions are general, they lead to stronger impossibility results than when restricted to specific models of computation. Second, our framework greatly simplifies the proofs and makes it clear what the key mechanisms behind the proofs are (§3.3.2 describes this in terms of phase transitions of minimisers). The following discussions are self-contained.

3.1. **Algorithmic preliminaries: a user-friendly guide.** We begin with a definition of a computational problem, which is deliberately general in order to capture any computational problem.

**Definition 3.1** (Computational problem). *Let $\Omega$ be some set, which we call the domain, and $\Lambda$ be a set of complex valued functions on $\Omega$ such that for $\iota_1, \iota_2 \in \Omega$, then $\iota_1 = \iota_2$ if and only if $f(\iota_1) = f(\iota_2)$ for all $f \in \Lambda$, called an evaluation set. Let $(\mathcal{M}, d)$ be a metric space, and finally let $\Xi : \Omega \to \mathcal{M}$ be a function which we call*

*the problem function. We call the collection* $\{\Xi, \Omega, \mathcal{M}, \Lambda\}$ *a computational problem. When it is clear what* $\mathcal{M}$ *and* $\Lambda$ *are, we write* $\{\Xi, \Omega\}$ *for brevity.*

**Remark 3.2** (Multivalued problems). In some cases, such as when considering the optimisation problems $(P_j)$ that may have more than one solution, we consider $\Xi(\iota) \subset \mathcal{M}$. With an abuse of notation, we then set $d(x, \Xi(\iota)) = \text{dist}(x, \Xi(\iota)) = \inf_{y \in \Xi(\iota)} d(x, y)$ and this distinction will be made clear from context. ⊠

The set $\Omega$ is the set of objects that give rise to our computational problems. The problem function $\Xi : \Omega \to \mathcal{M}$ is what we are interested in computing. Finally, the set $\Lambda$ is the collection of functions that provide us with the information we are allowed to read as input to an algorithm. For example, $\Omega$ could consist of a collection of matrices $A$ and data $y$ in (1.1), $\Lambda$ could consist of the pointwise entries of the vectors and matrices in $\Omega$, $\Xi$ could represent the solution set (with the possibility of more than one solution as in Remark 3.2) of any of the problems $(P_j)$ and $(\mathcal{M}, d)$ could be $\mathbb{C}^N$ with the usual Euclidean metric (or any other suitable metric).

Given the definition of a computational problem, we need the definition of a general algorithm, whose conditions hold for any reasonable notion of a deterministic algorithm. Throughout this paper, we deal with the case that $\Lambda = \{f_j\}_{j \in \beta}$, where $\beta$ is some (at most) countable index set.

**Definition 3.3** (General Algorithm). *Given a computational problem* $\{\Xi, \Omega, \mathcal{M}, \Lambda\}$, *a general algorithm is a mapping* $\Gamma : \Omega \to \mathcal{M}$ *such that for each* $\iota \in \Omega$

  (i) *There exists a non-empty finite subset of evaluations* $\Lambda_\Gamma(\iota) \subset \Lambda$,
  (ii) *The action of* $\Gamma$ *on* $\iota$ *only depends on* $\{\iota_f\}_{f \in \Lambda_\Gamma(\iota)}$ *where* $\iota_f := f(\iota)$,
  (iii) *For every* $\kappa \in \Omega$ *such that* $\kappa_f = \iota_f$ *for every* $f \in \Lambda_\Gamma(\iota)$, *it holds that* $\Lambda_\Gamma(\kappa) = \Lambda_\Gamma(\iota)$.

*If, in addition, there exists a canonical ordering* $\Lambda_\Gamma(\iota) = \{f^\Gamma_{\iota,1} = f_{k_1}, ..., f^\Gamma_{\iota,S_\Gamma(\iota)} = f_{k_{S_\Gamma(\iota)}}\}$, *where* $S_\Gamma(\iota) = |\Lambda_\Gamma(\iota)|$, *such that if* $\kappa \in \Omega$ *and* $f^\Gamma_{\iota,j}(\iota) = f^\Gamma_{\iota,j}(\kappa)$ *for all* $j \leq r < S_\Gamma(\iota)$, *then* $f^\Gamma_{\iota,j} = f^\Gamma_{\kappa,j}$ *for all* $j \leq r+1$, *then we call* $\Gamma$ *a Sequential General Algorithm. In this case, we use the notation* $k_j(\Gamma, \iota)$ *to denote the ordered indices corresponding to the evaluation functions that the algorithm reads.*

The three properties of a general algorithm are the most basic natural properties we would expect any deterministic computational device to obey. The first condition says that the algorithm can only take a finite amount of information, though it is allowed adaptively to choose, depending on the input, the finite amount of information that it reads. The second condition ensures that the algorithm's output only depends on its input, or rather the information that it has accessed (or "read"). The final condition is very important and ensures that the algorithm produces outputs and accesses information consistently. In other words, if it sees the same information for two different inputs, then it cannot behave differently for those inputs. Note that the definition of a general algorithm is more general than the definition of a Turing machine [24] or a Blum–Shub–Smale (BSS) machine [10], which can be thought of as digital and analog computational devices respectively. In particular, a general algorithm has no restrictions on the operations allowed. The extra condition for a sequential general algorithm is satisfied by any algorithm defined by a computational machine with input of readable information (one should think of the ordered indices of the evaluation functions as corresponding to sequentially reading the tape which encodes the input information). Hence, a sequential general algorithm is still more general than a Turing or a BSS machine. Complete generality in Definition 3.3 is used for two primary reasons:

  (i) *Strongest possible bounds:* Since Definition 3.3 is completely general, the lower bounds hold in any model of computation, such as a Turing machine or a BSS machine. On the other hand, the algorithms we construct in this paper are made to work using only arithmetic operations over the rationals. Hence, we obtain the strongest possible lower bounds and the strongest possible upper bounds.
  (ii) *Simplified exposition:* Using the concept of a general algorithm considerably simplifies the proofs of lower bounds and allows us to see precisely the mechanisms behind the proofs.

Next, we consider the definition of a randomised general algorithm, which again is more general than a probabilistic Turing or probabilistic BSS machine. Randomised algorithms are widely used in practice

in areas such as optimisation, algebraic computation, machine learning, and network routing. In the case of Turing machines, it is currently unknown, in the sense of polynomial runtime, whether randomisation is beneficial from a complexity class viewpoint [7, Ch. 7], however, rather intriguingly this is not the case for BSS machines [10, Ch. 17] (some of the proofs in this reference are non-constructive - it is an open problem whether any probabilistic BSS machine can be simulated by a deterministic machine having the same machine constants and with only a polynomial slowdown). Nevertheless, randomisation is an extremely useful tool in practice. From a machine learning point of view, we also want to consider randomised algorithms to capture procedures such as stochastic gradient descent which are commonly used to train NNs. As developed in [8], the concept of a general algorithm can be extended to a randomised general algorithm. This concept allows for universal impossibility results regardless of the computational model.

**Definition 3.4** (Randomised General Algorithm)**.** *Given a computational problem $\{\Xi, \Omega, \mathcal{M}, \Lambda\}$, a Randomised General Algorithm (RGA) $\Gamma^{\mathrm{ran}}$ is a collection $X$ of general algorithms $\Gamma : \Omega \to \mathcal{M}$, a sigma-algebra $\mathcal{F}$ on $X$ and a family of probability measures $\{\mathbb{P}_\iota\}_{\iota \in \Omega}$ on $\mathcal{F}$ such that the following conditions hold:*

*(1) For each $\iota \in \Omega$, the mapping $\Gamma^{\mathrm{ran}}_\iota : (X, \mathcal{F}) \to (\mathcal{M}, \mathcal{B})$ defined by $\Gamma^{\mathrm{ran}}_\iota(\Gamma) = \Gamma(\iota)$ is a random variable, where $\mathcal{B}$ is the Borel sigma-algebra on $\mathcal{M}$.*

*(2) For each $n \in \mathbb{N}$ and $\iota \in \Omega$, the set $\{\Gamma \in X \,:\, \sup\{m \in \mathbb{N} : f_m \in \Lambda_\Gamma(\iota)\} \leq n\} \in \mathcal{F}$.*

*(3) For each $\iota_1, \iota_2 \in \Omega$ and $E \in \mathcal{F}$, such that for every $\Gamma \in E$ we have $f(\iota_1) = f(\iota_2)$ for every $f \in \Lambda_\Gamma(\iota_1)$, then $\mathbb{P}_{\iota_1}(E) = \mathbb{P}_{\iota_2}(E)$.*

*With slight abuse of notation, we denote the family of randomised general algorithms by* RGA.

The first two conditions are measure theoretic to avoid pathological cases and ensure that "natural sets" one might define for a random algorithm (such as notions of stopping times) are measurable. These conditions hold for all standard probabilistic machines (such as a Turing or BSS machine). The third condition ensures consistency, namely, that in the case of identical evaluations, the laws of the output cannot change. Finally, we will use the standard definition of a probabilistic Turing machine (which is a particular case of Definition 3.4). However, to make sense of probabilistic Turing machines in our context (in particular, to restrict operations to the rationals which can be encoded by the natural numbers), we must define the notion of inexact input.

Suppose we are given a computational problem $\{\Xi, \Omega, \mathcal{M}, \Lambda\}$, and that $\Lambda = \{f_j\}_{j \in \beta}$, where we remind the reader that $\beta$ is some index set that can be finite or countably infinite. However, obtaining $f_j$ may be a computational task on its own, which is exactly the problem in most areas of computational mathematics. In particular, for $\iota \in \Omega$, $f_j(\iota)$ could be the number $e^{\frac{\pi}{j}i}$ for example. Hence, we cannot access or store $f_j(\iota)$ on a computer, but rather $f_{j,n}(\iota)$ where $f_{j,n}(\iota) \to f_j(\iota)$ as $n \to \infty$. This idea is formalised in the definition below, however, to put this in perspective it is worth mentioning the Solvability Complexity Index (SCI) hierarchy.

**Remark 3.5** (The Solvability Complexity Index (SCI) hierarchy)**.** The SCI of a computational problem is the smallest number of limits needed in order to compute the solution. The full hierarchy is described in [9], and the mainstay of the hierarchy are the $\Delta^\alpha_k$ classes. The $\alpha$ denotes the model of computation. Informally, we have the following description. Given a collection $\mathcal{C}$ of computational problems, then

(i) $\Delta^\alpha_0$ is the set of problems that can be computed in finite time, the SCI $= 0$.

(ii) $\Delta^\alpha_1$ is the set of problems that can be computed using one limit (the SCI $= 1$) with control of the error, i.e. $\exists$ a sequence of algorithms $\{\Gamma_n\}$ such that $d(\Gamma_n(\iota), \Xi(\iota)) \leq 2^{-n}$, $\forall \iota \in \Omega$.

(iii) $\Delta^\alpha_2$ is the set of problems that can be computed using one limit (the SCI $= 1$) without error control, i.e. $\exists$ a sequence of algorithms $\{\Gamma_n\}$ such that $\lim_{n \to \infty} \Gamma_n(\iota) = \Xi(\iota)$, $\forall \iota \in \Omega$.

(iv) $\Delta^\alpha_{m+1}$, for $m \in \mathbb{N}$, is the set of problems that can be computed by using $m$ limits, (the SCI $\leq m$), i.e. $\exists$ a family of algorithms $\{\Gamma_{n_m,\ldots,n_1}\}$ with $\lim_{n_m \to \infty} \ldots \lim_{n_1 \to \infty} \Gamma_{n_m,\ldots,n_1}(\iota) = \Xi(\iota)$, $\forall \iota \in \Omega$.  ⊠

The above hierarchy gives rise to the concept of '$\Delta_1$-information.' That is, in informal terms, the problem of obtaining the inexact input to the computational problem is a $\Delta_1$ problem. One may think of an algorithm

taking the number $\exp(1)$ or $\sqrt{2}$ as input. Indeed, one can never produce an exact version of these numbers to the algorithm, however, one can produce an approximation to an arbitrarily small error.

**Definition 3.6** ($\Delta_1$-information). *Let $\{\Xi, \Omega, \mathcal{M}, \Lambda\}$ be a computational problem. We say that $\Lambda$ has $\Delta_1$-information if each $f_j \in \Lambda$ is not available, however, there are mappings $f_{j,n} : \Omega \to \mathbb{Q} + i\mathbb{Q}$ such that $|f_{j,n}(\iota) - f_j(\iota)| \leq 2^{-n}$ for all $\iota \in \Omega$. Finally, if $\widehat{\Lambda}$ is a collection of such functions described above such that $\Lambda$ has $\Delta_1$-information, we say that $\widehat{\Lambda}$ provides $\Delta_1$-information for $\Lambda$. Moreover, we denote the family of all such $\widehat{\Lambda}$ by $\mathcal{L}^1(\Lambda)$.*

We want to have algorithms that can handle all computational problems $\{\Xi, \Omega, \mathcal{M}, \widehat{\Lambda}\}$ whenever $\widehat{\Lambda} \in \mathcal{L}^1(\Lambda)$. In order to formalise this, we define what we mean by a computational problem with $\Delta_1$-information.

**Definition 3.7** (Computational problem with $\Delta_1$-information). *A computational problem where $\Lambda$ has $\Delta_1$-information is denoted by $\{\Xi, \Omega, \mathcal{M}, \Lambda\}^{\Delta_1} := \{\widetilde{\Xi}, \widetilde{\Omega}, \mathcal{M}, \widetilde{\Lambda}\}$, where*

$$\widetilde{\Omega} = \left\{\widetilde{\iota} = \{f_{j,n}(\iota)\}_{j,n \in \beta \times \mathbb{N}} : \iota \in \Omega, \{f_j\}_{j \in \beta} = \Lambda, |f_{j,n}(\iota) - f_j(\iota)| \leq 2^{-n}\right\},$$

*Moreover, if $\widetilde{\iota} = \{f_{j,n}(\iota)\}_{j,n \in \beta \times \mathbb{N}} \in \widetilde{\Omega}$ then we define $\widetilde{\Xi}(\widetilde{\iota}) = \Xi(\iota)$ and $\widetilde{f}_{j,n}(\widetilde{\iota}) = f_{j,n}(\iota)$. We also set $\widetilde{\Lambda} = \{\widetilde{f}_{j,n}\}_{j,n \in \beta \times \mathbb{N}}$. Note that $\widetilde{\Xi}$ is well-defined by Definition 3.1 of a computational problem and the definition of $\widetilde{\Omega}$ includes all possible instances of $\Delta_1$-information $\widehat{\Lambda} \in \mathcal{L}^1(\Lambda)$.*

We can now define a probabilistic Turing machine for $\{\Xi, \Omega, \mathcal{M}, \Lambda\}$, where the algorithm $\Gamma$ is executed by a Turing machine [24], that has an oracle tape consisting of $\{\widetilde{\iota}_f\}_{f \in \widetilde{\Lambda}}$. In what follows, we have deliberately not written down the (lengthy) definition of a Turing machine (found in any standard text [7]), which one should think of as an effective algorithm or computer programme (the famous Church–Turing thesis).

**Definition 3.8.** *Given the definition of a Turing machine, a probabilistic Turing machine for $\{\Xi, \Omega, \mathcal{M}, \Lambda\}$ is a Turing machine that has an oracle tape consisting of $\{\widetilde{\iota}_f\}_{f \in \widetilde{\Lambda}}$ (for $\widetilde{\iota} \in \widetilde{\Omega}$), with an additional read-only tape containing independent binary random numbers (0 or 1 with equal probability), and which halts with probability one and outputs a single element of $\mathcal{M}$. The law of such a machine will be denoted by $\mathbb{P}$. With an abuse of notation, we sometimes denote the probabilistic Turing machine by $(\Gamma, \mathbb{P})$.*

**Remark 3.9** (Where does the output live?). Strictly speaking, when we say that the output of a probabilistic Turing machine lies in $\mathcal{M}$, we mean that the output corresponds, via an encoding, to an element of a subset of $\mathcal{M}$ such as $(\mathbb{Q} + i\mathbb{Q})^N \subset \mathbb{C}^N$. However, we follow the usual convention of suppressing such encodings. ⊠

One should think of Definition 3.8 as an algorithm for the computational problem with inexact input, but with the additional ability to generate random numbers (corresponding to the binary input tape) and execute commands based on the sequence of random numbers that are generated. The reader should intuitively think of this as a computer program with a random number generator. For equivalent definitions and the basic properties of such machines, see [7]. For simplicity, we have only considered probabilistic Turing machines that halt with probability one, though extensions can be made to non-halting machines. Note that Definition 3.8 is a special case of Definition 3.4, where $\mathbb{P}_\iota = \mathbb{P}$ is fixed across different $\iota$. In particular, given a probabilistic Turing machine, the sigma-algebra and probability distribution generated by the standard product topology on $\{0,1\}^\mathbb{N}$ induce the relevant collection $X$ of Turing machines and sigma-algebra $\mathcal{F}$, as well as $\mathbb{P}$.

Finally, we recall standard definitions of condition used in optimisation [10, 12]. The classical condition number of an invertible matrix $A$ is given by $\mathrm{Cond}(A) = \|A\|\|A^{-1}\|$. For different types of condition numbers related to a possibly multivalued (signified by the double arrow) mapping $\Xi : \Omega \subset \mathbb{C}^n \rightrightarrows \mathbb{C}^N$ we need to establish what types of perturbations we are interested in. For example, if $\Omega$ denotes the set of diagonal matrices (which we treat as elements of $\mathbb{C}^n$ for some $n$), we may not be interested in perturbations in the off-diagonal elements as they will always be zero. In particular, we may only be interested in perturbations in the coordinates that are varying in the set $\Omega$. Thus, given $\Omega \subset \mathbb{C}^n$ we define the active coordinates of $\Omega$ to

be $\text{Act} = \text{Act}(\Omega) = \{j : \exists x, y \in \Omega, x_j \neq y_j\}$. Moreover, for $\nu > 0$ (including the obvious extension to $\nu = \infty$),

$$\Omega_\nu = \{x : \exists y \in \Omega \text{ such that } \|x - y\|_{l^\infty} \leq \nu, x_{\text{Act}^c} = y_{\text{Act}^c}\}.$$

In other words, $\Omega_\nu$ is the set of $\nu$-perturbations along the non-constant coordinates of elements in $\Omega$. We can now recall some of the classical condition numbers from the literature [12, 10].

(1) *Condition of a mapping*: Let $\Xi : \Omega \subset \mathbb{C}^n \rightrightarrows \mathbb{C}^m$ be a linear or non-linear mapping, and suppose that $\Xi$ is also defined on $\Omega_\nu$ for some $\nu > 0$. Then,

$$\text{Cond}(\Xi, \Omega) = \sup_{x \in \Omega} \lim_{\epsilon \to 0^+} \sup_{\substack{x+z \in \Omega_\nu \\ 0 < \|z\|_{l^2} \leq \epsilon}} \left\{ \frac{\text{dist}(\Xi(x+z), \Xi(x))}{\|z\|_{l^2}} \right\},$$

where we allow for multivalued functions by defining

$$\text{dist}(\Xi(x+z), \Xi(z)) = \inf_{w_1 \in \Xi(x+z), w_2 \in \Xi(z)} \|w_1 - w_2\|_{l^2}$$

(see Remark 3.2). We will use this notion of condition number for $(P_1)$, $(P_2)$ and $(P_3)$.

(2) *Distance to infeasibility - the Feasibility Primal condition number*: For the problem $(P_1)$ of basis pursuit (for $(P_2)$ and $(P_3)$ the following condition number is always zero) we set

$$\nu(A, y) = \sup\left\{\epsilon \geq 0 : \|\hat{y}\|_{l^2}, \|\hat{A}\| \leq \epsilon, (A+\hat{A}, y+\hat{y}) \in \Omega_\infty \Rightarrow (A+\hat{A}, y+\hat{y}) \text{ are feasible inputs to } \Xi_{P_1}\right\},$$

and define the *Feasibility Primal* (FP) local condition number $C_{\text{FP}}(A, y) := \frac{\max\{\|y\|_{l^2}, \|A\|\}}{\nu(A, y)}$. We then define the FP global condition number via $C_{\text{FP}}(\Xi_{P_1}, \Omega) := \sup_{(A,y) \in \Omega} C_{\text{FP}}(A, y)$.

3.2. **Phase transitions.** To prove Theorem 2, we use the following lemmas which describe phase transitions of the minimisers of the respective optimisation problems ($e_j$ correspond to the canonical basis of $\mathbb{C}^N$).

**Lemma 3.10** (**Phase transition for basis pursuit**). *Let $N \geq 2$ and consider the problem $(P_1)$ for*

$$A = \begin{pmatrix} \frac{w_1}{\rho_1} & \frac{w_2}{\rho_2} & \cdots & \frac{w_N}{\rho_N} \end{pmatrix} \in \mathbb{C}^{1 \times N}, \quad y = 1, \quad \epsilon \in [0, 1),$$

*where $\rho_j > 0$ for $j = 1, ..., N$. Then the set of solutions is given by*

$$\sum_{j=1}^N \left[t_j(1-\epsilon)\frac{\rho_j}{w_j}\right] e_j, \quad s.t. \quad t_j \in [0, 1], \sum_{j=1}^N t_j = 1 \quad \text{and} \quad t_j = 0 \quad \text{if} \quad \rho_j > \min_k \rho_k. \quad (3.1)$$

*Proof.* Let $\hat{x}_j = x_j w_j \rho_j^{-1}$, then the optimisation problem becomes

$$\text{argmin}_{\hat{x} \in \mathbb{C}^N} f(\hat{x}) := \sum_{j=1}^N \rho_j |\hat{x}_j| \text{ such that } \left|1 - \sum_{j=1}^N \hat{x}_j\right| \leq \epsilon. \quad (3.2)$$

Since $\epsilon < 1$ and the $(\rho_1, \rho_2, ...\rho_N)$ weighted $l^1$ norm is convex, it follows that the solution must lie on the hypersurface segment $\hat{x}_1 + \hat{x}_2 + ... + \hat{x}_j = 1 - \epsilon$ for $\hat{x}_j \in \mathbb{R}_{\geq 0}$. We now claim that if $\hat{x}$ is a solution of (3.2), and $\rho_j > \min_k \rho_k$, then $\hat{x}_j = 0$. Suppose for a contradiction that there exists a solution $\hat{x}$ of (3.2) where $\hat{x}_j > 0$ and $\rho_j > \min_k \rho_k$. Pick any $l$ such that $\rho_l = \min_k \rho_k$, then $\hat{x} + \hat{x}_j(e_l - e_j)$ is feasible with $f(\hat{x} + \hat{x}_j(e_l - e_j)) < f(\hat{x})$, a contradiction. Similarly, if $x$ is of the form given in (3.1), then $f(\hat{x}) = (1-\epsilon) \min_k \rho_k$. In particular, the objective function is constant over the set of all such vectors and the result follows. □

**Lemma 3.11** (**Phase transition for LASSO**). *Let $N \geq 2$ and consider the problem $(P_2)$ for*

$$A = \lambda \begin{pmatrix} \frac{w_1}{\rho_1} & \frac{w_2}{\rho_2} & \cdots & \frac{w_N}{\rho_N} \end{pmatrix} \in \mathbb{C}^{1 \times N}, \quad y = 1,$$

*where $0 < \rho_j < 2$ for $j = 1, ..., N$. Then the set of solutions is given by*

$$\left(1 - \frac{\min_k \rho_k}{2}\right) \sum_{j=1}^N \frac{\rho_j t_j}{\lambda w_j} e_j, \quad s.t. \quad t_j \in [0, 1], \sum_{j=1}^N t_j = 1 \quad \text{and} \quad t_j = 0 \quad \text{if} \quad \rho_j > \min_k \rho_k. \quad (3.3)$$

*Proof.* Let $\hat{x}_j = x_j \lambda w_j \rho_j^{-1}$, then the optimisation problem becomes $\operatorname{argmin}_{\hat{x} \in \mathbb{C}^N} f(\hat{x}) := |1 - \sum_{j=1}^N \hat{x}_j|^2 + \sum_{j=1}^N \rho_j |\hat{x}_j|$. It is clear that any optimal solution must be real and hence we restrict our argument to real $\hat{x}$. Define the $2^N$ quadrant subdomains $D_{k_1,\ldots,k_N} = \{\hat{x}_j \cdot (-1)^{k_j} > 0\}$ for $k_j \in \{0, 1\}$, and notice that

$$\nabla f(\hat{x}) = \begin{bmatrix} -2(1 - \sum_{j=1}^N \hat{x}_j) + (-1)^{k_1} \rho_1 \\ \vdots \\ -2(1 - \sum_{j=1}^N \hat{x}_j) + (-1)^{k_N} \rho_N \end{bmatrix}, \quad \text{for } \hat{x} \in D_{k_1,\ldots,k_N}.$$

We first look for stationary points of the objective function in the subdomains $D_{k_1,\ldots,k_N}$. The condition for a stationary point in the interior of such a domain leads to the constraint that $k_1 = k_2 = \ldots = k_N$. If $k_1 = k_2 = \ldots = k_N = 1$, then $\nabla f = 0$ leads to the contradiction $\rho_j = 2(\hat{x}_1 + \ldots + \hat{x}_N) - 2 < 0$. Finally, in the case (and only in the case) of $\rho_1 = \rho_2 = \ldots = \rho_N$, there is a hypersurface segment of stationary points in $D_{0,0,\ldots,0}$ given by $\hat{x}_1 + \ldots + \hat{x}_N = 1 - \rho_1/2$ (recall that we assumed $\rho_1 < 2$ so this segment exists).

First, consider the case that $\rho_1 = \ldots = \rho_N$. Then any optimal solution must either lie on the boundary of some $D_{k_1,\ldots,k_N}$ or on the hypersurface segment $\hat{x}_1 + \ldots + \hat{x}_N = 1 - \rho_1/2$ in $D_{0,0,\ldots,0}$. A simple case by case analysis now yields that the solutions $\hat{x}$ are given by convex combinations of $(1 - \rho_j/2)e_j$ for $j = 1, \ldots, N$. Now consider the case that not all of the $\rho_j$ are equal. Then any optimal solution must lie on the boundary of some $D_{k_1,\ldots,k_N}$. A simple case by case analysis now yields that the solutions $\hat{x}$ are given by convex combinations of $(1 - \rho_j/2)e_j$ for $j$ such that $\rho_j = \min_k \rho_k$. Rescaling back to $x$ gives the result. $\square$

**Lemma 3.12 (Phase transition for square-root LASSO).** *Let $N \geq 2$ and consider the problem $(P_3)$ for*

$$A = \lambda \begin{pmatrix} \frac{w_1}{\rho_1} & \frac{w_2}{\rho_2} & \cdots & \frac{w_N}{\rho_N} \end{pmatrix} \in \mathbb{C}^{1 \times N}, \quad y = 1,$$

*where $0 < \rho_j < 1$ for $j = 1, \ldots, N$. Then the set of solutions is given by*

$$\sum_{j=1}^N \frac{\rho_j t_j}{\lambda w_j} e_j, \quad \text{s.t.} \quad t_j \in [0,1], \sum_{j=1}^N t_j = 1 \quad \text{and} \quad t_j = 0 \quad \text{if} \quad \rho_j > \min_k \rho_k. \tag{3.4}$$

*Proof.* Let $\hat{x}_j = x_j \lambda w_j \rho_j^{-1}$, then the optimisation problem becomes $\operatorname{argmin}_{\hat{x} \in \mathbb{C}^N} f(\hat{x}) := |1 - \sum_{j=1}^N \hat{x}_j| + \sum_{j=1}^N \rho_j |\hat{x}_j|$. It is clear that any optimal solution must be real and hence we restrict our argument to real $\hat{x}$. The objective function is piecewise affine and since $\rho_j < 1$, the gradient of $f$ is non-vanishing on the interior of any of the domains $D_{k_1,\ldots,k_N} = \{\hat{x}_j \cdot (-1)^{k_j} > 0\}$ for $k_j \in \{0, 1\}$. It follows that the optimal solutions must lie on the boundaries of the domains $D_{k_1,\ldots,k_N}$. A simple case by case analysis shows that the solutions $\hat{x}$ are given by convex combinations of $e_j$ for $j$ such that $\rho_j = \min_k \rho_k$. Rescaling back to $x$ gives the result. $\square$

We will also need the following propositions, which give useful criteria for impossibility results.

**Proposition 3.13.** *Let $\{\Xi, \Omega, \mathcal{M}, \Lambda\}$ be a computational problem. Suppose that there are two sequences $\{\iota_n^1\}_{n \in \mathbb{N}}, \{\iota_n^2\}_{n \in \mathbb{N}} \subset \Omega$ satisfying the following conditions:*

(a) *There are sets $S^1, S^2 \subset \mathcal{M}$ and $\kappa > 0$ such that $\inf_{x_1 \in S^1, x_2 \in S^2} d(x_1, x_2) > \kappa$ and $\Xi(\iota_n^j) \subset S^j$ for $j = 1, 2$.*

(b) *For every $f \in \Lambda$ there is a $c_f \in \mathbb{C}$ such that $|f(\iota_n^j) - c_f| \leq 1/4^n$ for all $n \in \mathbb{N}$ and $j = 1, 2$.*

*Then, if we consider $\{\Xi, \Omega, \mathcal{M}, \Lambda\}^{\Delta_1}$, we have the following:*

(i) *For any sequential general algorithm $\Gamma$ and $M \in \mathbb{N}$, there exists $\iota \in \Omega$ and $\widehat{\Lambda} \in \mathcal{L}^1(\Lambda)$ such that*

$$\operatorname{dist}(\Gamma(\iota), \Xi(\iota)) > \kappa/2 \quad \text{or} \quad S_\Gamma(\iota) > M.$$

(ii) *If there is an $\iota^0 \in \Omega$ such that for every $f \in \Lambda$ we have that (b) is satisfied with $c_f = f(\iota^0)$, then for any RGA $\Gamma$ and $p \in [0, 1/2)$, there exists $\iota \in \Omega$ and $\widehat{\Lambda} \in \mathcal{L}^1(\Lambda)$ such that $\mathbb{P}_\iota(\operatorname{dist}(\Gamma(\iota), \Xi(\iota)) \geq \kappa/2) > p$.*

*Proof.* Without loss of generality, we assume that $\Omega = \{\iota_n^1\}_{n \in \mathbb{N}} \cup \{\iota_n^2\}_{n \in \mathbb{N}}$. Part (ii) follows immediately from a Proposition of [8], so we only prove part (i). Let $\Gamma$ be a sequential general algorithm and $M \in \mathbb{N}$. We will construct the required $\widehat{\Lambda} \in \mathcal{L}^1(\Lambda)$ inductively. By Definition 3.3 and the setup of $\Delta_1$-information for $\Lambda$, there exists some $f_{k_1} \in \Lambda$ and $n_1 \in \mathbb{N}$ such that for all $\iota \in \Omega$ and for all $\widehat{\Lambda} \in \mathcal{L}^1(\Lambda)$, we have $f_{\iota,1}^\Gamma = f_{k_1,n_1}$. We set $f_{k_1,n_1}(\iota_m^j) = c_{f_{k_1}}$ for all $m \geq n_1$ and choose $f_{k_1,n_1}(\iota_m^j)$ consistently for $m < n_1$. Again by Definition 3.3 and the setup of $\Delta_1$-information for $\Lambda$, it follows that there exists $f_{k_2} \in \Lambda$ and $n_2 \in \mathbb{N}$ (which without loss of generality $\geq n_1$) such that for all $m \geq n_1$, either $\Lambda_\Gamma(\iota_m^j) = \{f_{k_1,n_1}\}$ or $f_{\iota_m^j,2}^\Gamma = f_{k_2,n_2}$. In the latter case, we set $f_{k_2,n_2}(\iota_m^j) = c_{f_{k_2}}$ for all $m \geq n_2$ and choose $f_{k_2,n_2}(\iota_m^j)$ consistently for $m < n_2$. We continue this process for a maximum of $M$ steps up to $f_{\iota_m^j, \min\{M, |\Lambda_\Gamma(\iota_m^j)|\}}^\Gamma$ as follows. At the $q$th step after defining $f_{k_q,n_q}$, by Definition 3.3 and the setup of $\Delta_1$-information for $\Lambda$, it follows that there exists $f_{k_{q+1}} \in \Lambda$ and $n_{q+1} \in \mathbb{N}$ (which without loss of generality $\geq n_q$) such that for all $m \geq n_q$, either $\Lambda_\Gamma(\iota_m^j) \subset \{f_{k_1,n_1}, ..., f_{k_q,n_q}\}$ or $f_{\iota_m^j,q+1}^\Gamma = f_{k_{q+1},n_{q+1}}$. In the latter case, we set $f_{k_{q+1},n_{q+1}}(\iota_m^j) = c_{f_{k_{q+1}}}$ for all $m \geq n_{q+1}$ and choose $f_{k_{q+1},n_{q+1}}(\iota_m^j)$ consistently for $m < n_{q+1}$. We can then choose the rest of the function values to obtain $\widehat{\Lambda}$.

Given this $\widehat{\Lambda} \in \mathcal{L}^1(\Lambda)$, suppose for a contradiction that for any $\iota \in \Omega$, $\text{dist}(\Gamma(\iota), \Xi(\iota)) \leq \kappa/2$ and $S_\Gamma(\iota) \leq M$. Without loss of generality, we assume that the above construction is carried out for $M$ steps. It follows that we must have $f(\iota_{n_M}^1) = f(\iota_{n_M}^2)$ for all $f \in \widehat{\Lambda}_\Gamma(\iota_{n_M}^1)$. By (ii) and (iii) of Definition 3.3, it follows that $\Gamma(\iota_{n_M}^1) = \Gamma(\iota_{n_M}^2)$. Let $\epsilon > 0$ be arbitrary. Since $\text{dist}(\Gamma(\iota), \Xi(\iota)) \leq \kappa/2$ for all $\iota \in \Omega$, there exists $s_j \in S^j$ such that $d(\Gamma(\iota_{n_M}^j), s_j) < \kappa/2 + \epsilon$. It follows that

$$\inf_{x_1 \in S^1, x_2 \in S^2} d(x_1, x_2) \leq d(s_1, s_2) \leq d(\Gamma(\iota_{n_M}^1), s_1) + d(\Gamma(\iota_{n_M}^2), s_2) < \kappa + 2\epsilon.$$

Since $\epsilon > 0$ was arbitrary, we have $\inf_{x_1 \in S^1, x_2 \in S^2} d(x_1, x_2) \leq \kappa$, the required contradiction. □

**Proposition 3.14.** *Let $\{\Xi, \Omega, \mathcal{M}, \Lambda\}$ be a computational problem and $u \geq 2$ be a positive integer. Suppose that there are $u$ sequences $\{\iota_n^j\}_{n \in \mathbb{N}} \subset \Omega$, for $j = 1, ..., u$, satisfying the following conditions:*

*(a) There are sets $S^j \subset \mathcal{M}$, for $j = 1, ..., u$, and $\kappa > 0$ such that $\inf_{x_j \in S^j, x_k \in S^k} d(x_j, x_k) > \kappa$ for any $j \neq k$ and $\Xi(\iota_n^j) \subset S^j$ for $j = 1, ..., u$.*

*(b) For every $f \in \Lambda$, there is a $c_f \in \mathbb{C}$ such that $|f(\iota_n^j) - c_f| \leq 1/4^n$ for all $n \in \mathbb{N}$ and $j = 1, ..., u$.*

*Then, if we consider $\{\Xi, \Omega, \mathcal{M}, \Lambda\}^{\Delta_1}$, for any halting probabilistic Turing machine $(\Gamma, \mathbb{P})$, $M \in \mathbb{N}$ and $p \in [0, \frac{u-1}{u})$, there exists $\iota \in \Omega$ and $\widehat{\Lambda} \in \mathcal{L}^1(\Lambda)$ such that $\mathbb{P}\Big(\text{dist}(\Gamma(\iota), \Xi(\iota)) > \kappa/2 \quad \text{or} \quad S_\Gamma(\iota) > M\Big) > p$.*

*Proof.* Without loss of generality, we can assume that $\Omega = \cup_{j=1}^u \{\iota_n^j\}_{n \in \mathbb{N}}$. Let $(\Gamma, \mathbb{P})$ be a halting probabilistic Turing machine and $M \in \mathbb{N}$, $p \in [0, (u-1)/u)$. Suppose for a contradiction that for all $\iota \in \Omega$ and all $\widehat{\Lambda} \in \mathcal{L}^1(\Lambda)$, we have $\mathbb{P}(\text{dist}(\Gamma(\iota), \Xi(\iota)) > \kappa/2$ or $S_\Gamma(\iota) > M) \leq p$. We will construct the required $\widehat{\Lambda} \in \mathcal{L}^1(\Lambda)$ inductively. Let $\beta \in (0, 1)$ be such that $(1-\beta)^M - p > 1/u$. Such a $\beta$ exists since $p \in [0, (u-1)/u)$. Since the Turing machine must halt with probability one, there exists finite sets $K_1, N_1 \subset \mathbb{N}$ such that with probability (w.r.t. $\mathbb{P}$) at least $1 - \beta$, for all $\iota \in \Omega$ and for all $\widehat{\Lambda} \in \mathcal{L}^1(\Lambda)$, it holds that $f_{\iota,1}^\Gamma = f_{k_1,n_1}$ for some $k_1 \in K_1$ and $n_1 \in N_1$. We set $f_{k_1,n_1}(\iota_m^j) = c_{f_{k_1}}$ for all $m \geq \max\{n_1 : n_1 \in N_1\}$ and choose $f_{k_1,n_1}(\iota_m^j)$ consistently otherwise.

We continue this process inductively for $M$ steps up to $f_{\iota_m^j, \min\{M, |\Lambda_\Gamma(\iota_m^j)|\}}^\Gamma$ as follows. At the $q$th step after defining $f_{k_q,n_q}$ for $k_q \in K_q$ and $n_q \in N_q$, it follows (since the Turing machine halts with probability one) that there exists finite sets $K_{q+1}, N_{q+1} \subset \mathbb{N}$ with the following property. Let $E_{q+1}$ be the event that for all $\iota_m^j \in \Omega$ with $m \geq \max\{n : n \in N_1 \cup ... \cup N_q\}$, either $f_{\iota_m^j, q+1}^\Gamma = f_{k_{q+1},n_{q+1}}$, for some $k_{q+1} \in K_{q+1}$ and $n_{q+1} \in N_{q+1}$, or $|\Lambda(\iota_m^j)| \leq q$. Then $\mathbb{P}(E_{q+1} | \cap_{k \leq q} E_k) \geq 1 - \beta$. We then set $f_{k_{q+1},n_{q+1}}(\iota_m^j) = c_{f_{k_{q+1}}}$ for all $m \geq \max\{n : n \in N_1 \cup ... \cup N_{q+1}\}$ and choose $f_{k_{q+1},n_{q+1}}(\iota_m^j)$ consistently otherwise. This ensures the existence of $K_{q+2}$ and $N_{q+2}$. After the $M$th step, we can choose the rest of the function values to obtain $\widehat{\Lambda}$.

It follows that for $m \geq \max\{n : n \in N_1 \cup ... \cup N_M\}$, the outputs $\Gamma(\iota_m^j)$ conditional on the event $E_1 \cap ... \cap E_M \cap \{S_\Gamma(\cdot) \leq M\}$ are equal for $j = 1, ..., u$. Since $\inf_{x_j \in S^j, x_k \in S^k} d(x_1, x_2) > \kappa$ for $j \neq k$, it follows that the events $F_j := \{\text{dist}(\Gamma(\iota_m^j), \Xi(\iota_m^j)) \leq \kappa/2\} \cap \{S_\Gamma(\iota_m^j) \leq M\} \cap E_1 \cap ... \cap E_M$, $j = 1, ..., u$,

are disjoint. Moreover, using the fact that $\mathbb{P}(A \cap B) = \mathbb{P}(A) + \mathbb{P}(B) - \mathbb{P}(A \cup B)$,

$$\mathbb{P}(F_j) \geq \mathbb{P}(\{\operatorname{dist}(\Gamma(\iota_m^j), \Xi(\iota_m^j)) \leq \kappa/2\} \cap \{S_\Gamma(\iota_m^j) \leq M\}) + \mathbb{P}(E_1 \cap ... \cap E_M) - 1$$
$$\geq \mathbb{P}(\{\operatorname{dist}(\Gamma(\iota_m^j), \Xi(\iota_m^j)) \leq \kappa/2\} \cap \{S_\Gamma(\iota_m^j) \leq M\}) + (1-\beta)^M - 1 \geq (1-\beta)^M - p > 1/u.$$

But this contradicts the disjointness of the $F_j$'s. □

3.3. **Proof of Theorem 2.**

*Proof of Theorem 2.* We will argue for $m = 1$ and construct such an $\Omega$ for this case. The general case of $m > 1$ follows by embedding our construction for $A \in \mathbb{C}^{1 \times (N+1-m)}$ in the first row of matrices and vectors of the form

$$\hat{A} = \begin{pmatrix} A & 0 \\ 0 & \alpha I \end{pmatrix}, \quad \hat{y} = (y, 0)^\top, \quad (A, y) \in \Omega,$$

where $I \in \mathbb{C}^{(m-1) \times (m-1)}$ denotes the $(m-1) \times (m-1)$ identity matrix and $\alpha = \alpha(A)$ is chosen such that $\hat{A}\hat{A}^*$ is a multiple of the identity. In particular, such an embedding does not effect the relevant condition numbers (it is straightforward to see that the matrix norm, distance to infeasibility for $(P_1)$ and condition numbers of the mappings are all unchanged). For the classes we consider, the setup of Theorem 2 coincides with the $\Delta_1$-information model discussed in §3.3.1. In particular, we can use Lemmas 3.10, 3.11 and 3.12 to derive the relevant $x_{s,n}$'s in (1.7). This means that we can apply Propositions 3.13 and 3.14 with the metric corresponding to the $l^2$-norm. Recall that for this theorem, we assume that $w_1 = w_2 = ... = w_N = 1$.

**Step 1:** Proof for $(P_1)$. First, consider the class defined by

$$\Omega_1 = \left\{ (A(\gamma_1; \rho), y) : A(\gamma_1; \rho) := \gamma_1 \begin{pmatrix} \frac{1}{\rho_1} & \frac{1}{\rho_2} & \cdots & \frac{1}{\rho_N} \end{pmatrix}, y = 1, \rho_j \in [1 - 2\delta, 1 - \delta] \right\},$$

for fixed $\gamma_1 > 10$ and $\delta \in (0, 1/4)$. We choose $\gamma_1$ and $\delta$ such that

$$\frac{1-\epsilon}{\gamma_1} \cdot \sup_{\rho_j, \rho_k \in [1-2\delta, 1-\delta], j \neq k} \|\rho_j e_j - \rho_k e_k\|_{l^2} = 3 \cdot 10^{-K}, \tag{3.5}$$

$$\frac{1-\epsilon}{\gamma_1} \cdot \inf_{\rho_j, \rho_k \in [1-2\delta, 1-\delta], j \neq k} \|\rho_j e_j - \rho_k e_k\|_{l^2} > 2 \cdot 10^{-K}, \tag{3.6}$$

where the $e_j$ denote the canonical basis of $\mathbb{C}^N$. Note that we can ensure $\gamma_1 > 10$ since $\epsilon \leq 1/2$ and $K > 2$. If $\rho_j \in [1-2\delta, 1-\delta)$, for $j = 1, 2$, then by (a simple rescale of) Lemma 3.10,

$$\Xi_{P_1}(A(\gamma_1; (\rho_1, 1-\delta, ..., 1-\delta)), 1) = \frac{(1-\epsilon)\rho_1}{\gamma_1} e_1, \Xi_{P_1}(A(\gamma_1; (1-\delta, \rho_2, 1-\delta, ..., 1-\delta)), 1) = \frac{(1-\epsilon)\rho_2}{\gamma_1} e_2.$$

Since (3.6) holds, it follows by selecting appropriate sequences $\iota_n^j$ for choices of $\rho_j = \rho_j^n \uparrow 1 - \delta$ that the conditions of Proposition 3.13 hold for $\Omega_1$ with

$$S^j = \left\{ \frac{1-\epsilon}{\gamma_1} \rho e_j : \rho \in [1-2\delta, 1-\delta] \right\}, \quad \kappa = 2 \cdot 10^{-K}. \tag{3.7}$$

Moreover, the condition for part (ii) of Proposition 3.13 also holds with $\iota^0 = (A(\gamma_1; (1-\delta, 1-\delta, ..., 1-\delta)), 1)$.

Now suppose for a contradiction that there exists a (halting) RGA (with input $\iota_{A,S}$) and $p > 1/2$ that produces a NN $\phi_A$ such that $\min_{y \in S_A} \inf_{x^* \in \Xi_{P_1}(A,y)} \|\phi_A(y) - x^*\|_{l^2} \leq 10^{-K}$ holds with probability at least $p$ for all $(A, y) \in \Omega_1$. Then there exists a (halting) RGA, $\Gamma$, taking $\iota_{A,S}$ as input that computes a solution of $(P_1)$ to $K$ correct digits with probability at least $p$ on each input in $\Omega_1$. However, this contradicts Proposition 3.13 (ii). Next, consider the class defined by

$$\Omega_2 = \left\{ (A(\gamma_2; \rho), y) : A(\gamma_2; \rho) = \gamma_2 \begin{pmatrix} \frac{1}{\rho_1} & \frac{1}{\rho_2} & \cdots & \frac{1}{\rho_N} \end{pmatrix}, y = 1, \rho_j \in [1-2\delta, 1-\delta], \rho_j \neq \rho_k \text{ if } j \neq k \right\},$$

where $\gamma_2 = \gamma_1/10 > 1$ so that

$$\sup_{\rho_j,\rho_k \in [1-2\delta,1-\delta], j \neq k} \|\rho_j e_j - \rho_k e_k\|_{l^2} = 3 \cdot 10^{-K+1} \cdot \frac{\gamma_2}{1-\epsilon}, \quad (3.8)$$

$$\inf_{\rho_j,\rho_k \in [1-2\delta,1-\delta], j \neq k} \|\rho_j e_j - \rho_k e_k\|_{l^2} > 2 \cdot 10^{-K+1} \cdot \frac{\gamma_2}{1-\epsilon}. \quad (3.9)$$

By extending the argument above to $u = N + 1 - m = N$ (recall without loss of generality that $m = 1$) sequences and sets $S^j$ defined as in (3.7), the conditions of Proposition 3.14 hold with $\kappa = 2 \cdot 10^{-K+1}$. Now suppose that there exists a (halting) probabilistic Turing machine $(\Gamma, \mathbb{P})$, $M \in \mathbb{N}$ and $p \in \left[0, \frac{N-m}{N+1-m}\right)$, such that for any $(A, 1) \in \Omega_2$, $\Gamma$ computes a NN $\phi_A$ with

$$\mathbb{P}\left(\inf_{x^* \in \Xi_{P_1}(A,y)} \|\phi_A(y) - x^*\|_{l^2} > 10^{1-K} \text{ or the sample size needed to construct } \phi_A > M\right) \leq p.$$

Then there exists a (halting) probabilistic Turing machine that computes a solution of $(P_1)$ to $K - 1$ correct digits on each input in $\Omega_2$ with sample size at most $M$ with probability at least $1 - p$. However, this contradicts Proposition 3.14.

We now set $\Omega = \Omega_1 \cup \Omega_2$. Note that the negative statements of part (i) and (ii) follow from the above arguments by considering restrictions to $\Omega_1$ and $\Omega_2$ respectively. Hence, we are left with proving the condition number bounds, part (iii) and the positive part of part (ii). First, note that $\text{Cond}(AA^*) = 1$ for any $(A, y) \in \Omega$. For any $(A, y) \in \Omega$, we have $\nu(A, y) = \|A\| \geq 1 = \|y\|_{l^2}$ and hence $C_{\text{FP}}(\Xi_{P_1}, \Omega) \leq 1$. To bound the final condition number, first note that if $\rho_j, \rho'_j \leq 1$, then $\|\rho - \rho'\|_{l^2} \leq \|\sum_{j=1}^N (\frac{1}{\rho_j} - \frac{1}{\rho'_j})e_j\|_{l^2}$. Let $(A(\gamma_1; \rho), 1) \in \Omega_1$, then if $(A(\gamma_1; \rho'), 1) \in \Omega_1$ with $\Delta(\rho, \rho') := \gamma_1 \|\sum_{j=1}^N (\frac{1}{\rho_j} - \frac{1}{\rho'_j})e_j\|_{l^2}$ sufficiently small,

$$\text{dist}(\Xi_{P_1}(A(\gamma_1; \rho'), 1), \Xi_{P_1}(A(\gamma_1; \rho), 1)) \leq \frac{1-\epsilon}{\gamma_1}\|\rho - \rho'\|_{l^2}.$$

It follows that

$$\lim_{\beta \downarrow 0} \sup_{\substack{(A(\gamma_1;\rho'),1) \in \Omega_1 \\ \Delta(\rho,\rho') \leq \beta}} \frac{\text{dist}(\Xi_{P_1}(A(\gamma_1;\rho'),1), \Xi_{P_1}(A(\gamma_1;\rho),1))}{\Delta(\rho,\rho')} \leq \frac{1-\epsilon}{\gamma_1^2} < 1.$$

A similar argument holds for $(A(\gamma_2; \rho), 1) \in \Omega_2$, and hence $\text{Cond}(\Xi_{P_1}, \Omega) \leq 1$.

We now prove the positive parts of (ii) and (iii). We begin with (ii) and describe the algorithm informally, noting that the output of the algorithm, $\Gamma(A)$, yields a NN which maps $y = 1$ to $\Gamma(A) \in \mathbb{C}^N$. Given an input $(A, y) \in \Omega$, the algorithm first tests the size of $A_{1,1}$ to determine whether $(A, y) \in \Omega_1$ or $(A, y) \in \Omega_2$. Explicitly, we note that $A_{1,1}$ is positive and bounded away from 0. Hence, with one sample from $\iota_{A,S}$ we can determine $A_{1,1}$ to an accuracy of at least $0.01 \cdot A_{1,1}$ and such that, simultaneously, the corresponding approximation of $A_{1,1}^{-1}$ is accurate to at least $10^{-K}$. If $(A, y) \in \Omega_1$, then $A_{1,1} \in \gamma_1 \cdot [1, 2]$ whereas if $(A, y) \in \Omega_2$, then $A_{1,1} \in \gamma_2 \cdot [1, 2]$. Since $\gamma_2 = \gamma_1/10$, this level of accuracy is enough to determine whether $(A, y) \in \Omega_1$ or $(A, y) \in \Omega_2$. Next, if the algorithm determines $(A, y) \in \Omega_1$, it outputs the corresponding approximation of $(1 - \epsilon)A_{1,1}^{-1}e_1 = \frac{1-\epsilon}{\gamma_1}\rho_1 e_1$ correct to $10^{-K}$ in the $l^2$-norm from the sample. Since

$$\sup_{\rho_j,\rho_k \in [1-2\delta,1-\delta], j \neq k} \|\rho_j e_j - \rho_k e_k\|_{l^2} = 3 \cdot 10^{-K} \cdot \frac{\gamma_1}{1-\epsilon},$$

it follows that $\inf_{x^* \in \Xi_{P_1}(A,y)} \|\Gamma(A) - x^*\|_{l^2} \leq 4 \cdot 10^{-K} < 10^{-K+1}$. On the other hand, if the algorithm determines $(A, y) = (A(\gamma_2; \rho), y) \in \Omega_2$, then we know that all of the $\rho_j$ are distinct. The algorithm continues to sample $\iota_{A,S}$ until we determine $j$ such that $\rho_j = \min_k \rho_k$. It then outputs an approximation of $(1 - \epsilon)A_{1,j}^{-1}e_j = \Xi_{P_1}(A, y)$ correct to $10^{-K}$ in the $l^2$-norm. Such as approximation can be computed using $\iota_{A,S}$. It then follows that $\|\Gamma(A) - \Xi_{P_1}(A, y)\|_{l^2} \leq 10^{-K} < 10^{-K+1}$ and this finishes the proof of (ii). Finally, to prove (iii), note that the arguments above show that, given an input $(A, y) \in \Omega$, we can use one sample

($L = 1$) of $\iota_{A,S}$ to compute an approximation of $A_{1,1}^{-1}$ with error bounded by $10^{-K}$. We simply set $\Gamma(A)$ to be $(1-\epsilon)e_1$ multiplied by the approximation of $A_{1,1}^{-1}$. Using (3.5) and (3.8), it follows that

$$\inf_{x^* \in \Xi_{P_1}(A,y)} \|\Gamma(A) - x^*\|_{l^2} \leq 3 \cdot 10^{-K+1} + 10^{-K} < 10^{-K+2}.$$

**Step 2:** Proof for ($P_2$). This is almost identical step 1 with replacing Lemma 3.10 with Lemma 3.11. The other changes are replacing $\epsilon$ with the suitable $\rho_i/2$ in the solution of each LASSO problem, including the additional scale $\lambda$ in the definition of the matrices (see Lemma 3.11) and choosing $\gamma_1 \lambda > 10$ and $\delta \in (0, 1/4)$ such that (recall that $\lambda \leq 1$)

$$\sup_{\rho_j, \rho_k \in [1-2\delta, 1-\delta], j \neq k} \frac{1}{\lambda \gamma_1} \cdot \|\rho_j(1-\rho_j/2)e_j - \rho_k(1-\rho_k/2)e_k\|_{l^2} = 3 \cdot 10^{-K} \tag{3.10}$$

$$\inf_{\rho_j, \rho_k \in [1-2\delta, 1-\delta], j \neq k} \frac{1}{\lambda \gamma_1} \cdot \|\rho_j(1-\rho_j/2)e_j - \rho_k(1-\rho_k/2)e_k\|_{l^2} > 2 \cdot 10^{-K}. \tag{3.11}$$

Let $f(x) = (1 - x/2)x$, then for $x \in [0,1]$, $|f'(x)| \leq 1$. It follows that

$$\left\|\sum_{j=1}^{N} \left(\rho_j(1-\rho_j/2) - \rho'_j(1-\rho'_j/2)\right) e_j\right\|_{l^2} \leq \left\|\sum_{j=1}^{N} \left(\frac{1}{\rho_j} - \frac{1}{\rho'_j}\right) e_j\right\|_{l^2}.$$

Hence, for sufficiently small $\delta$, for any $(A(\gamma_1; \rho), 1) \in \Omega_1$ (recall the additional factor of $\lambda$) and $\rho'$ sufficiently close to $\rho$ with $(A(\gamma_1; \rho'), 1) \in \Omega_1$,

$$\frac{\mathrm{dist}(\Xi_{P_2}(A(\gamma_1; \rho'), 1), \Xi_{P_2}(A(\gamma_1; \rho), 1))}{\gamma_1 \lambda \left\|\sum_{j=1}^{N} \left(\frac{1}{\rho_j} - \frac{1}{\rho'_j}\right) e_j\right\|_{l^2}} \leq \frac{1}{\gamma_1^2 \lambda^2} < 1,$$

with the same bound holding for $\Omega_2$. It follows that $\mathrm{Cond}(\Xi_{P_2}, \Omega) \leq 1$.

**Step 3:** Proof for ($P_3$). This is almost identical step 2 with replacing Lemma 3.11 with Lemma 3.12, and deleting the corresponding factors of $1 - \rho_j/2$. $\square$

3.4. **Details on the numerical example following Theorem 2 of the main text.** In this section, we elaborate on the numerical example following Theorem 2 of the main text. The example is a simplification of the arguments found in the proof of Theorem 2 that uses Lemma 3.12 extensively. In our experiment, we use $N_1 = 2$ and $\lambda = 1$, but for full generality, we do not keep these parameters fixed in the discussion below. We assume throughout that $\lambda \in (0, 1]$ and $N_1 \geq 2$. The experiment is done for real matrices so that the LISTA network architecture can be used.

Let $\gamma > 0$, $\rho \in \mathbb{R}^{N_1} \setminus \{0\}$, and $D \in \mathbb{C}^{N_2+1 \times N_2+1}$ be a unitary discrete cosine transform matrix. Define

$$A(\gamma, \rho) := D \begin{pmatrix} a(\gamma, \rho)^\top & 0 \\ 0 & \|a(\gamma, \rho)\|_{l^2} I \end{pmatrix}, \quad \text{where} \quad a(\gamma, \rho)^\top := \gamma \left(\frac{1}{\rho_1} \cdots \frac{1}{\rho_{N_1}}\right) \in \mathbb{R}^{1 \times N_1},$$

and $I \in \mathbb{R}^{N_2 \times N_2}$ is the identity matrix. Observe that $A(\gamma, \rho) \in \mathbb{R}^{m \times N}$, with $N = N_1 + N_2$ and $m = N_2 + 1$ and that $A$ has irrational entries (hence only approximations can be used in real-life computations). Furthermore, let $\delta = 1/6$ and $\gamma_K = \frac{\sqrt{2}}{3\lambda}(1 - \delta) \cdot 10^K$, where $K$ is the parameter from Theorem 2. Also let

$$y(x^{(2)}) = D \begin{pmatrix} 1 & \|a(\gamma, \rho)\|_{l^2} x^{(2)} \end{pmatrix}^\top \in \mathbb{R}^{N_2+1} \quad \text{for} \quad x^{(2)} \in \mathbb{R}^{N_2} \tag{3.12}$$

and $\Omega_K = \left\{ \left(y(x^{(2)}), A(\gamma_K, \rho)\right) : \rho_j \in [1 - 2\delta, 1 - \delta], x^{(2)} \in \mathbb{R}^{N_2} \right\}$. Next define

$$\rho' = \begin{pmatrix} \rho'_1 & 1-\delta & \cdots & \cdots & 1-\delta \end{pmatrix} \quad \text{and} \quad \rho^\sharp = \begin{pmatrix} 1-\delta & \rho_2^\sharp & 1-\delta & \cdots & 1-\delta \end{pmatrix}$$

where $\rho'_1, \rho_2^\sharp \in [1 - 2\delta, 1 - \delta)$. We let $e_i$ denote the $i$'th canonical basis vector for $\mathbb{R}^{N_1}$ and define $x' = \frac{\rho'_1}{\lambda \gamma_K} e_1$ and $x^\sharp = \frac{\rho_1^\sharp}{\lambda \gamma_K} e_2$. For this choice of parameters we have from Lemma 3.12 and the fact that $D$ is unitary that $(x', x^{(2)})^\top \in \Xi_3\left(A(\gamma_K, \rho'), y(x^{(2)})\right)$ and $(x^\sharp, x^{(2)})^\top \in \Xi_3\left(A(\gamma_K, \rho^\sharp), y(x^{(2)})\right)$.

Observe that we can let $A(\gamma_K, \rho')$ and $A(\gamma_K, \rho^\sharp)$ become arbitrary close by letting $\rho'_1, \rho_2^\sharp \uparrow 1 - \delta$. We will let $\rho'_1 = \rho_2^\sharp$, and notice that for this choice the data $y(x^{(2)})$ are the same for both inputs. However, regardless

of the choice of $\rho'_1, \rho_2^\sharp \in [1 - 2\delta, 1 - \delta)$ the minimisers for the two problems are bounded away from each other. In particular, we have that

$$\inf_{\rho'_1, \rho_2^\sharp \in [1-2\delta, 1-\delta]} \frac{1}{\lambda \gamma_K} \|\rho'_1 e_1 - \rho_2^\sharp e_2\|_{l^2} > 2 \cdot 10^K \quad \text{and} \quad \sup_{\rho'_1, \rho_2^\sharp \in [1-2\delta, 1-\delta]} \frac{1}{\lambda \gamma_K} \|\rho'_1 e_1 - \rho_2^\sharp e_2\|_{l^2} = 3 \cdot 10^K$$

which implies that $10^K < \|(x', x^{(2)})^\top - (x^\sharp, x^{(2)})^\top\|_{l^2} < 10^{K+1}$.

In the numerical experiment, we take $A = A(\gamma_K, \rho')$ with $\rho'_1 = 1 - \delta + 2^{-n-1}$, and approximate this matrix with the matrix $A_n = A(\gamma_K, \rho^\sharp)$, where the parameter $\rho_2^\sharp = 1 - \delta + 2^{-n-1}$. This ensures that $\|A - A_n\| \leq 2^n$. For the trained neural network, we used 8000 triples of the form

$$\iota_{A,S,n} = \left\{ \left( y_{k,n}(x_{k,n}^{(2)}), A_n, (x_n^\sharp, x_{k,n}^{(2)})^\top \right) : k = 1, \ldots, 8000, \text{ and } x^{(2)} \text{ is 5-sparse} \right\}. \quad (3.13)$$

for $n = 10, 20, 30$. Note that it is not necessary to make the $x^{(2)}$ component sparse. This is done merely to make the experiment more realistic, as the main usage of the problems $(P_j)$ are for recovery of sparse vectors.

## 4. PROOF OF THEOREM 3

A roadmap for the proof is as follows. We consider the problem $(P_3)$ and unroll iterations of Chambolle and Pock's primal-dual algorithm [13, 14]. These iterations are approximated by NNs in Theorem 5, where we obtain bounds on a rescaled version of the objective function in (4.9). The assumption of weighted rNSPL then allows us to relate the bounds proven in Theorem 5 to bounds on the distance of the output of the NN to the wanted vector and also, simultaneously, prove stability. This also allows the acceleration to exponential convergence through a restart scheme (with a reweighting at each restart). We begin with the proof of Lemma 1.8, which allows us to consider the approximation matrices $A_l$ in the construction of the NNs. We also state some results from compressed sensing that are needed in our proofs. We then discuss preliminary results on unrolling iterative algorithms for $(P_3)$, which are used in the proof of Theorem 3. When writing out NNs in the proofs, we will use $\xrightarrow{\text{NL}}$ arrows to denote the non-linear maps and $\xrightarrow{\text{L}}$ arrows to denote the affine maps.

### 4.1. Some results from compressed sensing.

*Proof of Lemma 1.8.* Let $\Delta$ be a $(\mathbf{s}, \mathbf{M})$ support set and $x \in \mathbb{C}^N$, then

$$\|x_\Delta\|_{l^2} \leq \frac{\rho \|x_{\Delta^c}\|_{l^1_w}}{\sqrt{\xi}} + \gamma \|Ax\|_{l^2} \leq \frac{\rho \|x_{\Delta^c}\|_{l^1_w}}{\sqrt{\xi}} + \gamma \|\hat{A}x\|_{l^2} + \gamma \|\hat{A} - A\| \|x\|_{l^2}. \quad (4.1)$$

Note that $\min_{k=1,\ldots,r} w_{(k)}^2 \sum_{j \in \Delta^c} |x_j|^2 \leq (\sum_{j \in \Delta^c} |x_j| w_j)^2 = \|x_{\Delta^c}\|_{l^1_w}^2$ and hence, (4.1) implies that

$$\|x_\Delta\|_{l^2} \leq \frac{\rho \|x_{\Delta^c}\|_{l^1_w}}{\sqrt{\xi}} + \gamma \|\hat{A}x\|_{l^2} + \gamma \|\hat{A} - A\| \left( \|x_\Delta\|_{l^2} + \frac{\|x_{\Delta^c}\|_{l^1_w}}{\min_{k=1,\ldots,r} w_{(k)}} \right).$$

Rearranging now gives the result. □

The following results are taken from the compressed sensing literature [1].

**Lemma 4.1** (rNSPL implies $l^1_w$ distance bound). *Suppose that $A$ has the weighted rNSPL of order $(\mathbf{s}, \mathbf{M})$ with constants $0 < \rho < 1$ and $\gamma > 0$. Let $x, z \in \mathbb{C}^N$, then*

$$\|z - x\|_{l^1_w} \leq \frac{1 + \rho}{1 - \rho} \left( 2\sigma_{\mathbf{s}, \mathbf{M}}(x)_{l^1_w} + \|z\|_{l^1_w} - \|x\|_{l^1_w} \right) + \frac{2\gamma}{1 - \rho} \sqrt{\xi} \|A(z - x)\|_{l^2}. \quad (4.2)$$

**Lemma 4.2** (rNSPL implies $l^2$ distance bound). *Suppose that $A$ has the weighted rNSPL of order $(\mathbf{s}, \mathbf{M})$ with constants $0 < \rho < 1$ and $\gamma > 0$. Let $x, z \in \mathbb{C}^N$, then*

$$\|z - x\|_{l^2} \leq \left( \rho + \frac{(1 + \rho)\kappa^{1/4}}{2} \right) \frac{\|z - x\|_{l^1_w}}{\sqrt{\xi}} + \left( 1 + \frac{\kappa^{1/4}}{2} \right) \gamma \|A(z - x)\|_{l^2}. \quad (4.3)$$

4.2. **Preliminary constructions of neural networks.** When constructing NNs, we will make use of the following maps from $\mathbb{C}^M$ to $\mathbb{C}^M$, defined for various $M \in \mathbb{N}$ and $\beta \in \mathbb{Q}_{>0}$ by $\psi_\beta^0(x) = \max\{0, 1 - \beta/\|x\|_{l^2}\}x$, $\quad \psi^1(x) = \min\{1, \|x\|_{l^2}^{-1}\}x$.

**Lemma 4.3.** *Let $M \in \mathbb{N}$, $\beta \in \mathbb{Q}_{>0}$ and $\theta \in \mathbb{Q}_{>0}$. Then there exists neural networks $\phi_{\beta,\theta}^0, \phi_\theta^1 \in \mathcal{N}_{\mathbf{D},3,2}$ with $\mathbf{D} = (M, 2M, M+1, M)$ such that $\|\phi_{\beta,\theta}^0(x) - \psi_\beta^0(x)\|_{l^2} \leq \theta$ and $\|\phi_\theta^1(x) - \psi^1(x)\|_{l^2} \leq \theta$ for all $x \in \mathbb{C}^M$, and the non-linear maps can be computed from $\mathrm{sqrt}_\theta$ and finitely many arithmetic operations and comparisons.*

*Proof.* We deal only with the case of $\psi_\beta^0$ since the case of $\psi^1$ is nearly identical. Consider the maps $\phi_{\beta,\theta}^0$:

$$x \xrightarrow{L} \begin{pmatrix} x \\ x \end{pmatrix} \xrightarrow{NL} \begin{pmatrix} |x_1|^2 \\ |x_2|^2 \\ \vdots \\ |x_M|^2 \\ x \end{pmatrix} \xrightarrow{L} \begin{pmatrix} \sum_{j=1}^M |x_j|^2 \\ x \end{pmatrix} \xrightarrow{NL} \begin{pmatrix} 0 \\ \max\left\{0, 1 - \frac{\beta}{\mathrm{sqrt}_\theta(\|x\|_{l^2}^2)}\right\}x \end{pmatrix} \xrightarrow{L} \max\left\{0, 1 - \frac{\beta}{\mathrm{sqrt}_\theta(\|x\|_{l^2}^2)}\right\}x.$$

The first, third and final arrows are simple affine maps. The second arrow applies pointwise modulus squaring, which can be done using finitely many arithmetic operations. The penultimate arrow applies a non-linear map which can be computed from one application of $\mathrm{sqrt}_\theta$ and finitely many arithmetic operations and comparisons. The bound $\|\psi_\beta^0(x) - \phi_{\beta,\theta}^0(x)\|_{l^2} \leq \theta$ follows from a simple case by case analysis. $\square$

The final piece of machinery needed is a NN approximation of applying a pointwise version of $\psi_\beta^0$.

**Lemma 4.4.** *Let $s, \theta \in \mathbb{Q}_{>0}$, $w \in \mathbb{Q}_{>0}^N$ and for $\hat{x} \in \mathbb{C}^N$ consider the minimisation problem*

$$\mathrm{argmin}_{x \in \mathbb{C}^N} \|x\|_{l_w^1} + s\|x - \hat{x}\|_{l^2}^2. \tag{4.4}$$

*Let $\tilde{x}_s(\hat{x})$ denote the solution of (4.4). Then, there exists $\phi_{s,\theta} \in \mathcal{N}_{\mathbf{D},2,1}$ such that*

$$\|\phi_{s,\theta}(\hat{x}) - \tilde{x}_s(\hat{x})\|_{l^2} \leq \theta\|w\|_{l^2}, \quad \forall \hat{x} \in \mathbb{C}^N \tag{4.5}$$

*and $\mathbf{D} = (N, N, N)$. Each affine map in the NN is linear and is an arithmetic function of $w$. Moreover, the non-linear maps used can be computed from $\mathrm{sqrt}_\theta$ and finitely many arithmetic operations and comparisons.*

*Proof.* Let $B = \mathrm{diag}(w_1, ..., w_N) \in \mathbb{Q}^{N \times N}$ and consider the function $F(y) = \|By\|_{l^1}/(2s) = \|y\|_{l_w^1}/(2s)$. We write the minimisation problem in (4.4) as $\mathrm{prox}_F(\hat{x})$. Given $y \in \mathbb{C}^N$, we identify $y = (y_1, y_2)^\top \in \mathbb{R}^{2N}$.

First, for $\beta > 0$ and $x \in \mathbb{R}^n$ recall that the proximal operator of a multiple of the $l^2$-norm is

$$\mathrm{prox}_{\beta\|\cdot\|_{l^2}}(x) = \max\{0, 1 - \beta/\|x\|_{l^2}\}x. \tag{4.6}$$

Thus, for $\beta > 0$ we define $\varphi_\beta(y) = (v(y, \beta) * y_1, v(y, \beta) * y_2)^\top$, where $*$ denotes pointwise multiplication and $v(y, \beta)_j = \max\{0, 1 - \beta/\sqrt{y_{1,j}^2 + y_{2,j}^2}\}$ for $j = 1, ..., N$. The function $\varphi_\beta$ simply corresponds to a proximity map of the $l^2$-norm applied component-wise to the complexified version of $y$. Using (4.6), we have

$$\mathrm{prox}_F(y) = \mathrm{argmin}_{z \in \mathbb{C}^N} \frac{1}{2s}\|Bz\|_{l^1} + \frac{1}{2}\|z - y\|_{l^2}^2$$

$$= \mathrm{argmin}_{z \in \mathbb{C}^N} \sum_{j=1}^N \left(\frac{B_{jj}}{2s}\sqrt{z_{1,j}^2 + z_{2,j}^2} + \frac{1}{2}\left((z_{1,j} - y_{1,j})^2 + (z_{2,j} - y_{2,j})^2\right)\right)$$

It follows that (in complex vector form) $[\mathrm{prox}_F(y)]_j = \{0, 1 - \frac{B_{jj}/(2s)}{|y_j|}\}y_j$, for $j = 1, \ldots, N$. We can therefore write $\mathrm{prox}_F(y) = B\varphi_{(2s)^{-1}}(B^{-1}y)$. We unroll the computation of $\mathrm{prox}_F(\hat{x})$ via:

$$\hat{x} \xrightarrow{L} B^{-1}\hat{x} \xrightarrow{NL} \varphi_{(2s)^{-1}}(B^{-1}\hat{x}) \xrightarrow{L} B\varphi_{(2s)^{-1}}(B^{-1}y).$$

The first arrow is a simple linear map, the second applies $\varphi_{(2s)^{-1}}$ and the third is a linear map. We approximate this by replacing $v(y, \beta)_j y_j$ with $\phi_{\beta,\theta}^0(y_{1,j} + y_{2,j}i)$ (denoting the replacement of $\varphi_{(2s)^{-1}}$ by $\varphi_{(2s)^{-1}}^\theta$) where

$\phi^0_{\beta,\theta}$ is the NN from Lemma 4.3 with $M = 1$. This clearly gives $\phi_{s,\theta} \in \mathcal{N}_{\mathbf{D},2,1}$, so we need to only bound the error. From Lemma 4.3 we have

$$\left\|\mathrm{prox}_F(\hat{x}) - B\varphi^\theta_{(2s)^{-1}}(B^{-1}\hat{x})\right\|_{l^2} = \left\|B\left(\varphi_{(2s)^{-1}}(B^{-1}\hat{x}) - \varphi^\theta_{(2s)^{-1}}(B^{-1}\hat{x})\right)\right\|_{l^2} \leq \theta \|w\|_{l^2}.$$

The bound in (4.5) now follows. □

The following theorem proves that one can construct NNs with objective function bounds. The proof constructs approximations of unrolled iterations of Chambolle and Pock's primal-dual algorithm [13, 14]. We have used $b$ to denote part of the inputs of the NNs, instead of $y$, to avoid a clash of notation with the usual notation for primal-dual iterations ($y$ is used to denote a dual variable). The bounds in part 2 of Theorem 5 will be combined with results from §4.4.1 to construct the families of NNs in Theorem 3.

**Theorem 5.** *Let $A \in \mathbb{Q}[i]^{m \times N}$ and $\theta \in \mathbb{Q}_{>0}$. Suppose also that $L_A \in \mathbb{Q}_{\geq 1}$ is an upper bound for $\|A\|$, and that $\tau, \sigma \in \mathbb{Q}_{>0}$ are such that $\tau\sigma L_A^2 < 1$. Let $\lambda \in \mathbb{Q}_{>0}$, $w \in \mathbb{Q}_{>0}^N$ and consider the resulting optimisation problem $(P_3)$. Then there exists an algorithm that constructs a sequence of neural networks $\{\phi^A_{n,\lambda}, \theta\}$ with the following properties:*

(1) *(Size) Each $\phi^A_{n,\lambda} : \mathbb{C}^{m+N} \to \mathbb{C}^N$ takes as input data $b \in \mathbb{C}^m$ and an initial guess $x_0 \in \mathbb{C}^N$, both of which are completely general. Also, $\phi^A_{n,\lambda} \in \mathcal{N}_{\mathbf{D}_n, 3n+1, 3}$ with*

$$\mathbf{D}_n = (m + N, \underbrace{2N + m, 2(N+m), 2N + m + 1}_{\text{repeated } n \text{ times}}, N).$$

(2) *($\mathcal{O}(n^{-1} + n\theta)$ Error Control) Let*

$$C = (1 + \|w\|_{l^2} + 2\sigma\|A\|\|w\|_{l^2})\sqrt{\frac{\tau + \sigma}{1 - \tau\sigma L_A^2}}\sqrt{\frac{\tau + \sigma}{\tau\sigma}}, \tag{4.7}$$

*then for any inputs $b \in \mathbb{C}^m$ and $x_0 \in \mathbb{C}^N$, there exists a vector $\psi_n(b, x_0) \in \mathbb{C}^N$ with*

$$\left\|\psi_n(b, x_0) - \phi^A_{n,\lambda}(b, x_0)\right\|_{l^2} \leq n\theta C \tag{4.8}$$

*such that for any $x \in \mathbb{C}^N$ and $\eta \in [0, 1]$, it holds that*

$$\lambda\|\psi_n(b, x_0)\|_{l^1_w} - \lambda\|x\|_{l^1_w} + \eta\|A\psi_n(b, x_0) - b\|_{l^2} - \|Ax - b\|_{l^2} \leq \frac{1}{n}\left(\frac{\|x - x_0\|^2_{l^2}}{\tau} + \frac{\eta^2}{\sigma}\right). \tag{4.9}$$

*Proof.* We use the notation $b \in \mathbb{C}^m$ to denote an input vector for our NNs throughout the proof and reserve $y$ to denote dual vectors, consistent with the literature on primal-dual algorithms for saddle point problems.

**Step 1:** The first step is to consider an equivalent optimisation problem over $\mathbb{R}$ instead of $\mathbb{C}$, and rewrite the problem as a saddle point problem. For $x \in \mathbb{C}^N$, let $x_1 = \mathrm{real}(x)$ and $x_2 = \mathrm{imag}(x)$ and consider $x = (x_1, x_2)^\top$ as a vector in $\mathbb{R}^{2N}$ (and likewise for the dual variables). With an abuse of notation, we use the same notation for complex $x \in \mathbb{C}^N$ and the corresponding vector in $\mathbb{R}^{2N}$, though it will be clear from the context whether we refer to the complex or real case. We let $c = (\mathrm{real}(b), \mathrm{imag}(b))^\top$. Define the matrices

$$K_1 = \begin{pmatrix} \mathrm{real}(A) & -\mathrm{imag}(A) \\ \mathrm{imag}(A) & \mathrm{real}(A) \end{pmatrix} \in \mathbb{R}^{2m \times 2N}, \quad K_2 = \begin{pmatrix} \mathrm{real}(B) & -\mathrm{imag}(B) \\ \mathrm{imag}(B) & \mathrm{real}(B) \end{pmatrix} \in \mathbb{R}^{2N \times 2N},$$

corresponding to multiplication by the matrices $A$ and $B := \mathrm{diag}(w_1, ..., w_N)$ respectively. Let $\widetilde{F}_1 : \mathbb{R}^{2N} \to \mathbb{R}$ be defined by $\widetilde{F}_1(x) = \sum_{j=1}^N \sqrt{(K_2 x)_j^2 + (K_2 x)_{j+N}^2}$ and $\widetilde{F}_3(x) = \lambda \widetilde{F}_1(x)$. Then $(P_3)$ is equivalent to $\min_{x \in \mathbb{R}^{2N}} \widetilde{F}_3(x) + \|K_1 x - c\|_{l^2}$ and $L_A$ is an upper bound for $\|K_1\|$. The saddle point formulation of the problem is given by

$$\min_{x \in \mathbb{R}^{2N}} \max_{y \in \mathbb{R}^{2m}} \mathcal{L}(x, y) := \langle K_1 x, y \rangle + \widetilde{F}_3(x) - f_3^*(y), \tag{4.10}$$

where $f_3^*(y) = \chi_{B_1(0)}(y) + \langle c, y \rangle$, and $\chi_S$ denotes the indicator function of a set $S$, taking the value 0 on $S$ and $+\infty$ otherwise, and $B_1(0)$ denotes the closed $l^2$ unit ball.

**Step 2:** We will solve (4.10) by approximating Chambolle and Pock's primal-dual algorithm [13] (with a shift of updates considered in [14]) with a NN. We will write the iteration as an instance of the proximal point algorithm [22] and gain a non-expansive map in a norm which we relate to the standard Euclidean norm.

We start by setting $x^0 = x_0$ (one of the inputs of the NN) and $y^0 = 0$. Recall that for a convex function $h$, we have that $x = \operatorname{prox}_h(z)$ if and only if $z \in x + \partial h(x)$, where $\partial h$ denotes the subdifferential of $h$, see, for example, [16, Prop. B.23]. Letting $g = \widetilde{F}_3$ and $f^* = f_3^*$, the exact iterates can be written as

$$\begin{aligned} x^{k+1} &= \operatorname{argmin}_{x \in \mathbb{R}^{2N}} g(x) + \frac{1}{2\tau}\|x - (x^k - \tau K_1^* y^k)\|_{l^2}^2 = (I + \tau \partial g)^{-1}(x^k - \tau K_1^* y^k) \\ y^{k+1} &= \operatorname{argmin}_{y \in \mathbb{R}^{2m}} f^*(y) + \frac{1}{2\sigma}\|y - (y^k + \sigma K_1(2x^{k+1} - x^k))\|_{l^2}^2 \\ &= (I + \sigma \partial f^*)^{-1}\left[y^k + \sigma K_1(2x^{k+1} - x^k)\right]. \end{aligned} \quad (4.11)$$

Note that the solutions of these proximal mappings are given by Lemmas 4.3 and 4.4 and their proofs, as we describe explicitly below in step 4. The function $f^*$ also depends on the input data $b$.

Let $z = (x, y)^\top$ and define the matrix

$$M_{\tau\sigma} = \begin{pmatrix} \frac{1}{\tau}I & -K_1^* \\ -K_1 & \frac{1}{\sigma}I \end{pmatrix} \in \mathbb{R}^{2(m+N) \times 2(m+N)},$$

which is positive definite by the assumption $\tau\sigma L_A^2 < 1$ and hence induces a norm denoted by $\|\cdot\|_{\tau\sigma}$. We can write the iterations as (see, for example, [14, Sec. 3])

$$0 \in M_{\tau\sigma}^{-1}\begin{pmatrix} \partial g & K_1^* \\ -K_1 & \partial f^* \end{pmatrix}z^{k+1} + (z^{k+1} - z^k) \Rightarrow z^{k+1} = \left[I + M_{\tau\sigma}^{-1}\begin{pmatrix} \partial g & K_1^* \\ -K_1 & \partial f^* \end{pmatrix}\right]^{-1} z^k.$$

The multi-valued operator

$$M_{\tau\sigma}^{-1}\begin{pmatrix} \partial g & K_1^* \\ -K_1 & \partial f^* \end{pmatrix}$$

is maximal monotone with respect to the inner product induced by $M_{\tau\sigma}$ [22] and hence the iterates are non-expansive in the norm $\|\cdot\|_{\tau\sigma}$. We also have that

$$\|(x,y)^\top\|_{\tau\sigma}^2 \leq \frac{\|x\|_{l^2}^2}{\tau} + \frac{\|y\|_{l^2}^2}{\sigma} + 2L_A\|x\|_{l^2}\|y\|_{l^2} \leq \left(\frac{L_A}{\nu} + \tau^{-1}\right)\|x\|_{l^2}^2 + \left(L_A\nu + \sigma^{-1}\right)\|y\|_{l^2}^2,$$

for any $\nu > 0$ by the generalised AM–GM inequality. Choosing $\nu = \sigma L_A$ and using $\tau\sigma L_A^2 < 1$, we have that

$$\|(x,y)^\top\|_{\tau\sigma}^2 \leq (\tau^{-1} + \sigma^{-1})\|(x,y)^\top\|_{l^2}^2. \quad (4.12)$$

A similar calculation yields that

$$\|(x,y)^\top\|_{l^2}^2 \leq \frac{\tau + \sigma}{1 - \tau\sigma L_A^2}\|(x,y)^\top\|_{\tau\sigma}^2. \quad (4.13)$$

**Step 3:** Next, we use convergence guarantees proven in [14] to obtain inequalities that closely resemble (4.9). Define the ergodic averages $X^k = \frac{1}{k}\sum_{j=1}^k x^j, Y^k = \frac{1}{k}\sum_{j=1}^k y^j$. By convexity, the map from $(x^1, y^1)^\top$ to $(X^k, Y^k)^\top$ is also non-expansive in the norm $\|\cdot\|_{\tau\sigma}$. It also holds (see [14] Theorem 1 and remarks) that

$$\mathcal{L}(X^k, y) - \mathcal{L}(x, Y^k) \leq \frac{1}{k}\left(\frac{\|x - x_0\|_{l^2}^2}{\tau} + \frac{\|y\|_{l^2}^2}{\sigma}\right), \quad \forall x \in \mathbb{R}^{2N}, \forall y \in \mathbb{R}^{2m}. \quad (4.14)$$

Let $y$ be parallel to $KX^k - c$ such that $\|y\|_{l^2} = \eta \leq 1$, and $x$ be general in (4.14). This gives

$$\widetilde{F}_3(X^k) - \widetilde{F}_3(x) + \langle K_1 X^k - c, y\rangle + \langle c - K_1 x, Y^k\rangle \leq \frac{1}{k}\left(\frac{\|x - x_0\|_{l^2}^2}{\tau} + \frac{\eta^2}{\sigma}\right).$$

Since $\|Y^k\|_{l^2} \leq 1$ (otherwise we gain a contradiction in that the left-hand side of (4.14) is infinite), this implies

$$\widetilde{F}_3(X^k) - \widetilde{F}_3(x) + \eta\|K_1 X^k - c\|_{l^2} - \|K_1 x - c\|_{l^2} \leq \frac{1}{k}\left(\frac{\|x - x_0\|_{l^2}^2}{\tau} + \frac{\eta^2}{\sigma}\right). \quad (4.15)$$

**Step 4:** The next step is to unroll the iterations in (4.11) as (complex-valued) NNs that approximate the $X^k$. We unroll via the following steps:

$$\begin{pmatrix} X^k \\ x^k \\ y^k \end{pmatrix} \xrightarrow{L} \begin{pmatrix} X^k \\ x^k - \tau A^* y^k \\ y^k - \sigma A x^k \end{pmatrix} \xrightarrow{NL} \begin{pmatrix} X^k \\ x^{k+1} \\ y^k - \sigma A x^k \end{pmatrix} \xrightarrow{L} \begin{pmatrix} X^{k+1} \\ x^{k+1} \\ u^k \end{pmatrix} \xrightarrow{NL} \begin{pmatrix} X^{k+1} \\ x^{k+1} \\ y^{k+1} \end{pmatrix},$$

with $u^k = y^k + \sigma A(2x^{k+1} - x^k) - \sigma b$. The first arrow is a simple linear map, the second computes $x^{k+1} = (I + \tau\lambda \partial F_1^A)^{-1}(x^k - \tau A^* y^k)$. The third is an affine map and the final arrow applies $\psi^1$ to $u^k$. We now define the approximations $\tilde{Z}^k$ and $\tilde{z}^k$ (of $Z^k = (X^k, Y^k)^\top$ and $z^k = (x^k, y^k)^\top$ respectively) defined by replacing $\psi^1$ with $\phi_\theta^1$ (Lemma 4.3) and the computation of $(I + \tau\lambda\partial F_1^A)^{-1}(x^k - \tau A^* y^k)$ with $\phi_{(2\tau\lambda)^{-1},\theta}(\tilde{x}^k - \tau A^* \tilde{y}^k)$ (Lemma 4.4). We initialise the network with $\tilde{x}^0 = x_0$ and $y^0 = 0$. Since the composition of two affine maps is affine, it follows that the mapping from $(b, x_0)$ to $\tilde{X}^n$ can be realised by $\phi_{n,\lambda}^A \in \mathcal{N}_{\mathbf{D}_n, 3n+1, 3}$ with

$$\mathbf{D}_n = (m + N, \underbrace{2N + m, 2(N + m), 2N + m + 1}_{\text{repeated } n \text{ times}}, N).$$

Clearly, the sequence of NNs are NNs in the sense of §1.1.2.1 and can be constructed by an algorithm (see §3.3.1).

**Step 5:** Finally, we bound the difference between $Z^k$ and $\tilde{Z}^k$ to deduce (4.8), and the error bound in the objective function using the inequalities in Step 3. We write $\tilde{x}^k = x^k + e_1^k$, $\tilde{y}^k = y^k + e_2^k$ and clearly have that $e_1^0 = 0$ and $e_2^0 = 0$. We can write $\tilde{x}^{k+1} = \phi_{(2\tau\lambda)^{-1}}(\tilde{x}^k - \tau A^* \tilde{y}^k) + e_3^{k+1}$, with $\|e_3^{k+1}\|_{l^2} \leq \theta \|w\|_{l^2}$ by Lemma 4.4. We also have that $\tilde{y}^{k+1} = \psi^1(\tilde{y}^k + \sigma A(2\tilde{x}^{k+1} - \tilde{x}^k) - \sigma b) + e_4^{k+1}$, with $\|e_4^{k+1}\|_{l^2} \leq \theta$ by Lemma 4.3. Since $\psi^1$ is non-expansive, it follows that $\tilde{y}^{k+1} = \psi^1(\tilde{y}^k + \sigma A(2(\tilde{x}^{k+1} - e_3^{k+1}) - \tilde{x}^k) - \sigma b) + e_5^{k+1}$, with $\|e_5^{k+1}\|_{l^2} \leq \theta(1 + 2\sigma \|A\| \|w\|_{l^2})$. We can then use the fact that the iterates applied with the exact proximal maps are non-expansive in the norm $\|\cdot\|_{\tau\sigma}$, along with (4.13) and (4.12), to conclude that

$$\|X^n - \tilde{X}^n\|_{l^2} \leq \sqrt{\frac{\tau + \sigma}{1 - \tau\sigma L_A^2}} \|Z^n - \tilde{Z}^n\|_{\tau\sigma}$$

$$\leq \sqrt{\frac{\tau + \sigma}{1 - \tau\sigma L_A^2}} \left[ \|Z^{n-1} - \tilde{Z}^{n-1}\|_{\tau\sigma} + \theta\sqrt{\frac{\tau + \sigma}{\tau\sigma}} \left(1 + \|w\|_{l^2} + 2\sigma \|A\| \|w\|_{l^2}\right) \right]$$

$$\leq n\theta(1 + \|w\|_{l^2} + 2\sigma \|A\| \|w\|_{l^2}) \sqrt{\frac{\tau + \sigma}{1 - \tau\sigma L_A^2}} \sqrt{\frac{\tau + \sigma}{\tau\sigma}}.$$

It follows that (4.8) holds with $\psi_n(b, x_0) = X^n$ and the complex version of (4.15) implies (4.9). □

4.3. **Proof of Theorem 3.** **Step 1:** The first step is to derive a bound on the distance between vectors using the square-root LASSO objective function and rNSPL. For any inputs $A$ (the rational approximations $\{A_l\}$), $\rho$ and $\gamma$ described in the theorem, we can compute, using Lemma 1.8, a positive integer $l$ in finitely many arithmetic operations and comparisons, such that $A_l \in \mathbb{Q}[i]^{m \times N}$ satisfies the rNSPL with constants $(1 + \rho)/2 \in (0, 1)$, $2\gamma > 0$. Lemmas 4.1 and 4.2 therefore imply that for any pair $z_1, z_2 \in \mathbb{C}^N$ we have

$$\|z_1 - z_2\|_{l_w^1} \leq \frac{3 + \rho}{1 - \rho} \left(2\sigma_{\mathbf{s},\mathbf{M}}(z_2)_{l_w^1} + \|z_1\|_{l_w^1} - \|z_2\|_{l_w^1}\right) + \frac{8\gamma\sqrt{\xi}}{1 - \rho} \|A_l(z_1 - z_2)\|_{l^2}, \quad (4.16)$$

$$\|z_1 - z_2\|_{l^2} \leq \left(\frac{1 + \rho}{2} + \frac{(3 + \rho)\kappa^{1/4}}{4}\right) \frac{\|z_1 - z_2\|_{l_w^1}}{\sqrt{\xi}} + \left(2 + \kappa^{1/4}\right) \gamma \|A_l(z_1 - z_2)\|_{l^2}. \quad (4.17)$$

Combining these two inequalities, we obtain the bound

$$\|z_1 - z_2\|_{l^2} \leq \frac{2C_1}{\sqrt{\xi}} \sigma_{\mathbf{s},\mathbf{M}}(z_2)_{l_w^1} + \frac{C_1}{\sqrt{\xi}} \left(\|z_1\|_{l_w^1} - \|z_2\|_{l_w^1}\right) + C_2 \|A_l(z_1 - z_2)\|_{l^2}$$

$$\leq \frac{2C_1}{\sqrt{\xi}} \sigma_{\mathbf{s},\mathbf{M}}(z_2)_{l_w^1} + 2C_2 \|A_l z_2 - y\|_{l^2} + \frac{C_1}{\lambda\sqrt{\xi}} \left(\lambda\|z_1\|_{l_w^1} - \lambda\|z_2\|_{l_w^1} + \|A_l z_1 - y\|_{l^2} - \|A_l z_2 - y\|_{l^2}\right),$$

(4.18)

where the second inequality follows from the fact that $\|A_l(z_1 - z_2)\|_{l^2} \leq \|A_l z_1 - y\|_{l^2} + \|A_l z_2 - y\|_{l^2}$ and we chose a positive rational $\lambda \leq C_1/(C_2\sqrt{\xi})$ (we will specify how small $|\lambda - C_1/(C_2\sqrt{\xi})|$ must be later, and always assume $\lambda \sim C_1/(C_2\sqrt{\xi})$). For notational convenience, we define

$$G(z_1, z_2, y) := \lambda \|z_1\|_{l^1_w} - \lambda \|z_2\|_{l^1_w} + \|A_l z_1 - y\|_{l^2} - \|A_l z_2 - y\|_{l^2}, \tag{4.19}$$

the difference between the values of the objective function $F_3^A$ for arguments $z_1$ and $z_2$. We also define

$$c(z, y) := \frac{2C_1}{C_2\sqrt{\xi}} \cdot \sigma_{\mathbf{s},\mathbf{M}}(z)_{l^1_w} + 2\|A_l z - y\|_{l^2}. \tag{4.20}$$

It follows from (4.18) and $\lambda \leq C_1/(C_2\sqrt{\xi})$ that

$$\|z_1 - z_2\|_{l^2} \leq \frac{C_1}{\lambda\sqrt{\xi}} \left( c(z_2, y) + G(z_1, z_2, y) \right), \tag{4.21}$$

which also implies the bound $G(z_1, z_2, y) \geq -c(z_2, y)$. These bounds hold for general $z_1, z_2$ and $y$.

**Step 2:** We now apply Theorem 5 using a suitable scaling to define a family of parametrised NNs, which we iterate later in the proof (this corresponds to restarting primal-dual iterations with different parameters). Let $\sigma = \tau \in (4\|A_l\|^{-1}/5, 5\|A_l\|^{-1}/6)$ be positive rational numbers. We can compute such parameters by approximating $\|A_l\|$ via any standard algorithm that approximates the largest singular value of a rectangular matrix using finitely many arithmetic operations and comparisons. We now use Theorem 5 (with $\theta$ specified below) with input $y/(p\beta)$ and $x_0/(p\beta)$ for a given $p \in \mathbb{N}$, and $\beta \in \mathbb{Q}_{>0}$ (which we explicitly define below). Given $\phi_{p,\lambda}^{A_l}(y/(p\beta), x_0/(p\beta))$, Theorem 5 ensures the existence of a vector $\psi_p = \psi_p(y/(p\beta), x_0/(p\beta))$ satisfying

$$\left\| \psi_p \left( \frac{y}{p\beta}, \frac{x_0}{p\beta} \right) - \phi_{p,\lambda}^{A} \left( \frac{y}{p\beta}, \frac{x_0}{p\beta} \right) \right\|_{l^2} \leq pC\theta$$

where $C$ is given in (4.7) and

$$\lambda \|\psi_p\|_{l^1_w} - \lambda \left\| \frac{x}{p\beta} \right\|_{l^1_w} + \left\| A\psi_p - \frac{y}{p\beta} \right\|_{l^2} - \frac{1}{p\beta} \|Ax - y\|_{l^2} \leq \frac{1}{p} \left( \frac{\|x(p\beta)^{-1} - x_0(p\beta)^{-1}\|_{l^2}^2}{\tau} + \frac{1}{\sigma} \right) \tag{4.22}$$

for any $x \in \mathbb{C}^N$ (and we have taken $\eta = 1$ in (4.9)). Define the map $H_p^\beta : \mathbb{C}^m \times \mathbb{C}^N \to \mathbb{C}^N$ by

$$H_p^\beta(y, x_0) = p\beta \phi_{p,\lambda}^{A_l} \left( \frac{y}{p\beta}, \frac{x_0}{p\beta} \right).$$

The additional scaling factors can be incorporated so that $H_p^\beta \in \mathcal{N}_{\mathbf{D}_p, 3p+1, 3}$. Rescaling (4.22) yields the existence of a vector $\hat{\psi}_p(y, x_0) \in \mathbb{C}^N$ (where the $\hat{\cdot}$ denotes an appropriate rescaling by multiplying by $p\beta$) such that

$$G\left( \hat{\psi}_p(y, x_0), x, y \right) \leq \frac{5}{4} \left( \frac{\|A_l\|}{p^2 \beta} \|x - x_0\|_{l^2}^2 + \|A_l\|\beta \right), \tag{4.23}$$

where we have used $\tau^{-1} = \sigma^{-1} \leq 5\|A_l\|/4$. Moreover, the constant $C$ in Theorem 5 is bounded by

$$C = (1 + \|w\|_{l^2} + 2\sigma\|A_l\|\|w\|_{l^2}) \sqrt{\frac{\tau + \sigma}{1 - \tau\sigma L_A^2}} \sqrt{\frac{\tau + \sigma}{\tau\sigma}} \leq \hat{C}_1(1 + \|w\|_{l^2}), \tag{4.24}$$

for a constant $\hat{C}_1$ that we can explicitly compute. Hence, upon rescaling (4.8), we arrive at

$$\left\| \hat{\psi}_p(y, x_0) - H_p^\beta(y, x_0) \right\|_{l^2} \leq p^2\theta\beta\hat{C}_1(1 + \|w\|_{l^2}).$$

Using Hölder's inequality, this also implies that

$$\left\| \hat{\psi}_p(y, x_0) - H_p^\beta(y, x_0) \right\|_{l^1_w} \leq p^2\theta\beta\hat{C}_1(1 + \|w\|_{l^2})\|w\|_{l^2}.$$

It follows from the reverse triangle inequality that

$$G\left( H_p^\beta(y, x_0), x, y \right) \leq G\left( \hat{\psi}_p(y, x_0), x, y \right) + p^2\theta\beta\hat{C}_1(1 + \|w\|_{l^2})(\|A_l\| + \lambda\|w\|_{l^2}). \tag{4.25}$$

Using this bound in (4.23), and the fact that $\lambda \lesssim (\gamma\sqrt{\xi})^{-1}$, we can choose $\theta \in \mathbb{Q}_{>0}$ such that

$$\theta^{-1} \lesssim p^2(1+\|w\|_{l^2})\max\left\{1, \frac{\lambda\|w\|_{l^2}}{\|A_l\|}\right\} \lesssim p^2(1+\|w\|_{l^2})\max\left\{1, \frac{\|w\|_{l^2}}{\|A\|\gamma\sqrt{\xi}}\right\},$$

and, simultaneously,

$$G\left(H_p^\beta(y,x_0),x,y\right) \leq \frac{4}{3}\left(\frac{\|A_l\|}{p^2\beta}\|x-x_0\|_{l^2}^2 + \|A_l\|\beta\right).$$

Combining this with (4.21), we obtain the key inequality

$$G\left(H_p^\beta(y,x_0),x,y\right) \leq \frac{4C_1^2\|A_l\|}{3p^2\beta\lambda^2\xi}\left[c(x,y)+G(x_0,x,y)\right]^2 + \frac{4}{3}\|A_l\|\beta. \tag{4.26}$$

**Step 3:** In this step, we specify the choice of $p$ and $\beta$. So far, we have not used any information regarding the vectors $x$ and $y$. Recall that for our recovery theorem, we restricted to pairs $(x,y)$ such that

$$\frac{2C_1}{C_2\sqrt{\xi}} \cdot \sigma_{\mathbf{s},\mathbf{M}}(x)_{l_w^1} + 2\|Ax-y\|_{l^2} \leq \delta, \quad \|x\|_{l^2} \leq b_1, \quad \|y\|_{l^2} \leq b_2.$$

Using this, we can choose $l$ larger if necessary such that for any such $(x,y)$, we have the bound

$$c(x,y) \leq \frac{2C_1}{C_2\sqrt{\xi}} \cdot \sigma_{\mathbf{s},\mathbf{M}}(x)_{l_w^1} + 2\|Ax-y\|_{l^2} + 2\|A-A_l\|\|x\|_{l^2} \leq 2\delta.$$

The following lemma shows how to choose $\beta$ and $p$ to gain a decrease in $G$ by a factor of $\upsilon \in (0,1)$, up to small controllable error terms.

**Lemma 4.5.** *Let $\upsilon \in (0,1) \cap \mathbb{Q}_{>0}$, $\epsilon_0 \in \mathbb{Q}_{>0}$ and choose $\beta \in \mathbb{Q}_{>0}$ such that $8\|A_l\|\beta = 3\upsilon_0\upsilon(\epsilon_0+2\delta)$ for some $\upsilon_0 \in [1,2)$. Then for any $x_0$ with $G(x_0,x,y) \leq \epsilon_0$ and positive integer $p \geq \left\lceil \frac{8C_1\|A_l\|}{3\upsilon\lambda\sqrt{\xi}\sqrt{(2-\upsilon_0)\upsilon_0}} \right\rceil$ the following bound holds*

$$G\left(H_p^\beta(y,x_0),x,y\right) \leq \upsilon(2\delta+\epsilon_0). \tag{4.27}$$

*Proof.* The choice of $\beta$ ensures that $\frac{4}{3}\|A_l\|\beta \leq \frac{\upsilon_0\upsilon}{2}(2\delta+\epsilon_0)$. Using (4.26), and the fact that $0 \leq c(x,y)+G(x_0,x,y) \leq 2\delta+\epsilon_0$, the bound in (4.27) therefore holds if

$$\frac{32C_1^2\|A_l\|^2}{9p^2\upsilon_0\upsilon\lambda^2\xi}(2\delta+\epsilon_0) \leq \frac{(2-\upsilon_0)\upsilon}{2}(2\delta+\epsilon_0).$$

Rearranging and taking the square root gives the result, where the ceiling function ensures $p$ is an integer. □

We denote the choice of $\beta$ in Lemma 4.5 by $\beta(\upsilon,\epsilon_0)$. Since $8/3 < 3$, we can, by taking $l$ larger and by making $\lambda$ closer to $C_1/(C_2\sqrt{\xi})$ if necessary, and through an appropriate choice of $\upsilon_0$, ensure that we can compute (using finitely many arithmetic operations and comparisons) a choice $p(\upsilon) \leq \left\lceil \frac{3C_2\|A\|}{\upsilon} \right\rceil$ such that the conclusion of the lemma holds.

**Step 4:** We are now ready to construct our NNs. Note first that $G(0,x,y) \leq \|y\|_{l^2} \leq b_2$, for any $y$ in our desired input. Given $n \in \mathbb{N}$, we set $\epsilon_0 = b_2$ and for $j = 2,...,n$ set $\epsilon_j = \upsilon(2\delta+\epsilon_{j-1})$. By summing a geometric series, this implies $\epsilon_n \leq \frac{2\upsilon\delta}{1-\upsilon} + \upsilon^n b_2$. We define $\phi_n(y)$ iteratively as follows. We set $\phi_1(y) = H_{p(\upsilon)}^{\beta(\upsilon,\epsilon_0)}(y,0)$ and for $j = 2,...,n$ we set $\phi_j(y) = H_{p(\upsilon)}^{\beta(\upsilon,\epsilon_{j-1})}(y,\phi_{j-1}(y))$. Clearly this algorithmically constructs a NN $\phi_n$. We can concatenate (by combining affine maps) the NNs corresponding to the $H_p^\beta$ maps to see that $\phi_n \in \mathcal{N}_{\mathbf{D}(n,p),3np+1,3}$. Moreover, Lemma 4.5 implies the bound $G(\phi_n(y),x,y) \leq \epsilon_n \leq \frac{2\upsilon\delta}{1-\upsilon} + \upsilon^n b_2$. Combining this with (4.18),

$$\|\phi_n(y)-x\|_{l^2} \leq \frac{2C_1}{\sqrt{\xi}}\sigma_{\mathbf{s},\mathbf{M}}(x)_{l_w^1} + 2C_2\|Ax-y\|_{l^2} + 2C_2\|A-A_l\|_{l^2}b_1 + \frac{C_1}{\lambda\sqrt{\xi}}\left(\frac{2\upsilon\delta}{1-\upsilon}+\upsilon^n b_2\right), \tag{4.28}$$

Again, we can apriori choose $l$ and $\lambda$ to ensure that

$$2C_2\|A-A_l\|_{l^2}b_1 + \frac{C_1}{\lambda\sqrt{\xi}}\left(\frac{2\upsilon\delta}{1-\upsilon}+\upsilon^n b_2\right) \leq C_2\left(\frac{2\upsilon\delta}{1-\upsilon}+\delta+\upsilon^n b_2\right).$$

Applying this bound to (4.28) yields (1.16).

Finally, we argue for the error in the weighted $l_w^1$-norm. Note that since $\rho < 1$, the choice of $\lambda$ ensures that $\frac{8\gamma\sqrt{\xi}}{1-\rho} < \frac{3+\rho}{1-\rho}\frac{1}{\lambda}$. It follows from (4.16), using the same argument for the $l^2$ case, that

$$\|\phi_n(y) - x\|_{l_w^1} \leq \frac{3+\rho}{1-\rho}\left(2\sigma_{\mathbf{s},\mathbf{M}}(x)_{l_w^1} + \frac{2}{\lambda}\|A_l x - y\|_{l^2} + \frac{1}{\lambda}G\left(\phi_n(y), x, y\right)\right), \tag{4.29}$$

Again, we can apriori adjust $l$ and $\lambda$ as necessary to obtain the bound

$$\|\phi_n(y) - x\|_{l_w^1} \leq \frac{3+\rho}{1-\rho}\left(2\sigma_{\mathbf{s},\mathbf{M}}(x)_{l_w^1} + \frac{2C_2\sqrt{\xi}}{C_1}\|Ax - y\|_{l^2} + \frac{C_2\sqrt{\xi}}{C_1}\left(\frac{2v\delta}{1-v} + v^n b_2\right) + \delta\frac{C_2\sqrt{\xi}}{C_1}\right),$$

where the final term in brackets corresponds to this final approximation. Simplifying this yields (1.17). □

To end this section, we provide a brief proof sketch of the bounds in Remark 1.10. The argument is similar to the proof of Theorem 3. We set $\widehat{\phi}_n(y, x_0) = \beta\phi_{n,\lambda}^{A_l}\left(\frac{y}{\beta}, \frac{x_0}{\beta}\right)$, and the arguments in Theorem 3 show that we can choose $\tau, \sigma, l$ and $\theta_n$ with $\theta_n^{-1} = \mathcal{O}(n^2)$ such that for any $x, x_0 \in \mathbb{C}^N$ and $y \in \mathbb{C}^m$,

$$G\left(\widehat{\phi}_n(y, x_0), x, y\right) \leq \frac{3}{2}\frac{\|A\|}{n}\left(\frac{\|x - x_0\|_{l^2}^2}{\beta} + \beta\right).$$

If $\|x\|_{l^2} \leq b_1$, then we can choose $l$ such that $2\|A - A_l\|b_2 \leq \|A\|\beta/(2n)\min\{C_1/(C_2\lambda\sqrt{\xi}), 1\}$ and hence (1.18) follows from (4.18). Similarly, we can use the corresponding bound in (4.29) to show (1.19).

## 5. PROOF OF THEOREM 4

For the benefit of the reader, we first recall the orthonormal bases used. We then provide coherence estimates which are used to obtain bounds on the number of samples needed, and end this section with the proof of Theorem 4. It will be convenient to sometimes enumerate the vector or tensor elements starting from 0, or negative numbers. That is for $x \in \mathbb{C}^N$ with $d = 1$ we might denote its elements as $x = (x(0), \ldots, x(N-1))$, or $x = (x(-N/2+1), \ldots, x(N/2))$ and for $d > 1$ its $\boldsymbol{k} = (k_1, \ldots, k_d)$'th element is written as $x(\boldsymbol{k})$. It will always be clear from the context, which range of indices we consider. Furthermore, recall from §1.1.3 that we let $N = K^d$ and $K = 2^r$ for $r \in \mathbb{Z}_{\geq 0}$. This is assumed throughout this section.

### 5.1. Setup: the relevant orthonormal bases.

*Discrete Fourier transform.* For a $d$-dimensional signal $x = \{x(\boldsymbol{t})\}_{t_1,\ldots,t_d=0}^{K-1} \in \mathbb{C}^{K \times \cdots \times K}$ we denote its Fourier transform by $[\mathcal{F}x](\boldsymbol{\omega}) = \frac{1}{N^{1/2}}\sum_{t_1,\ldots,t_d=0}^{K-1} x(\boldsymbol{t})\exp(\frac{2\pi i \boldsymbol{\omega} \cdot \boldsymbol{t}}{K})$, $\boldsymbol{\omega} \in \mathbb{R}^d$. For discrete computations, it is customary to consider this transform at the integers $\boldsymbol{\omega} \in \{-K/2+1, \ldots, K/2\}^d$ and let $F^{(d)} \in \mathbb{C}^{K^d \times K^d}$ denote the corresponding matrix so that $F^{(d)}\text{vec}(x) = \{[\mathcal{F}x](\boldsymbol{\omega})\}_{\boldsymbol{\omega} \in \{-K/2+1,\ldots,K/2\}^d}$ for a suitable vectorisation $\text{vec}(x)$ of $x$ and ordering of the $\boldsymbol{\omega}$'s. Let

$$\vartheta_{\boldsymbol{\omega}} = \left\{N^{-1/2}\exp\left(-2\pi i K^{-1}\boldsymbol{\omega} \cdot \boldsymbol{t}\right) : \boldsymbol{t} \in \{0, \ldots, K-1\}^d\right\} \subset \mathbb{C}^{K \times \cdots \times K}.$$

Then

$$\left\{\text{vec}(\vartheta_{\boldsymbol{\omega}}) : \boldsymbol{\omega} \in \{-K/2+1, \ldots, K/2\}^d\right\} \tag{5.1}$$

is an orthonormal basis for $\mathbb{C}^{K^d} = \mathbb{C}^N$. Furthermore, recall from §1.1.3, that we divide the different frequencies into dyadic bands. For $d = 1$ we let $B_1 = \{0, 1\}$ and

$$B_k = \left\{-2^{k-1}+1, \ldots, -2^{k-2}\right\} \cup \left\{2^{k-2}+1, \ldots, 2^{k-1}\right\}, \quad k = 2, \ldots, r.$$

In the general $d$-dimensional case we set $B_{\boldsymbol{k}}^{(d)} = B_{k_1} \times \ldots \times B_{k_d}$ for $\boldsymbol{k} = (k_1, \ldots, k_d) \in \mathbb{N}^d$.

*Walsh transform.*

**Definition 5.1.** *The Walsh functions* $v_n \colon [0,1) \to \{+1, -1\}$ *are defined by*

$$v_\omega(z) = (-1)^{\sum_{j=1}^\infty (\omega^{(j)} + \omega^{(j+1)}) z^{(j)}}, \quad z \in [0,1), \quad \omega \in \mathbb{Z}_{\geq 0}, \tag{5.2}$$

*where* $(z^{(i)})_{i \in \mathbb{N}}$ *denotes the binary expansion of $z$ (terminating if $z$ is a dyadic rational) and we write* $\omega = \sum_{j=1}^\infty \omega^{(j)} 2^{j-1}$ *for* $\omega^{(j)} \in \{0,1\}$. *For* $\bm{z} \in [0,1)^d$ *and* $\bm{\omega} \in \mathbb{Z}_{\geq 0}^d$, *we let* $v_{\bm{\omega}}(\bm{z}) = v_{\omega_1}(z_1) \cdots v_{\omega_d}(z_d)$.

For $x \in \mathbb{C}^{K \times \cdots \times K}$ and $K = 2^r$ we let its $d$-dimensional Walsh transform be denoted by

$$[\mathcal{W}x](\bm{\omega}) = \frac{1}{N^{1/2}} \sum_{t_1,\ldots,t_d=0}^{K-1} x(\bm{t}) v_{\bm{\omega}}(\bm{t}/K), \quad \bm{\omega} \in \{0,\ldots,2^r-1\}^d.$$

As in the Fourier case, we let $W^{(d)} \in \mathbb{C}^{N \times N}$ so that $W^{(d)} \mathrm{vec}(x) = \{[\mathcal{W}x](\bm{\omega})\}_{\bm{\omega} \in \{0,\ldots,K-1\}^d}$ for a suitable vectorisation of $x$ and ordering of the $\bm{\omega}$'s. We let $\varrho_{\bm{\omega}} = \{N^{-1/2} v_{\bm{\omega}}(\bm{t}/K) : \bm{t} \in \{0,\ldots,K-1\}^d\} \subset \mathbb{C}^{K \times \cdots \times K}$ and note that

$$\{\mathrm{vec}(\varrho_{\bm{\omega}}) : \bm{\omega} \in \{0,\ldots,K-1\}^d\} \tag{5.3}$$

is an orthonormal basis for $\mathbb{C}^N$. As in the Fourier case we recall the frequency bands introduced in §1.1.3. Let $B_1 = \{0,1\}$ and $B_k = \{2^{k-1},\ldots,2^k-1\}$ for $k = 2,\ldots,r$ in the one-dimensional case, and $B_{\bm{k}}^{(d)} = B_{k_1} \times \ldots \times B_{k_d}$, $\bm{k} = (k_1,\ldots,k_d) \in \mathbb{N}^d$. Whether the notation refers to the Walsh or Fourier frequency bands will always be clear from the context.

*Haar-wavelet transform.* On $\mathbb{C}^K$ the Haar wavelet vectors are defined as

$$\psi_{j,p}(i) = \begin{cases} 2^{\frac{j-r}{2}}, & p 2^{r-j} \leq i < \left(p + \frac{1}{2}\right) 2^{r-j} \\ -2^{\frac{j-r}{2}}, & \left(p + \frac{1}{2}\right) 2^{r-j} \leq i < (p+1) 2^{r-j} \\ 0, & \text{otherwise}, \end{cases}$$

for $j = 0,\ldots,r-1$ and $p = 0,\ldots,2^j - 1$, and we can define the corresponding scaling vectors as $\varphi_{j,p}(i) = |\psi_{j,p}(i)|$. To simplify the notation we set $\psi_{j,k}^{(0)} = \varphi_{j,k}$ and $\psi_{j,k}^{(1)} = \psi_{j,k}$. For $d > 1$ and $\bm{q} = (q_1,\ldots,q_d) \in \{0,1\}^d$, $\bm{p} = (p_1,\ldots,p_d) \in \mathbb{Z}_{\geq 0}^d$ define the tensor product $\psi_{j,\bm{p}}^{\bm{q}} = \psi_{j,p_1}^{(q_1)} \otimes \ldots \otimes \psi_{j,p_d}^{(q_d)}$. Splitting these tensors by scale

$$C_1 = \{\mathrm{vec}(\psi_{0,0}^{\bm{q}}) : \bm{q} \in \{0,1\}^d\}, \quad C_j = \{\mathrm{vec}(\psi_{j-1,\bm{p}}^{\bm{q}}) : \bm{q} \in \{0,1\}^d \setminus \{0\}, p_k = 0,\ldots,2^{j-1}-1\},$$

for $j = 2,\ldots,r$, we get that $C_1 \cup \cdots \cup C_r$ is an orthonormal basis for $\mathbb{C}^N$. Next, let the vectors in $C_1 \cup \cdots \cup C_r$, form the rows of a matrix $\Phi \in \mathbb{C}^{N \times N}$. The matrix $\Psi$ is called the *discrete wavelet transform* (DWT) matrix, and its inverse $\Psi^{-1}$ is called the *inverse discrete wavelet transform* (IDWT) matrix. Notice that since $C_1 \cup \cdots \cup C_r$ is an orthonormal basis, we have the relation $\Psi^{-1} = \Psi^*$.

**5.2. Uniform recovery guarantees and coherence estimates.** We express $U = \left[U^{(\bm{k},j)}\right]_{\bm{k}=1, j=1}^{\|\bm{k}\|_{\ell^\infty} \leq r, r}$ in block form, where the entries in each $U^{\bm{k},j}$ consist of the inner products $\langle \varphi, \rho_{\bm{\omega}} \rangle$ for $\varphi \in C_j$ and where $\rho_{\bm{\omega}}$ is an element in either (5.1) or (5.3) with $\bm{\omega} \in B_{\bm{k}}^{(d)}$, depending on whether we consider Fourier or Walsh sampling. For this decomposition we define local coherence as follows.

**Definition 5.2.** *Let* $U = \left[U^{(\bm{k},j)}\right]_{\bm{k}=1, j=1}^{\|\bm{k}\|_{\ell^\infty} \leq r, r}$ *be defined as above. Then the* $(\bm{k},j)$*th local coherence of $U$ is*

$$\mu(U^{\bm{k},j}) = \left|B_{\bm{k}}^{(d)}\right| \max_{p,q} |(U^{\bm{k},j})_{pq}|^2, \quad \text{where } \left|B_{\bm{k}}^{(d)}\right| \text{ is the cardinality of } B_{\bm{k}}^{(d)}.$$

Recall from Definition 1.6, that for an $(\bm{s},\bm{M})$-sparse vector, $s = s_1 + \ldots + s_r$ denotes the total sparsity. Furthermore, $m = \sum_{\bm{k}=1}^{\|\bm{k}\|_{\ell^\infty} \leq r} m_{\bm{k}}$ denotes the total number of samples in an $(\bm{N},\bm{m})$-multilevel subsampling scheme. The following shows that to use Theorem 3, we need to bound the local coherences of $U$.

**Proposition 5.3** ([1]). *Let $\epsilon_{\mathbb{P}} \in (0,1)$, $(\mathbf{s}, \mathbf{M})$ be local sparsities and sparsity levels respectively with $2 \leq s \leq N$, and consider the $(\mathbf{N}, \mathbf{m})$-multilevel subsampling scheme to form a subsampled unitary matrix A as in Definitions 1.11 and 1.12. Let*

$$t_j = \min\left\{ \left\lceil \frac{\xi(\mathbf{s}, \mathbf{M}, w)}{w_{(j)}^2} \right\rceil, M_j - M_{j-1} \right\}, \quad j = 1, ..., r, \tag{5.4}$$

*and suppose that*

$$m_k \gtrsim \mathcal{L}' \cdot \sum_{j=1}^{r} t_j \mu(U^{k,j}), \quad k = 1, ..., l \tag{5.5}$$

*where $\mathcal{L}' = r \cdot \log(2m) \cdot \log^2(t) \cdot \log(N) + \log(\epsilon_{\mathbb{P}}^{-1})$. Then with probability at least $1 - \epsilon_{\mathbb{P}}$, A satisfies the weighted rNSPL of order $(\mathbf{s}, \mathbf{M})$ with constants $\rho = 1/2$ and $\gamma = \sqrt{2}$.*

The following bound the local coherences of $U$, with $\mathcal{M}_{\mathcal{F}}(\mathbf{s}, \mathbf{k})$ and $\mathcal{M}_{\mathcal{W}}(\mathbf{s}, \mathbf{k})$ defined in (1.20) and (1.21).

**Lemma 5.4** (**Coherence bound for Fourier case**). *Consider the d-dimensional Fourier–Haar–wavelet matrix with blocks $U^{\mathbf{k},j}$, then the local coherences satisfy*

$$\mu(U^{\mathbf{k},j}) \lesssim 2^{-2(j-\|\mathbf{k}\|_{l^\infty})_+} \prod_{i=1}^{d} 2^{-|k_i - j|}, \tag{5.6}$$

*where for $t \in \mathbb{R}$, $t_+ = \max\{0, t\}$. It follows that*

$$\sum_{j=1}^{r} s_j \mu(U^{\mathbf{k},j}) \lesssim \sum_{j=1}^{\|\mathbf{k}\|_{l^\infty}} s_j \prod_{i=1}^{d} 2^{-|k_i - j|} + \sum_{j=\|\mathbf{k}\|_{l^\infty}+1}^{r} s_j 2^{-2(j-\|\mathbf{k}\|_{l^\infty})} \prod_{i=1}^{d} 2^{-|k_i - j|} = \mathcal{M}_{\mathcal{F}}(\mathbf{s}, \mathbf{k}). \tag{5.7}$$

*Proof.* From the one-dimensional case treated in [5, See proof of Lem. 1], we have

$$\left| \left[ \mathcal{F} \psi_{j,p}^{(1)} \right] (\omega) \right|^2 \lesssim \begin{cases} 2^{-k} 2^{-|k-j|}, & \text{if } j \leq k, \\ 2^{-k} 2^{-3|k-j|}, & \text{otherwise} \end{cases},$$

We proceed by showing that $|[\mathcal{F}\psi_{j,p}^{(0)}](\omega)|^2 \lesssim 2^{-k} 2^{-|k-j|}$ in the one-dimensional case, before considering $d$ dimensions. Let $\omega \neq 0$ correspond to a frequency in $B_k$, $j \in \{0, ..., r-1\}$ and $p \in \{0, ..., 2^j - 1\}$. Then

$$\left[ \mathcal{F} \psi_{j,p}^{(0)} \right] (\omega) = 2^{\frac{j}{2}-r} e^{2\pi i \omega p 2^{-j}} \sum_{t=0}^{2^{r-j}-1} e^{2\pi i \omega t 2^{-r}} = 2^{\frac{j}{2}-r} e^{2^{1-j}\pi i \omega p} \frac{1 - e^{2\pi i \omega 2^{-j}}}{1 - e^{2\pi i \omega 2^{-r}}}.$$

A simple application of the double angle formula then yields

$$\left| \left[ \mathcal{F} \psi_{j,p}^{(0)} \right] (\omega) \right| \lesssim 2^{\frac{j}{2}-r} \frac{|\sin(\pi \omega 2^{-j})|}{|\sin(\pi \omega 2^{-r})|} = \frac{2^{\frac{j}{2}}}{|\omega|} \frac{|\omega 2^{-r}|}{|\sin(\pi \omega 2^{-r})|} |\sin(\pi \omega 2^{-j})| \lesssim 2^{\frac{j}{2}-k} |\sin(\pi \omega 2^{-j})|,$$

where the second inequality follows from $|\omega 2^{-r}| \leq 1/2$ and $2^k \lesssim |\omega|$. If $k > j$, this implies $|[\mathcal{F}\psi_{j,p}^{(0)}](\omega)|^2 \lesssim 2^{-k} 2^{-|k-j|}$. If $k \leq j$, we use that $|\sin(\pi t)| \leq \pi |t|$, $\forall t \in \mathbb{R}$ to get $|\sin(\pi \omega 2^{-j})| \lesssim 2^{k-j}$. Hence,

$$\left| \left[ \mathcal{F} \psi_{j,p}^{(0)} \right] (\omega) \right|^2 \lesssim 2^{j-2k} 2^{2k-2j} = 2^{-j} = 2^{-k} 2^{-|k-j|}.$$

If $\omega = 0$ then by definition we have $|[\mathcal{F}\psi_{j,p}^{(0)}](\omega)|^2 \lesssim 2^{-j} = 2^{-k} 2^{-|k-j|}$ and hence this bound still holds.

We now consider the general $d$-dimensional case. The above computations give that

$$\mu(U^{(\mathbf{k},1)}) \lesssim 2^{\sum_{i=1}^{d} k_i} \max_{\mathbf{q} \in \{0,1\}^d} \prod_{i=1}^{d} \max_{w \in B_{k_i}} \left| \left[ \mathcal{F} \psi_{0,0}^{(q_i)} \right] (\omega) \right|^2 \lesssim \prod_{i=1}^{d} 2^{-|k_i - 1|}.$$

Similarly for $j > 1$

$$\mu(U^{(\mathbf{k},j)}) \lesssim 2^{\sum_{i=1}^{d} k_i} \max_{\mathbf{q} \in \{0,1\}^d \setminus \{0\}} \prod_{i=1}^{d} \max_{w \in B_{k_i}} \max_{p_i \in \{0,...,2^{j-1}-1\}} \left| \left[ \mathcal{F} \psi_{j-1,p_i}^{(q_i)} \right] (\omega) \right|^2 \lesssim \max_{\mathbf{q} \in \{0,1\}^d \setminus \{0\}} \prod_{i=1}^{d} 2^{-|k_i - j| - 2q_i(j-k_i)_+}.$$

The maximum value of this estimate is obtained when the non-zero component of $\mathbf{q}$ corresponds to the maximum value of $k_i$. This gives precisely (5.6). □

Before proceeding with the Walsh–Haar–wavelet case, we recall the following lemma [6].

**Lemma 5.5.** *Let $\omega$ and $j \geq 0$ be integers so that $2^j \leq \omega < 2^{j+1}$ and let $\Delta_k^j = [k2^{-j}, (k+1)2^{-j})$ for $k \in \mathbb{Z}_{\geq 0}$. Then $v_\omega$ is constant on each of the intervals $\Delta_k^{j+1}$, $k \in \{0, \ldots, 2^{j+1} - 1\}$. Each of the intervals $\Delta_k^j$ can be decomposed into the intervals $\Delta_{2k}^{j+1}$ and $\Delta_{2k+1}^{j+1}$, where $v_\omega$ is equal to 1 on exactly one of them and equal to $-1$ on the other. When $\omega = 0$, we have $v_\omega \equiv 1$.*

**Lemma 5.6** (**Coherence bound for Walsh case**)**.** *Consider the $d$-dimensional Walsh–Haar–wavelet matrix with blocks $U^{(\mathbf{k},j)}$, then the local coherences satisfy*

$$\mu(U^{(\mathbf{k},j)}) \lesssim \begin{cases} \prod_{i=1}^d 2^{-|k_i - j|} & \text{if } k_i \leq j \text{ for } i = 1, \ldots, d \text{ with at least one equality,} \\ 0 & \text{otherwise} \end{cases}. \quad (5.8)$$

*It follows that*

$$\sum_{j=1}^r s_j \mu(U^{(\mathbf{k},j)}) \lesssim s_{\|\mathbf{k}\|_{l^\infty}} \prod_{i=1}^d 2^{-|k_i - \|\mathbf{k}\|_{l^\infty}|} = \mathcal{M}_{\mathcal{W}}(\mathbf{s}, \mathbf{k}). \quad (5.9)$$

*Proof.* We begin with some computations in the one-dimensional case. Let $I_{j,p} = \{p2^{r-j}, \ldots, (p+1)2^{r-j} - 1\}$. We recall that $\operatorname{supp}(\psi_{j,p}^{(0)}) = \operatorname{supp}(\psi_{j,p}^{(1)}) = I_{j,p}$. Using Lemma 5.5 it is clear that for $2^m \leq \omega < 2^{m+1}$, $\varrho_\omega$ is constant on $I_{m+1,k}$, for $k \in \{0, \ldots, 2^{m+1} - 1\}$ and that for any pair $I_{m+1,2t}, I_{m+1,2t+1}$, $\varrho_\omega$ changes sign. For $\omega = 0$, we have that $\varrho_\omega$ is all constant. Keeping track of the supports gives the relations

$$\left|\left\langle \psi_{j,p}^{(0)}, \varrho_\omega \right\rangle\right| = \begin{cases} 2^{-j/2} & \text{if } \omega < 2^j \\ 0 & \text{otherwise} \end{cases}, \quad \text{and} \quad \left|\left\langle \psi_{j,p}^{(1)}, \varrho_\omega \right\rangle\right| = \begin{cases} 2^{-j/2} & \text{if } 2^j \leq \omega < 2^{j+1} \\ 0 & \text{otherwise} \end{cases}.$$

In particular, we can rewrite this as $|\langle \psi_{j,p}^{(0)}, \varrho_\omega \rangle| = 2^{-j/2} = 2^{-k/2} 2^{-(j-k)/2}$ if $\omega \in B_k$, $k \leq j$ and 0 otherwise, and note that $|\langle \psi_{j,p}^{(1)}, \varrho_\omega \rangle| = 2^{-j/2}$ if $\omega \in B_{j+1}$ and 0 otherwise.

Turning to the general $d$-dimensional case. The above computations immediately give that $\mu(U^{(\mathbf{k},1)}) \lesssim \prod_{i=1}^d \delta_{k_i,1}$, where $\delta_{i,j}$ is the Kronecker-delta. Similarly for $j > 1$

$$\mu(U^{(\mathbf{k},j)}) = \left|B_{\mathbf{k}}^{(d)}\right| \max_{\mathbf{q} \in \{0,1\}^d \setminus \{0\}} \prod_{i=1}^d \max_{\omega_i \in B_{k_i}} \max_{p_i \in \{0,\ldots,2^{j-1}-1\}} \left|\langle \varrho_{\omega_i}, \psi_{j-1,p_i}^{(q_i)} \rangle\right|^2$$

$$\lesssim 2^{\sum_{i=1}^d k_i} \max_{\mathbf{q} \in \{0,1\}^d \setminus \{0\}} \prod_{i=1}^d \left(\delta_{q_i,0} \delta_{k_i < j} 2^{-k_i - |k_i - j|} + \delta_{q_i,1} \delta_{k_i,j} 2^{-k_i}\right)$$

$$\lesssim \max_{\mathbf{q} \in \{0,1\}^d \setminus \{0\}} \prod_{i=1}^d \left(\delta_{q_i,0} \delta_{k_i < j} 2^{-|k_i - j|} + \delta_{q_i,1} \delta_{k_i,j}\right).$$

This estimate is zero unless $k_i \leq j$ and at least one of the $k_i$ is equal to $j$. In this case the maximum corresponds to $q_i = 1$ if $k_i = j$ and $q_i = 0$ otherwise. This gives precisely (5.8). $\square$

**5.3. Proof of Theorem 4.** For the benefit of the reader, we recall that $A = P_{\mathcal{I}} D V \Psi$. We apply Proposition 5.3, noting that the $t_j$ in (5.4) satisfy

$$t_j \lesssim \frac{\xi(\mathbf{s}, \mathbf{M}, w)}{w_{(j)}^2} \leq s_j \cdot \kappa(\mathbf{s}, \mathbf{M}, w), \qquad t \lesssim s \cdot \kappa(\mathbf{s}, \mathbf{M}, w). \quad (5.10)$$

Therefore $\sum_{j=1}^r t_j \mu(U^{\mathbf{k},j}) \lesssim \kappa(\mathbf{s}, \mathbf{M}, w) \sum_{j=1}^r s_j \mu(U^{\mathbf{k},j})$. Combining with (5.10), note that (5.5) holds if

$$m_{\mathbf{k}} \gtrsim \kappa(\mathbf{s}, \mathbf{M}, w) \cdot \left(\sum_{j=1}^r s_j \mu(U^{\mathbf{k},j})\right) \cdot \mathcal{L}, \quad \text{where} \quad (5.11)$$

$$\mathcal{L} = \frac{r \cdot \log(2m)}{\log(2)} \cdot \log^2\left(s \cdot \kappa(\mathbf{s}, \mathbf{M}, w)\right) \cdot \log(N) + \log(\epsilon_{\mathbb{P}}^{-1}) = d \cdot r^2 \cdot \log(2m) \cdot \log^2\left(s \cdot \kappa(\mathbf{s}, \mathbf{M}, w)\right) + \log(\epsilon_{\mathbb{P}}^{-1}),$$

since $N = 2^{r \cdot d}$. In the Fourier sampling case, by Lemma 5.4, (5.11) holds if (1.22) holds. Similarly, in the Walsh sampling case, by Lemma 5.6, (5.11) holds if (1.23) holds. By Proposition 5.3, with probability at least $1 - \epsilon_\mathbb{P}$, $A$ satisfies the weighted rNSPL of order $(\mathbf{s}, \mathbf{M})$ with constants $\rho = 1/2$ and $\gamma = \sqrt{2}$. The conclusion of Theorem 3 then holds for the uniform recovery of the Haar wavelet coefficients $x = \Psi c \in \mathbb{C}^N$.

For the final part, we use Theorem 3. The only difference is that we have to compose the NNs with (an approximation of) the matrix $\Psi^*$ to recover approximations of $c$ from approximations of $x = \Psi c$. Recall that

$$\mathcal{Z} = \max\left\{1, \frac{\max_{j=1,\ldots,r} w_{(j)} \sqrt{(M_j - M_{j-1})}}{\sqrt{\xi(\mathbf{s}, \mathbf{M}, w)}}\right\}$$

and set $n_0 = \lceil \log\left(\delta^{-1} \mathcal{Z}\right) \kappa^{1/4} \mathcal{Z} \rceil$. Let $p$ be as in Theorem 3 and let $n_1 \in \mathbb{Z}_{\geq 0}$ such that $n_0 = n_1 p + n_2$ for $n_1 \in \{0, \ldots, p-1\}$ (the $n$ from the statement of the theorem corresponds to $n_1 p$). Set $\phi(y) = \Psi^* \left[\phi_{n_1}(y, 0)\right]$, where $\phi_{n_1}$ denotes the NN from Theorem 3 with $b_1 = 1$, $b_2 = \|A\| + \delta$ and $v = e^{-1}$. Strictly speaking, we need to approximate $\|A\|$ and $e^{-1}$, and also apply a rational approximation of the matrix $\Psi^*$ instead of $\Psi^*$, but we have avoided this extra notational clutter (the associated approximation errors can be made smaller than $\kappa^{1/4} \delta$ since the vectors we apply the matrix to are uniformly bounded). Now suppose that $y = P_\mathcal{I} DVc + e \in \mathcal{J}(\delta, \mathbf{s}, \mathbf{M}, w)$, and notice that for $\|c\|_{l^2} \leq 1$ we have that $\|y\|_{l^2} \leq \|A\| + \|e\|_{l^2} \leq b_2$ since $\Psi$ is an isometry. Then, since $C_1, C_2 \sim \kappa^{1/4}$ (using that $\kappa \geq 1$), (1.16) implies that

$$\|\phi(y) - c\|_{l^2} = \|\phi_{n_0}(y, 0) - \Psi c\|_{l^2} \lesssim \kappa^{1/4} \delta + b_2 \kappa^{1/4} e^{-n_1}.$$

The theorem follows if we can prove that $b_2 e^{-n_1} \lesssim \delta$.

Let $\mathbf{t}$ be as in (5.4) and let $\Delta_1, \Delta_2, \ldots$ be a partition of $\{1, \ldots, N\}$ such that each support set is $(\mathbf{t}, \mathbf{M})$-sparse. We can choose such as partition with at most

$$\max_{j=1,\ldots,r} \left\lceil \frac{M_j - M_{j-1}}{t_j} \right\rceil \lesssim \max_{j=1,\ldots,r} \left\lceil \frac{M_j - M_{j-1}}{\min\{\xi(\mathbf{s}, \mathbf{M}, w)/w_{(j)}^2, M_j - M_{j-1}\}} \right\rceil$$

sets. The proof of Proposition 5.3 shows that $A$ satisfies the RIPL of order $(\mathbf{t}, \mathbf{M})$ and hence for any $x \in \mathbb{C}^N$,

$$\|Ax\|_{l^2} \leq \sum_i \|A(x_{\Delta_i})\|_{l^2} \lesssim \sum_i \|x_{\Delta_i}\|_{l^2} \lesssim \max_{j=1,\ldots,r} \sqrt{\left\lceil \frac{M_j - M_{j-1}}{\min\{\xi(\mathbf{s}, \mathbf{M}, w)/w_{(j)}^2, M_j - M_{j-1}\}} \right\rceil} \|x\|_{l^2},$$

where we have used Hölder's inequality in the last step. It follows that $\|A\| \lesssim \mathcal{Z}$ and hence that $p \lesssim \kappa^{1/4} \mathcal{Z}$ and $b_2 \lesssim \mathcal{Z}$. This implies that $n_1 \gtrsim \log\left(\delta^{-1} \mathcal{Z}\right)$ and $b_2 e^{-n_1} \lesssim \delta$, completing the proof. $\square$



\end{document}